\documentclass[journal]{IEEEtai}

\usepackage[colorlinks,urlcolor=magenta,linkcolor=magenta,citecolor=magenta]{hyperref}

\usepackage{color,array}

\usepackage{graphicx}

\usepackage{amsmath,amsfonts,amssymb}

\usepackage{tabularx,booktabs}
\usepackage[flushleft]{threeparttable}
\usepackage[dvipsnames]{xcolor}
\usepackage[labelfont=normalfont]{subcaption} 
\usepackage{amsmath}
\usepackage{multirow}
\usepackage{color,soul}

\usepackage{cite}

\usepackage{algorithm}
\usepackage{algpseudocode}
\usepackage{relsize}

\usepackage{scalerel}
\usepackage{tikz}
\usetikzlibrary{svg.path}

\definecolor{orcidlogocol}{HTML}{A6CE39}
\tikzset{
    orcidlogo/.pic={
        \fill[orcidlogocol] svg{M256,128c0,70.7-57.3,128-128,128C57.3,256,0,198.7,0,128C0,57.3,57.3,0,128,0C198.7,0,256,57.3,256,128z};
        \fill[white] svg{M86.3,186.2H70.9V79.1h15.4v48.4V186.2z}
        svg{M108.9,79.1h41.6c39.6,0,57,28.3,57,53.6c0,27.5-21.5,53.6-56.8,53.6h-41.8V79.1z M124.3,172.4h24.5c34.9,0,42.9-26.5,42.9-39.7c0-21.5-13.7-39.7-43.7-39.7h-23.7V172.4z}
        svg{M88.7,56.8c0,5.5-4.5,10.1-10.1,10.1c-5.6,0-10.1-4.6-10.1-10.1c0-5.6,4.5-10.1,10.1-10.1C84.2,46.7,88.7,51.3,88.7,56.8z};
    }
}

\newcommand\orcidicon[1]{\href{https://orcid.org/#1}{\mbox{\scalerel*{
                \begin{tikzpicture}[yscale=-1,transform shape]
                \pic{orcidlogo};
                \end{tikzpicture}
            }{|}}}}

\begin{document}

\title{Haze Removal via Regional Saturation-Value Translation and Soft Segmentation}


\author{Le-Anh Tran$^{\textsuperscript{\orcidicon{0000-0002-9380-7166}}}$,~\IEEEmembership{Student Member,~IEEE,} Dong-Chul Park$^{\textsuperscript{\orcidicon{0009-0007-7442-2301}}}$,~\IEEEmembership{Senior Member,~IEEE}
\thanks{Le-Anh Tran and Dong-Chul Park are with the Department
of Electronics Engineering, Myongji University, Yongin, 17058, South Korea.}
\thanks{Corresponding author: Dong-Chul Park (parkd@mju.ac.kr)}}

\markboth{Manuscript, Vol. 00, No. 0, Month 202X}
{Le-Anh Tran \MakeLowercase{\textit{et al.}}: Haze Removal via Regional Saturation-Value Translation and Soft Segmentation}

\maketitle

\begin{abstract}
This paper proposes a single image dehazing prior, called Regional Saturation-Value Translation (RSVT), to tackle the color distortion problems caused by conventional dehazing approaches in bright regions. The RSVT prior is developed based on two key observations regarding the relationship between hazy and haze-free points in the HSV color space. First, the hue component shows marginal variation between corresponding hazy and haze-free points, consolidating a hypothesis that the pixel value variability induced by haze primarily occurs in the saturation and value spaces. Second, in the 2D saturation-value coordinate system, most lines passing through hazy-clean point pairs are likely to intersect near the atmospheric light coordinates. Accordingly, haze removal for the bright regions can be performed by properly translating saturation-value coordinates. In addition, an effective soft segmentation method based on a morphological min-max channel is introduced. By combining the soft segmentation mask with the RSVT prior, a comprehensive single image dehazing framework is devised. Experimental results on various synthetic and realistic hazy image datasets demonstrate that the proposed scheme successfully addresses color distortion issues and restores visually appealing images. The code of this work is available at \url{https://github.com/tranleanh/rsvt}.
\end{abstract}



\section{Introduction}
\label{sec:introduction}

\IEEEPARstart{N}{owadays}, the rise of industrialization has led to air pollution all over the world, resulting in dust and haze in the natural atmosphere. These occurrences present significant absorption and scattering of light during its propagation process and introduce detrimental effects to the radiance of the scene. As a result, the scene radiance received by the camera is degraded leading to the generation of low-quality images with a loss in contrast and low color fidelity. This issue has posed challenges for various computer applications that involve outdoor images such as target detection using aerospace devices. Therefore, there is a practical and strong demand for research on effective image dehazing techniques. Generally, haze removal algorithms can be classified into two main categories: \textit{prior-based} and \textit{deep learning-based} methods, each genre has its own strengths and weaknesses. Prior-based methods are effective for restoring visibility but they often result in over-saturation and artifacts in the output \cite{zhao2021refinednet}. On the other hand, deep learning-based approaches have the potential to improve the realism of the restored images, yet they heavily rely on training datasets containing hazy-clean image pairs from the same scene which can be costly for data preparation \cite{zhao2021refinednet}. It implies that there is still room for improvement in image dehazing research.

\begin{figure}
  \centering  
  \includegraphics[width=1.0\linewidth]{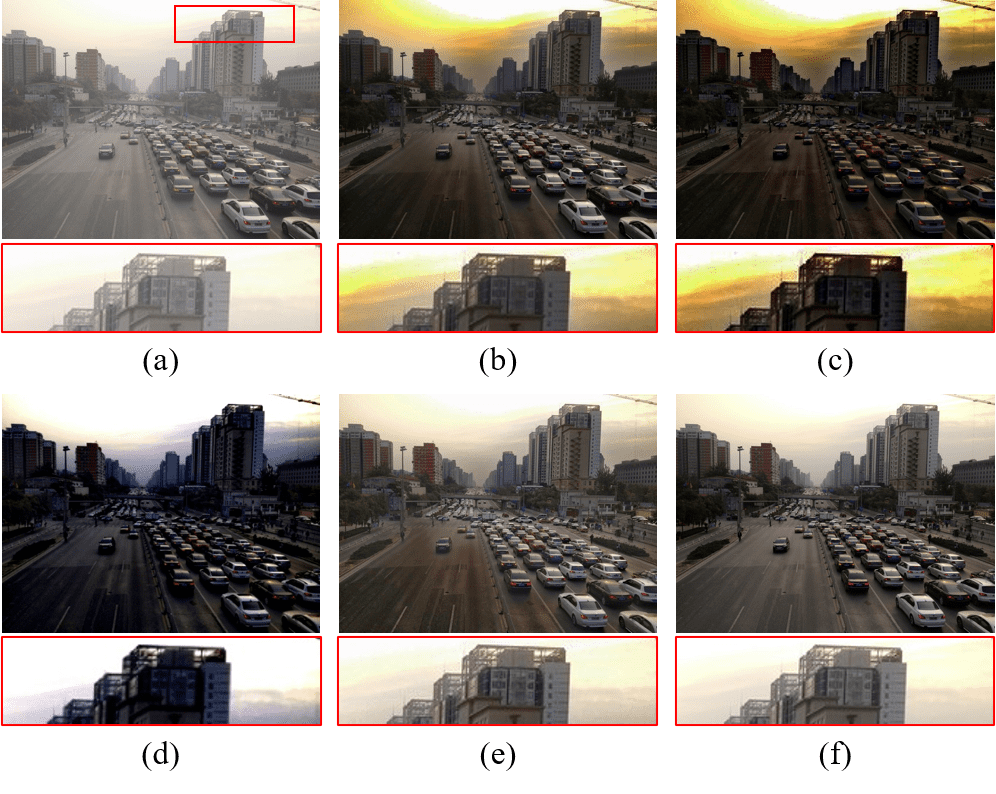} 
  \caption{Dehazing results of various approaches on hazy images containing bright regions: (a) input, (b) DCP \cite{he2010single}, (c) CEP \cite{bui2017single}, (d) NLID \cite{berman2016non}, (e) the proposed RSVT method, and (f) clean image.}
  \label{fig01:first_comparison}
\end{figure}

\begin{figure*}
  \centering  
  \includegraphics[width=1.0\linewidth]{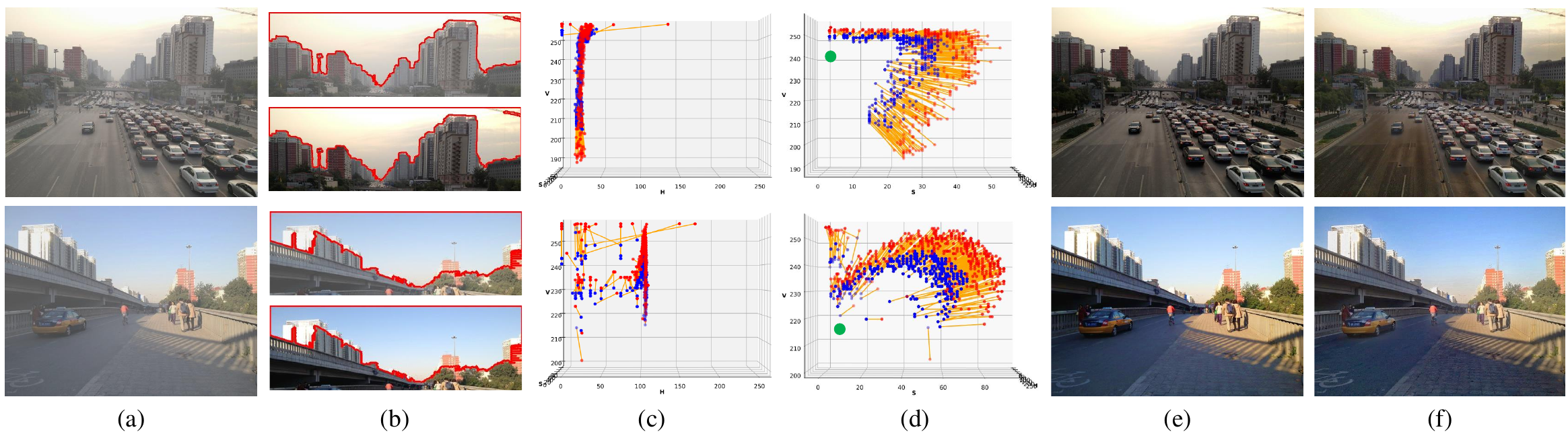}
  \caption{Regional Saturation-Value Transition: (a) hazy image, (b) sky regions in hazy (top) and clean (bottom) images marked in red color, (c,d) hazy (blue) and clean (red) points for the sky region in HSV color space from Hue-Value and Saturation-Value perspectives, respectively (orange segments connect corresponding hazy-clean pairs), (e) clean image, and (f) dehazed image by the proposed method.}
  \label{fig02:rsvt_examples}
\end{figure*}

This paper mainly focuses on addressing a common issue faced by conventional prior-based dehazing algorithms when dealing with hazy images that contain bright regions such as the sky or white objects. Typically, such images consist of two main parts: \textit{foreground} and \textit{background}. The background is usually sky regions while the foreground includes objects that are closer to the observer. While most prevailing prior-based dehazing algorithms perform quite well in restoring the visibility of the foreground regions, they often struggle with color distortion in the background areas. To address this issue, several approaches have been proposed which have tried to separately process the sky regions \cite{shi2014single, wang2017dehazing, salazar2020fast, yu2016image}. However, it is very difficult or unattainable to segment images captured under dense haze circumstances, and in fact, hard segmentation may never yield accurate results, particularly in dim areas where there is no distinct boundary between the foreground and background regions. Even when the sky region is not visible, color distortion can still occur in the output due to high-intensity regions like water surfaces or whitish buildings, as can be observed in Fig. \ref{fig01:first_comparison}b-\ref{fig01:first_comparison}d. From those observations, we have analyzed the relationship between hazy points and their respective haze-free points for the bright regions in the HSV color space. This analysis has yielded two significant findings. First, the difference in terms of hue component for a pair of hazy-clean points is marginal, as shown in Fig. \ref{fig02:rsvt_examples}c. This suggests that the impact of haze on the hue channel is insignificant, and the pixel value variations caused by haze may mainly occur in the saturation and value spaces. Second, when considering the 2D coordinate system formed by the saturation and value components, it is observed that most of the lines passing through corresponding pairs of hazy-clean points, referred to as \textit{S-V lines}, tend to intersect around the atmospheric light coordinates, as illustrated in Fig. \ref{fig02:rsvt_examples}d. Based on these two observations, we can assume that the haze-free pixel can be estimated by shifting the hazy point along the corresponding S-V line by an appropriate amount. In order to validate this hypothesis, various statistical analyses have been conducted. Consequently, an innovative image dehazing prior called \textit{Regional Saturation-Value Translation} (RSVT) is proposed to address the above-mentioned limitations of existing dehazing methods. On the other hand, dark channel prior (DCP) \cite{he2010single} is a widely recognized method in image dehazing field. DCP is developed based on the assumption that the intensity of at least one color channel is significantly lower in haze-free areas when compared with hazy regions. DCP is known for its efficiency and ability to produce favorable restoration outcomes in close-range and non-sky scenes. However, DCP is ineffective for bright regions such as the sky or white objects, as these areas naturally exhibit high-intensity pixels even in haze-free scenes, thereby violating the assumption of DCP. Accordingly, an effective single image dehazing scheme is proposed which synergies the potency of the proposed RSVT prior in dealing with the bright regions and the ability of DCP in processing foreground area.

In contrast to other approaches that adopt hard segmentation to decompose a hazy image into background and foreground regions \cite{shi2014single, yu2016image, wang2017dehazing}, the proposed method takes a different approach by segregating the input image into three main parts: \textit{hard foreground} (regions that are completely classified as foreground), \textit{hard background} (very bright areas such as the daytime sky or whitish objects), and \textit{middle ground} (dim areas that cannot be definitively classified as either background or foreground). To this end, we adopt a soft segmentation process based on a morphological min-max channel. In our proposed framework, the RSVT prior and DCP methods handle the hard background and hard foreground regions, respectively, while the restoration for the middle ground is the weighted average solution of these two algorithms. The proposed approach's effectiveness is typically illustrated through the result shown in Fig. \ref{fig01:first_comparison}e.  As can be observed from Fig. \ref{fig01:first_comparison}, it is evident that the proposed scheme significantly reduces color distortion and successfully recovers visually appealing images when compared with the other algorithms.

Part of this work was presented in \cite{tran2023single} where the preliminary idea of the RSVT prior was introduced. In this paper, we improve the image decomposition process and conduct further experiments to prove the effectiveness of the proposed dehazing scheme. The remainder of this paper is organized as follows: Section \ref{sec:preliminaries} presents the background materials including the haze imaging model, a brief review of single image dehazing approaches, and the guided filtering algorithm. The assumptions along with statistical analyses for the verification of the proposed prior are presented in Section \ref{sec:regional-saturation-value-variability}. In Section \ref{sec:proposed-dehazing-method}, a soft segmentation method for decomposing an input image is introduced, afterward a comprehensive single image dehazing framework is proposed. Section \ref{sec:experiments-discussions} presents the experimental results and comparisons. Section \ref{sec:conclusions} concludes the paper.


\section{Preliminaries}
\label{sec:preliminaries}

This section briefly reviews related background materials including the haze imaging model, prevailing single image dehazing methods, and the guided filtering algorithm.

\subsection{Haze Imaging Model}
\label{sec:haze-imaging-model}

Generally, a hazy image can be modeled as a per-pixel convex combination between the actual scene radiance and the global atmospheric light \cite{tran2022anovel}, this model is known as the scattering atmospheric model or Koschmieder’s law \cite{middleton1954vision}:
\begin{equation}
I(x)=J(x)t(x) + A(1-t(x)),       \label{equation1}
\end{equation} where $I(x)$, $J(x)$, $A$, and $t(x)$ denote the observed intensity, the scene radiance, the global atmospheric light, and the transmission map, respectively. From Eq. (\ref{equation1}), most of the prior-based methods try to estimate the actual scene radiance $J(x)$ via the following function:
\begin{equation}
J(x) = \frac{I(x) - A}{t(x)} + A.    \label{equation3} 
\end{equation} 
Relying on this recovering function, restored images can be obtained by solving the unknowns $t(x)$ and $A$. The value of $A$ is generally assumed to be a high and constant value, while that of $t(x)$ for the bright and distant regions is usually very small and is typically set to 0.1 \cite{he2010single}. Accordingly, a small difference in the magnitudes of $I(x)$ and $A$, for example, $|I(x) - A| = 10$, can cause a shift of 100 intensity levels in the restored image which yields noise and distorted color. Hence, the conventional hazy imaging model may be invalid for performing dehazing on images that contain bright and sky areas.

\begin{figure*}
  \centering  
  \begin{subfigure}{0.3\linewidth}
    \includegraphics[width=1.0\linewidth]{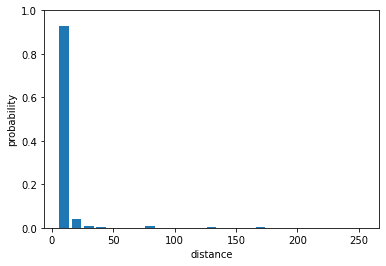} 
    \caption{}
    \label{fig03:statistics-a}
  \end{subfigure}
  \begin{subfigure}{0.3\linewidth}
    \includegraphics[width=1.0\linewidth]{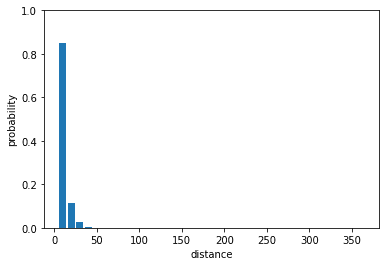} 
    \caption{}
    \label{fig03:statistics-b}
  \end{subfigure}
  \begin{subfigure}{0.305\linewidth}
    \includegraphics[width=1.0\linewidth]{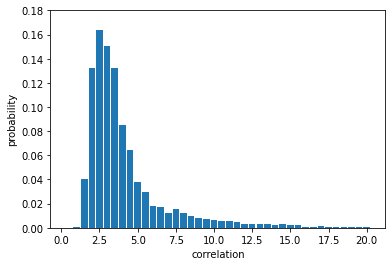}
    \caption{}
    \label{fig03:statistics-c}
  \end{subfigure}
  \caption{Statistical analyses: (a) differences between the hue components of hazy and haze-free point pairs within the sky regions
(each bin stands for 10 distance levels), (b) distances between the intersections of S-V lines and the global atmospheric light coordinates in the Saturation-Value coordinate system (each bin stands for 10 distance levels), and (c) distribution of the correlation between the S-V ratio and the transmission.}
  \label{fig03:statistics}
\end{figure*}

\subsection{Brief Review on Image Dehazing}
\label{sec:brief-review}

As mentioned in Section \ref{sec:introduction}, haze removal algorithms can be categorized into two groups: prior-based and deep learning-based methods. 

Prior-based methods have been developed based on strong prior knowledge or assumptions. Fattal \textit{et al.} \cite{fattal2008single} proposed a method that estimates the albedo of the scene and the transmission map under the assumption that the transmission and the surface shading are locally uncorrelated. This method, however, shows much difficulty when dealing with dense haze and may be ineffective if the assumption does not hold. Meng \textit{et al.} \cite{meng2013efficient} introduced a boundary constraint and contextual regularization (BCCR) method which enforces constraints on the image boundaries and applies contextual regularization to enhance the sharpness of dehazed images. Zhu \textit{et al.} \cite{zhu2015fast} proposed a color-attenuation prior (CAP) which models the scene depth of a hazy image and learns the model parameters through supervised learning. Berman \textit{et al.} \cite{berman2016non} introduced a non-local image dehazing (NLID) prior based on the observation that hazy pixels in RGB color space are distributed along lines passing through the atmospheric light coordinates. Even though showing promising results in various cases, this algorithm sometimes produces an estimated transmission smaller than a lower bound which results in a dark output and loss of details. Among popular prior-based approaches, dark channel prior (DCP) \cite{he2010single} has gained significant attention. DCP is developed based on a key observation that most local patches in haze-free outdoor images contain some pixels with very low intensity in at least one color channel. The prior can directly estimate the haze thickness and the haze-free image is recovered via Eq. (\ref{equation3}) with certain predetermined constraints. However, DCP encounters challenges when the scene object is similar to the airlight over a large local region and when there is no shadow cast on the objects. Although prior-based dehazing algorithms can achieve acceptable results in various circumstances, they are often incompetent in processing high-intensity regions (such as daytime sky and bright objects) and densely hazy areas, as discussed in Section \ref{sec:introduction}. Accordingly, dehazing for images containing sky regions has emerged as a niche research area that has also gained considerable attention. Numerous research works have explored this issue by adopting a fundamental approach of decomposing a hazy image into sky and non-sky regions. Wang \textit{et al.} \cite{wang2017dehazing} proposed to utilize the quad-tree method to identify a seed point for the sky regions and employed region-growing segmentation with a Gaussian filter applied afterward for image smoothing. This approach, however, encounters difficulties when the sky region is intersected by objects like trees or buildings. Sebastián \textit{et al.} \cite{salazar2020fast} adopted a similar procedure but replaced the quad-tree method with the local Shannon entropy calculation. Nevertheless, growing-based methods require accurate seed point estimation as well as the growing process control which is time-consuming in order to avoid under and/or over-segmentation \cite{nguyen20133d}. In addition to conventional methods, several studies \cite{liu2017single, liba2020sky} have also incorporated neural networks for sky segmentation. However, it is important to note that deep learning models are more computationally expensive when compared with pure image processing techniques. Moreover, it should be acknowledged that sky segmentation is not the primary objective in a dehazing system.  

On the other extreme, deep learning models have demonstrated their remarkable ability to learn non-linear mappings in image translation tasks. Various learning-based dehazing approaches, such as \cite{li2017aod, ren2016single, cai2016dehazenet}, have been proposed that can directly estimate $t(x)$ or $J(x)$ from a given input hazy image. These approaches commonly adopt convolutional neural networks (CNNs) as the primary backbone. However, the training process of CNNs requires extensive datasets with pairs of hazy and haze-free images captured in identical scenes, which are too difficult and/or unachievable to prepare under real-world circumstances. A common approach to this issue is to synthesize hazy image data using the haze imaging model, but there always exists a certain gap between the synthetic and real-world data. Furthermore, learning-based models often suffer from overfitting the training data, resulting in inadequate generalization capabilities toward realistic haze. 

Even though learning-based approaches are dominant in the present era of deep learning, we believe that innovative priors also hold significant importance owing to their consistent statistical nature and strong underlying assumptions. Moreover, learning-based approaches often entail expensive computational costs, whereas prior-based methods offer simplicity and efficient deployment. This has inspired us to come up with a novel and robust prior in an attempt to introduce a comprehensive haze removal framework that can restore high-quality haze-free images.

\subsection{Guided Filtering}
\label{sec:guidedfilter}

Guided filtering \cite{he2012guided} is an image processing technique for edge-preserving image smoothing by utilizing a bilateral filter to eliminate noise and unexpected texture artifacts. The process takes a noisy image and a guidance image as input signals, and yields a corrected image. Given a guidance image $I$, an input image $p$, and an output image $q$, the fundamental assumption of the guided filter is the existence of a local linear model between the guidance $I$ and the filtering output $q$ which can be determined as follows:
\begin{equation}
q_i = a_k I_i + b_k, \forall i \in \omega_k,
\end{equation}
where $i$ is the pixel index, $k$ is the index of a local patch $\omega$, and ($a_k$, $b_k$) are linear coefficients assumed to be constant in $\omega_k$. By minimizing the following cost function:
\begin{equation}
E(a_k,b_k) = \sum_{i \in \omega_k} ((a_k I_i + b_k - p_i)^2 + \epsilon a_k^2),
\end{equation}
where $\epsilon$ is a regularization parameter that controls the degree of smoothness, the solution for ($a_k$, $b_k$) is given by:
\begin{equation}
a_k = \frac{\frac{1}{|\omega|}\sum_{i \in \omega_k} I_i p_i - \mu_k \bar{p}_k}{\sigma_k^2 + \epsilon},
\end{equation}
\begin{equation}
b_k = \bar{p}_k - a_k\mu_k,
\end{equation}
where $\mu_k$ and $\sigma_k$ represent the mean and variance of $I$ in $\omega_k$, respectively, while $|\omega|$ denotes the number of pixels in $\omega_k$. Ultimately, the computation of the filtering output is carried out as follows:
\begin{equation}
q_i = GuidedFilter(p_i, I_i) = \bar{a_i} I_i + \bar{b_i},
\end{equation}
where $\bar{a_i}$ and $\bar{b_i}$ correspondingly are the average of $a$ and $b$ on the window $\omega_i$ centered at $i$.


\section{Regional Saturation-Value Variability}
\label{sec:regional-saturation-value-variability}

The proposed prior is formulated based on two primary observations made on the bright areas of hazy-clean image pairs in the HSV color space. First, the impact of haze on the hue channel is found to be insignificant. Second, in the 2D saturation-value coordinate system, it is observed that the majority of S-V lines tend to intersect around the atmospheric light coordinates. In order to verify these observations, a total of 1,000 hazy-clean image pairs which include high-intensity pixels are selected from the RESIDE database \cite{li2018benchmarking} for various statistical analyses. The high-intensity regions are easily identified by applying image thresholding segmentation with a high threshold and manual verification. Based on these efforts, three assumptions are formulated so as to confirm the correctness of the proposed prior.

\subsection{The 1st Assumption}
\label{subsec:1st-assumption}

The first assumption states that: the hue component of a pixel in a bright region of a hazy image is relatively similar to that of the corresponding pixel in a haze-free image, as depicted in Fig. \ref{fig02:rsvt_examples}c. This observation raises an assumption that the variation in pixel intensity caused by haze primarily occurs in the saturation and value spaces, as illustrated in Fig. \ref{fig02:rsvt_examples}d, and the dehazing process for bright hazy regions can be carried out without considering the hue channel.

To validate this assumption, the absolute differences between the hue components of hazy and haze-free point pairs within the bright areas are measured. This measurement yields a probability distribution as shown in Fig. \ref{fig03:statistics-a}, which reveals that approximately 93\% of the distances between the hue components are below 10, thus confirming the validity of our first assumption.

\subsection{The 2nd Assumption}
\label{subsec:2nd-assumption}

The second assumption states that: when disregarding the impact of haze on the hue space, it can be observed that the majority of S-V lines, which represent the lines connecting each pair of hazy-clean points, are likely to intersect near the atmospheric light coordinates which are typically found within the brightest pixels of a hazy image. 

In order to verify this assumption, the following procedure for each hazy-clean image pair is performed. First, the intersection point, denoted as $E$, of all the S-V lines is calculated. Subsequently, the Euclidean distance between $E$ and the estimated global atmospheric light $A$ is measured, where the estimation of $A$ is achieved by randomly applying DCP \cite{he2010single} and NLID \cite{berman2016non} methods:
\begin{equation}
||E - A|| =  \sqrt{(S_E - S_A)^2 + (V_E - V_A)^2},  \label{equation4}
\end{equation}
where $S$ and $V$ denote the saturation and value components, respectively. Note that $S, V \in [0,255]$, thus the maximum distance is $ \sqrt{255^2 + 255^2} \approx 360$. Note also that all the S-V lines do not intersect perfectly at one single point, hence, we collect every intersection of each pair of S-V lines and $E$ is computed as the average of all the intersections.

Fig. \ref{fig03:statistics-b} illustrates the likelihood of the measured distances between the intersection of the S-V lines $E$ and the global atmospheric light coordinates $A$. It is evident from the graph that the majority of distances are in proximity to small values, suggesting that the intersection is typically near the atmospheric light. This statistical analysis provides robust evidence in favor of the third assumption presented in the next section.

\subsection{The 3rd Assumption}
\label{subsec:3rd-assumption}

In the analysis presented in Fig. \ref{fig02:rsvt_examples}d, it can be observed that within the HSV color space, a clean point $c(x)$ is derived by shifting the corresponding hazy point $h(x)$ away from the intersection $E$ (indicated as a green point) by a magnitude that is directly proportional to the distance between the hazy point and the intersection, denoted as $||h(x) - E||$. The challenge at hand is to establish the relationship between $||h(x) - E||$ and any potential indicators associated with the distribution of haze in the input hazy image. Interestingly, upon revisiting the haze imaging model, the transmission in the RGB color space is represented as \cite{he2010single}:
\begin{equation}
t(x) =  \frac{||I(x) - A||}{||J(x) - A||},  \label{equation5}
\end{equation}
which implies that the proportion between the distance from the hazy point to the atmospheric light point and the distance from the haze-free point to the atmospheric light point in the RGB color space directly represents the transmission. This leads us to the exploration of a similar type of proportion in the HSV color space. To this end, we initially eliminate two types of lines: one consists of all the S-V lines where the hazy or haze-free points have a value or saturation component that is close to 0 or 255, as these points' magnitudes may have been truncated due to the 8-bit image property; and the other category includes excessively short lines which can be regarded as "outliers" when calculating the distribution. Subsequently, a specific proportion, referred to as the \textit{S-V ratio} $r(x)$ for each S-V line within the saturation-value space, is computed:
\begin{equation}
r(x) =  \frac{||h(x) - E||}{||c(x) - E||},  \label{equation6}
\end{equation}
and we examine the likelihood of the correlation between the S-V ratio and the transmission, denoted as $R(x)$:
\begin{equation}
R(x) =  \frac{r(x)}{t_b(x)},  \label{equ11}
\end{equation}
where $t_b(x)$ represents the transmission of the background region. In order to compute $t_b(x)$, we refer to the method described in \cite{huang2021image} which utilizes the dark channel with a small window size, i.e., $3 \times 3$, to initially estimate a coarse transmission $\tilde{t_b}(x)$:
\begin{equation}
\tilde{t_b}(x) =  1 - \omega \min_{y \in p_{3 \times 3}(x)} (\min_{c \in (r,g,b)} \frac{I_c(y)}{A_c}),  \label{equation7}
\end{equation}
where $\omega$ ($0 < \omega \leq 1$) denotes a predefined parameter to optionally keep a small amount of haze for the distant objects ($\omega$ is typically set to 0.95 \cite{he2010single}), afterward $t_b(x)$ can be obtained by refining $\tilde{t_b}(x)$ using guided filtering:
\begin{equation}
t_b(x) =  GuidedFilter(\tilde{t_b}(x), I_{gray}),
\end{equation}
where $I_{gray}$ denotes the gray image of the input. 

Fig. \ref{fig03:statistics-c} illustrates the obtained distribution of $R(x)$, which shows that the highest frequencies for $R(x)$ occur within the interval $[2.0,5.0]$ with approximately 85-90\% of cases. As a rough assumption, we consider setting $R(x)$ as a constant for any input image such that the intensities of the restored image are minimally affected when $R(x)$ varies within the range of its highest frequencies (this assumption is verified in Section \ref{sec:experiments-discussions}). However, it is worth noting that this assumption may not hold true in all circumstances, and in fact, $R(x)$ can be fine-tuned for each image to obtain an optimal result. In order to partially compensate for this limitation, a justification is presented in the following which explains the applicable range of $R(x)$.

By examining Fig. \ref{fig02:rsvt_examples}d, it can be observed that each clean point $c(x)$ is derived by shifting the corresponding hazy point $h(x)$ far away from the intersection $E$ (marked as a green point). That is, $h(x)$ tends to be closer to $E$ when compared with $c(x)$. Consequently, it can be concluded that $||h(x) - E|| \leq ||c(x) - E||$, indicating that the S-V ratio $r(x)$ is roughly less than 1:
\begin{equation}
r(x) \leq 1.   \label{equation_rx}
\end{equation}
On the other hand, $t_b(x)$ is used to indicate the transmission for the background regions. Typically, the minimum value for $t_b(x)$ is set to $0.1$ in most scenarios to represent the transmission of sky regions: $t_b(x) \geq 0.1$ \cite{he2010single}. However, in our situation, $t_b(x)$ not only represents the transmission for the sky but also for certain nearby bright objects. Therefore, the lower bound for $t_b(x)$ in our case can be adjusted slightly higher, i.e., 0.2: 
\begin{equation}
t_b(x) \geq 0.2 \text{ or }  1/t_b(x) \leq 5.    \label{equation_tbx}
\end{equation}
From Ep. (\ref{equation_rx}) and Ep. (\ref{equation_tbx}), we have: 
\begin{equation}
R(x) = r(x)/t_b(x) \leq 5.    \label{equation_rxtbx_constraint}
\end{equation}
This constraint is intriguingly justifiable with respect to the distribution shown in Fig. \ref{fig03:statistics-c}, where the value of $R(x)$ is generally below 5. According to this constraint and the distribution, we hypothesize that the potential values for $R(x)$ lie within the range of $[2.0,5.0]$ and we examine multiple configurations of $R(x)$ within this interval in Section \ref{sec:experiments-discussions}.


\begin{figure}
  \centering
  \includegraphics[width=1.0\linewidth]{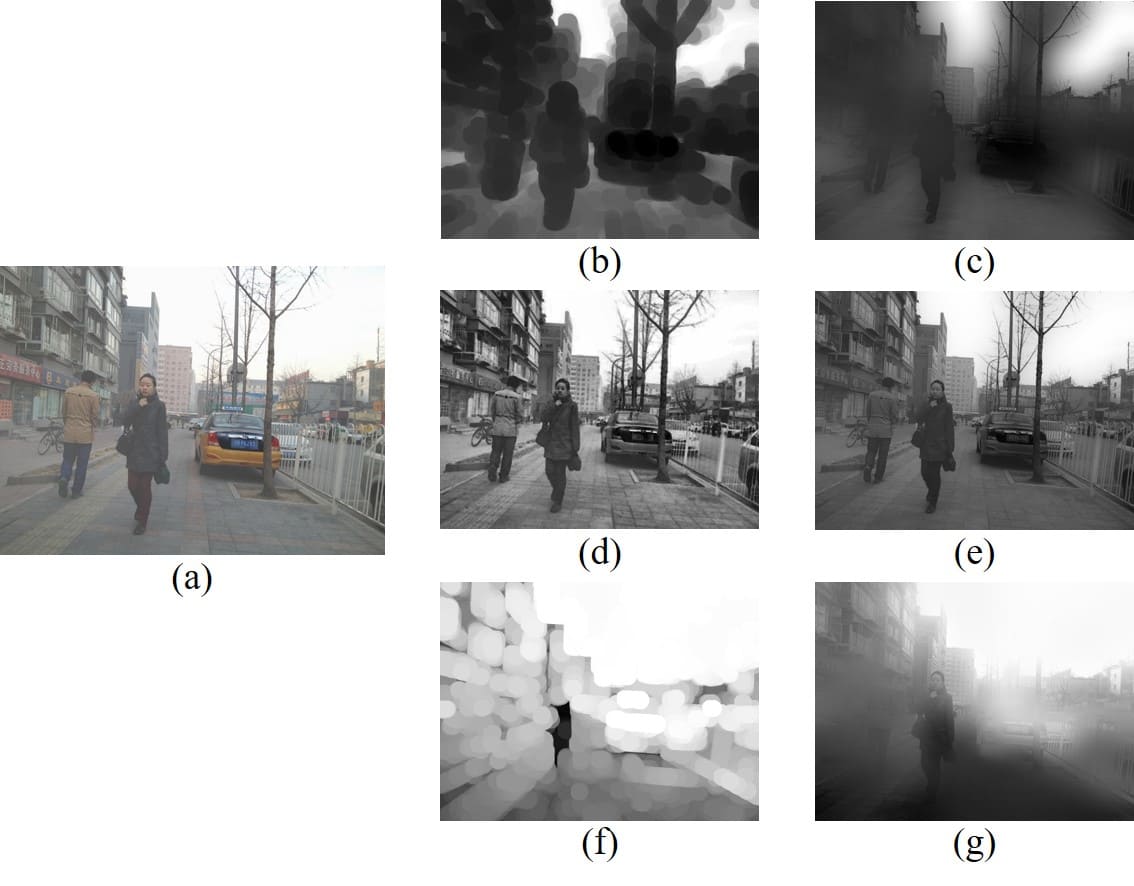} 
  \caption{Morphological min-max channel: (a) hazy image, (b) min channel with a large radius, (d) min channel with a small radius, (f) max channel with a large radius, and (c,e,g) refined images of (b,d,f), respectively.}
  \label{fig04:morphological-min-max-channels}
\end{figure}

\section{The Proposed Dehazing Method}
\label{sec:proposed-dehazing-method}

In this section, we first present a soft segmentation method utilizing a morphological min-max channel to effectively separate an input hazy image into three components: hard foreground, hard background, and middle ground. Subsequently, we propose a robust framework for single image dehazing that integrates the proposed RSVT prior with DCP.

\subsection{Morphological Min-Max Channel}
\label{subsec:hybrid-dark-channel}

In this section, a straightforward yet efficient soft segmentation strategy, called morphological min-max channel, which is inspired by dark/bright channel concepts \cite{he2010single, tao2017low} is proposed. Specifically, two channels, namely the foreground channel and the background channel, are extracted from a given hazy image. These channels are computed to emphasize the textures of their respective foreground/background regions. Subsequently, the two channels are combined to generate a unified mask that can effectually segment the input image.

Fig. \ref{fig04:morphological-min-max-channels}a shows an example of a hazy image that can be roughly segmented into two main regions: sky and non-sky. The non-sky region, which is considered the foreground, typically consists of various types of object textures and shows higher contrast compared to the sky region, which serves as the background. Consequently, the majority of non-sky regions should be connected with the top lowest-intensity pixels. Moreover, the foreground region usually contains floating bright pixels. To address this, a pixel-level min operation followed by a morphological erosion process is adopted to highlight the foreground. The purpose of the morphological erosion is to eliminate the floating bright pixels, thereby retaining only the significant foreground object textures. The computation of the coarse foreground channel $\tilde{I}_{fore}(x)$ for an input hazy image is as follows:
\begin{equation}
\tilde{I}_{fore}(x) =  \min_{c \in RGB} I_c(x) \ominus \Omega_f(x),
\end{equation}
where "$\ominus$" represents the morphological erosion operator while $\Omega_f(x)$ denotes a circle-shaped structuring element centered at $x$ for the erosion operation. Afterward the refined foreground channel $I_{fore}(x)$ is obtained by using guided filtering method:
\begin{equation}
I_{fore} =  GuidedFilter(\tilde{I}_{fore}, I_{gray}).
\end{equation}
In addition, to preserve the edge details and prevent the occurrence of shadow artifacts between the background and foreground areas, the radius of $\Omega_f(x)$ is typically set to a small value. In order to illustrate this effect of radius on the outcome, an example is given in Fig. \ref{fig04:morphological-min-max-channels}b and Fig. \ref{fig04:morphological-min-max-channels}d which show the coarse foreground channels with radius values of 30 and 3, respectively, while Fig. \ref{fig04:morphological-min-max-channels}c and Fig. \ref{fig04:morphological-min-max-channels}e depict the refined foreground channels of Fig. \ref{fig04:morphological-min-max-channels}b and Fig. \ref{fig04:morphological-min-max-channels}d, respectively. It can be observed clearly that a small radius can keep the details of objects while a large one can create shadow artifacts around the edges and result in a loss in object details.

\begin{figure}
  \centering
  \includegraphics[width=1.0\linewidth]{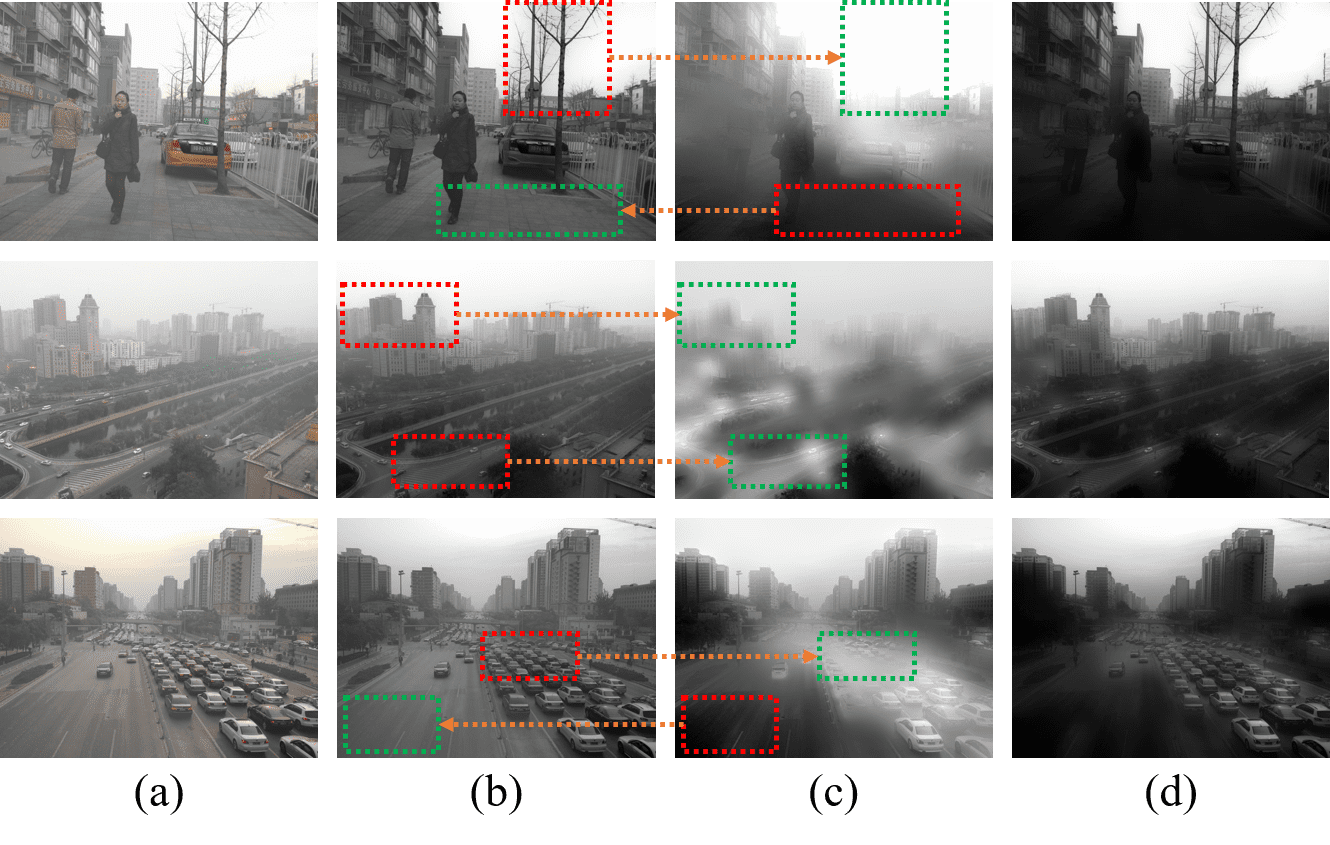} 
  \caption{Fused channel: (a) hazy image, (b) refined min channel, (c) refined max channel, and (d) fused channel of (b) and (c) through spatial-wise multiplication. The \textcolor{red}{red} and \textcolor{green}{green} boxes indicate oppositely low-intensity and high-intensity image patches from two channels at the same location, respectively.}
  \label{fig05:fused-channel}
\end{figure}

\begin{figure}
  \centering
  \includegraphics[width=1.0\linewidth]{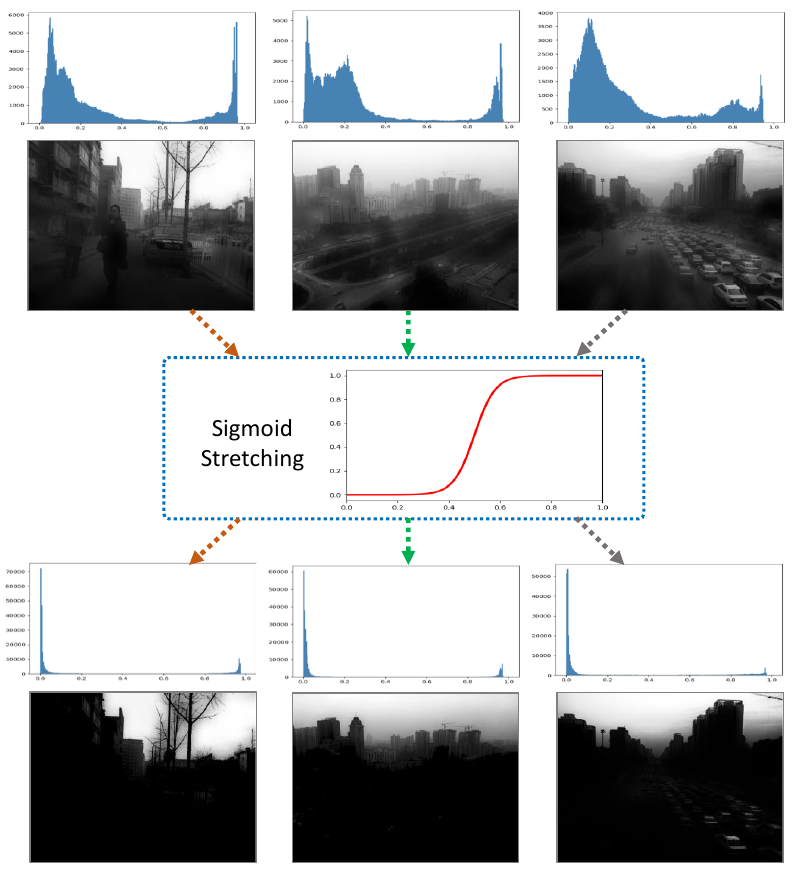} 
  \caption{Sigmoid stretching: the coarse soft segmentation masks and their respective histogram before (top) and after (bottom) stretching process.}
  \label{fig06:sigmoid-stretching}
\end{figure}

On the other hand, the majority of bright regions ought to be associated with the top highest-intensity pixels which can be obtained by adopting a pixel-wise max operation. Furthermore, these bright regions typically lack object textures and are spread out across a local region of an image. Hence, morphological dilation with a substantial radius can enhance the background characteristics in this context. As a result, the coarse background channel, denoted as $\tilde{I}_{back}(x)$, is accordingly computed as:
\begin{equation}
\tilde{I}_{back}(x) =  \max_{c \in RGB} I_c(x) \oplus \Omega_b(x),
\end{equation}
where "$\oplus$" denotes the morphological dilation operator while $\Omega_b(x)$ represents a circle-shaped structuring element centered at $x$ for the dilation operation. Then the refined background channel can be attained as:
\begin{equation}
I_{back} =  GuidedFilter(\tilde{I}_{back}, I_{gray}).
\end{equation}
Fig. \ref{fig04:morphological-min-max-channels}f and Fig. \ref{fig04:morphological-min-max-channels}g illustrate the coarse and refined background channels, respectively. As can be observed from Fig. \ref{fig04:morphological-min-max-channels}e, the refined foreground channel can capture fine details of object textures, but still contains some non-dark areas. Interestingly, those corresponding regions in the refined background channel, depicted in Fig. \ref{fig04:morphological-min-max-channels}g, can be leveraged in order to cover the uneven bright areas in the refined foreground channel. Therefore, these two channels are fused through spatial-wise multiplication, resulting in a coarse soft segmentation mask that roughly separates the sky and non-sky regions into bright and dark regions, respectively, as illustrated in Fig. \ref{fig05:fused-channel}. However, the intervals of the output values may be unsatisfactory due to the spatial-wise multiplication step. Therefore, a sigmoid model can be employed to stretch the values of the coarse mask:
\begin{equation}
I_{fuse} =  Sigmoid(I_{fore}I_{back}),  \label{equation8}
\end{equation}
where $I_{fuse}$ denotes the fusion outcome. Here the sigmoid model is adopted to stretch the foreground and background regions to have their weight values closer to 0 and 1, respectively. An illustration of this stretching process is presented in Fig. \ref{fig06:sigmoid-stretching}.

\subsection{Mask Refinement}
\label{subsec:refinement}

\begin{figure}
  \centering
  \includegraphics[width=1.0\linewidth]{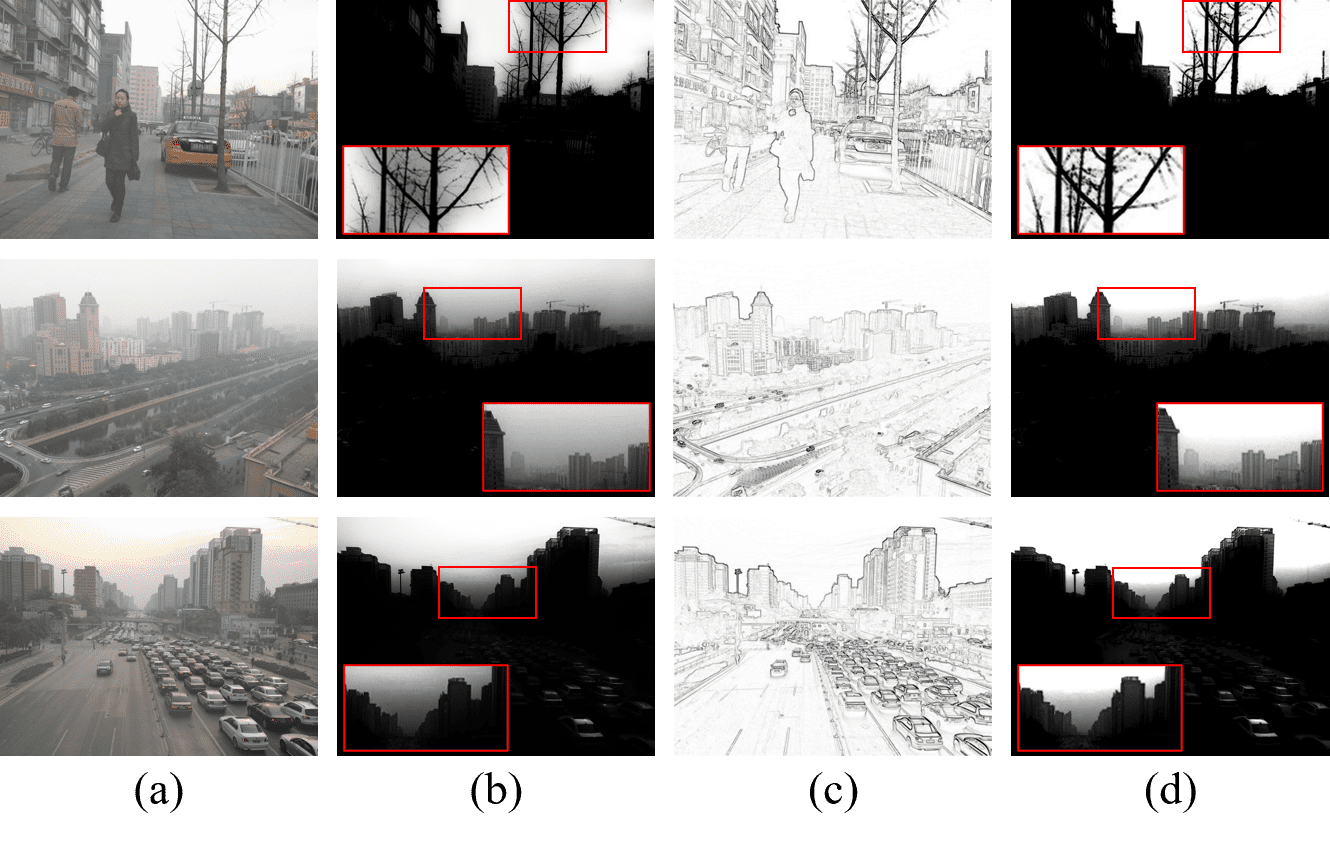} 
  \caption{Refinement of fused channel: (a) hazy image, (b) coarse fused channel, (c) inverse edge image, and (d) refined fused channel.}
  \label{fig07:mask_refinement}
\end{figure}

As can be seen from Fig. \ref{fig06:sigmoid-stretching}, the stretched mask can separate an input hazy image into two parts of background and foreground to a certain extent. However, it still contains residual shadow halos at some regions between background and foreground, as shown in red boxes in Fig. \ref{fig07:mask_refinement}b. To address this issue, a refinement process is carried out to attain an improved and refined mask. Typically, the background regions are spread out across a local area of the image with a low variation of pixel intensities. To take advantage of this property, the Sobel edge detection operator \cite{kanopoulos1988design} is adopted to compute the pixel gradients and extract the basic edge image of the input, followed by an image inverting step to bring the background intensities closer to 1:
\begin{equation}\label{eq:9s}
I_{edge} = 1 - Sobel(I_{gray}),
\end{equation}
where $I_{edge}$ denotes the obtained inverse edge image, as shown in Fig. \ref{fig07:mask_refinement}c.

On the other hand, from the stretched mask shown in Fig. \ref{fig07:mask_refinement}b, we observe that the regions having intensities closer to 0 have a higher probability of being in the foreground. Therefore, these regions with very low intensities can be considered \textit{hard foreground} with a constraint:
\begin{equation}\label{eq:9}
I_{fuse}(x) = 
\begin{cases} 
I_{fuse}(x), & \mbox{if } I_{fuse}(x) \geq \tau \\ 
0, & \mbox{if } I_{fuse}(x) < \tau 
\end{cases}
\end{equation}
where $\tau$ is a very low threshold and is typically set to 0.01. Subsequently, a spatial-wise multiplication is applied to the stretched mask and the inverse edge image. In so doing, the high intensity of the background in the inverse edge image can amplify the background region while the lower intensity of the stretched mask can down-weight the foreground region. This particular procedure can be referred to as a type of spatially varying detail enhancement \cite{li2014weighted} where the detail layer and the amplification factor are represented by the inverse edge image $I_{edge}$ and the stretched mask $I_{fuse}$, respectively. In practice, an outdoor input image may or may not contain sky regions, and the certainty of a pixel belonging to the foreground/background region can be considered random. Therefore, from the probability perspective, we assume that the stretched mask can be regarded as a random distribution, with element values ranging from 0 to 1 and a mean value $\phi = 0.5$.  Accordingly, the pixel-wise product of the stretched mask and the inverse edge image yields a darker image, as both input signals have values lower than 1, and the overall brightness of the detail layer could be degraded by $\phi$. To handle this, we adopt a compensation coefficient $\lambda$, with $\lambda = 1/\phi = 2$, to counterbalance the brightness degradation in the background. With the effect of $\lambda$, the outcome may contain values exceeding 1, which are then classified as \textit{hard background}. To obtain a refined and normalized soft segmentation mask, these pixel values are limited to a maximum of 1. The remaining pixels in the refined mask, which do not belong to either the hard foreground or hard background, are referred to as the \textit{middle ground}. Ultimately, the refined mask $M$ is obtained as:
\begin{equation}\label{eq:9ss}
M = min(\lambda I_{edge} I_{fuse}, 1).
\end{equation}

\subsection{Atmospheric Light Estimation}
\label{subsec:atmospheric-light-estimation}

In the DCP method, the estimation of atmospheric light is carried out by selecting the pixels with the highest intensity respective to the top 0.1\% brightest pixels in the dark channel. However, this strategy is not always valid and can lead to color shift in the final output. In order to overcome this limitation and achieve a more accurate selection of global atmospheric light, we propose referring to the top 0.1\% brightest pixels in the middle ground. This approach can avoid the selection of pixels that fall within excessively bright regions which may be associated with bright or white objects. The estimated atmospheric light is then computed as the average of these selected pixels.

\subsection{Regional Saturation-Value Translation}
\label{subsec:rsvt-prior}

As stated in Section \ref{subsec:3rd-assumption}, the haze-free point can be estimated by translating the hazy point far away from the atmospheric light by an amount directly proportional to its distance to the atmospheric light. To determine the precise amount of translation required for estimating the haze-free point, from Eq. (\ref{equation6}) we have:
\begin{equation}
\frac{||c(x) - A||}{||h(x) - A||} = \frac{1}{r(x)},   \label{equ25}
\end{equation}
note that the intersection $E$ is replaced with the estimated atmospheric light $A$ based on the second assumption presented in Section \ref{subsec:2nd-assumption}. Combining with Eq. (\ref{equ11}), Eq. (\ref{equ25}) is equivalent to:
\begin{equation}
\frac{||c(x) - h(x)||}{||h(x) - A||} + 1 = \frac{1}{R(x)t_s(x)},  \label{equation10}
\end{equation}
and the translation quantity $\delta(x)$ can be calculated as:
\begin{equation}
\delta(x) = ||c(x) - h(x)|| =  \frac{1 - R(x)t_s(x)}{R(x)t_s(x)}||h(x) - A||. \label{equation11}
\end{equation}
As a result, the haze-free image is restored by the following adjustments in the HSV color space:
\begin{equation}
H_c(x) = H_h(x), \label{equation11s}
\end{equation}
\begin{equation}\label{eq:9sc}
S_c(x) = 
\begin{cases} 
S_h(x) + \delta(x), & \mbox{if } S_h(x) \geq S_A \\ 
S_h(x) - \delta(x), & \mbox{if } S_h(x) < S_A 
\end{cases}
\end{equation}
\begin{equation}\label{eq:9scs}
V_c(x) = 
\begin{cases} 
V_h(x) + \delta(x), & \mbox{if } V_h(x) \geq V_A \\ 
V_h(x) - \delta(x), & \mbox{if } V_h(x) < V_A 
\end{cases}
\end{equation}
where $(H_c, S_c, V_c)$, $(H_h, S_h, V_h)$ denote three channels of the estimated HSV clean image and the HSV input hazy image, respectively.

\begin{algorithm}[t]
\caption{The proposed image dehazing framework.}\label{algo1}
\hspace*{0.0cm} \textbf{Input:} Hazy image $I(x)$, correlation value $R(x)$ \\
\hspace*{0.0cm} \textbf{Output:} Haze-free image $J(x)$
\begin{algorithmic}[1]
\State \textcolor{magenta}{\textit{\# Soft-segmentation Mask Extraction:}}
\State $I_{gray} = \text{rgb2gray}(I)$       \algorithmiccomment{RGB to gray}
\State $I_{edge} = 1 - \text{sobel}(I_{gray})$  \algorithmiccomment{inverse edge}
\State $\tilde{I}_{fore}(x) =  \min_{c \in RGB} I_c(x) \ominus \Omega_f(x)$
\State $I_{fore} =  \text{guidedfilter}(\tilde{I}_{fore}, I_{gray})$
\State $\tilde{I}_{back}(x) =  \max_{c \in RGB} I_c(x) \oplus \Omega_b(x)$
\State $I_{back} =  \text{guidedfilter}(\tilde{I}_{back}, I_{gray})$
\State $I_{fuse} =  \text{sigmoid}(I_{fore}I_{back})$    \algorithmiccomment{stretching}
\State $ I_{fuse}(x) = 
\begin{cases} 
I_{fuse}(x), & \mbox{if } I_{fuse}(x) \geq \tau \\ 
0, & \mbox{if } I_{fuse}(x) < \tau 
\end{cases}$
\State $M = \text{min}(\lambda I_{edge} I_{fuse}, 1)$  \algorithmiccomment{soft-seg mask}
\State
\State \textcolor{magenta}{\textit{\# Saturation-Value Translation:}}
\State $A = \text{average}(\text{top 0.1\%}( \forall M(x) | M(x) \neq 1 ))$
\State $\tilde{t_b}(x) =  1 - \omega \min_{y \in p_{3x3}(x)} (\min_{c \in (r,g,b)} \mathlarger{\frac{I_c(y)}{A_c}})$
\State $t_b(x) = \text{guidedfilter}(\tilde{t_b}(x), I_{gray})$
\State $h = \text{rgb2hsv}(I)$      \algorithmiccomment{RGB to HSV}
\State $\delta(x) = \mathlarger{\frac{1 - R(x)t_b(x)}{R(x)t_b(x)}}||h_{S,V}(x) - A_{S,V}||$
\State $H_c(x) = H_h(x)$
\State $S_c(x) = 
\begin{cases} 
S_h(x) + \delta(x), & \mbox{if } S_h(x) \geq S_A \\ 
S_h(x) - \delta(x), & \mbox{if } S_h(x) < S_A 
\end{cases}$
\State $V_c(x) = 
\begin{cases} 
V_h(x) + \delta(x), & \mbox{if } V_h(x) \geq V_A \\ 
V_h(x) - \delta(x), & \mbox{if } V_h(x) < V_A 
\end{cases}$
\State $D = \text{hsv2rgb}(\{H_c,S_c,V_c\})$    \algorithmiccomment{HSV to RGB}
\State
\State \textcolor{magenta}{\textit{\# Restoration:}}
\State $\tilde{t}(x) =  1 - \omega \min_{y \in p_{15x15}(x)} (\min_{c \in (r,g,b)}  \mathlarger{\frac{I_c(y)}{A_c}})$ 
\State $t(x) = \text{guidedfilter}(\tilde{t}(x), I_{gray})$
\State $J(x) = \mathlarger{\frac{I(x) - A(1 - t(x))}{max(t_0, t(x))}}(1-M(x)) + D(x)M(x)$
\end{algorithmic}
\end{algorithm}

\subsection{Haze-free Image Recovery}
\label{subsec:haze-free-recovery}

After acquiring the segmentation mask, the proposed RSVT prior and the DCP approach process the hard background and hard foreground regions, respectively. Meanwhile, the restoration of the middle ground is achieved through a weighted average solution derived from these two algorithms. The proposed dehazing scheme can be expressed as a per-pixel convex combination as follows:
\begin{equation}
J(x) = \frac{I(x) - A(1 - t(x))}{max(t_0, t(x))}(1-M(x)) + D(x)M(x), \label{equation12}
\end{equation} 
where $t(x)$ is estimated by DCP and $t_0$ (typically set to 0.1) is a lower bound of $t(x)$ \cite{he2010single}, while $D(x)$ is the dehazed image produced by RSVT formatted in the RGB color space. The Python-like pseudocode of the proposed dehazing algorithm is illustrated in Algorithm \ref{algo1}.

\section{Experiments and Discussions}
\label{sec:experiments-discussions}

In this section, the effectiveness of the proposed framework is examined in various experimental analyses. We first present the experimental settings including the datasets used in experiments as well as listing the methods utilized for comparisons. Subsequently, the performance of scene segmentation stage and the selection of $R(x)$ are discussed. Lastly, comparisons of the proposed dehazing method and other prevailing approaches in terms of dehazing effectiveness and processing time are provided.

\begin{figure*}
  \centering  
  \includegraphics[width=0.85\linewidth]{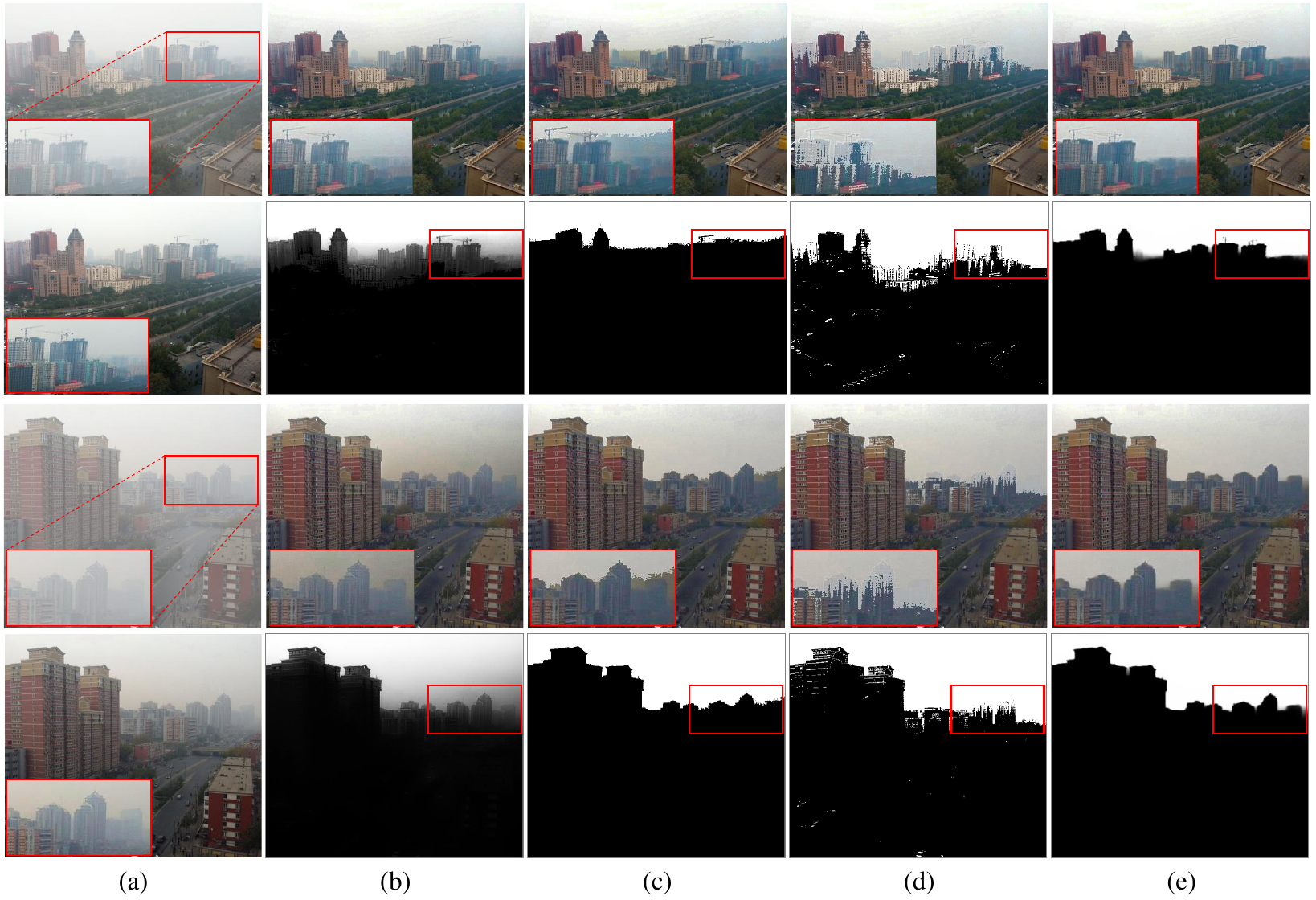}
  \caption{Dehazing results when RSVT is combined with various segmentation methods: (a) hazy (top) and clean (bottom) images, (b) the proposed morphological min-max channel method, (c) region growing-based method adopted in \cite{wang2017dehazing, salazar2020fast}, (d) GMM-based method used in \cite{yu2016image}, and (e) deep learning-based method using U\textsuperscript{2}-Net \cite{qin2020u2}. The respective segmentation mask is shown below each output.}
  \label{fig08:softseg-masks}
\end{figure*}

\begin{figure}
  \centering
  \includegraphics[width=1.0\linewidth]{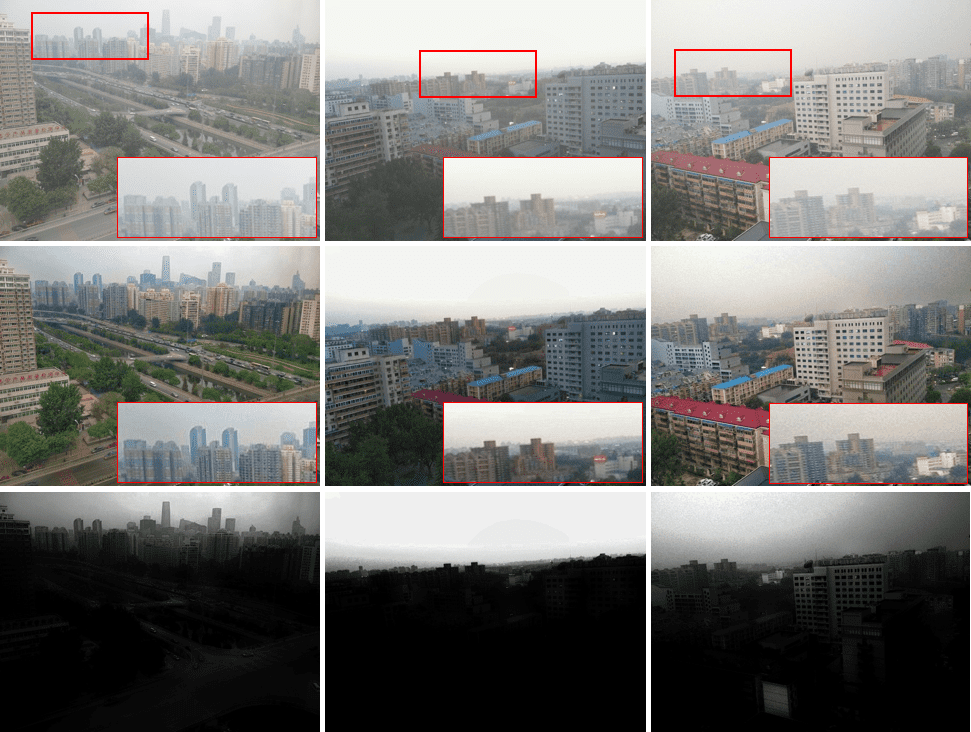} 
  \caption{Dehazing results by the proposed method (top: input image, middle: restored image, bottom: soft segmentation mask).}
  \label{fig09:visual-results}
\end{figure}

\begin{figure}
  \centering  
  \resizebox{0.50\textwidth}{!}{
  \begin{subfigure}{0.25\linewidth}
    \includegraphics[width=1.0\linewidth]{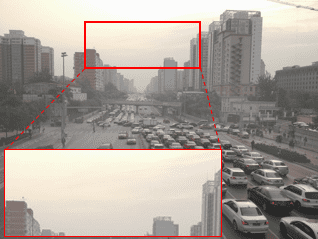}   

    \vspace{0.08cm}
    \includegraphics[width=1.01\linewidth]{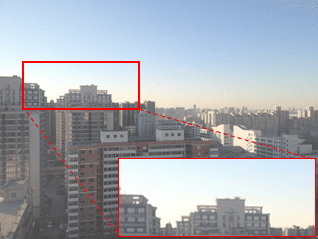}  
    \caption{Hazy}
    \label{fig10:Rx-a}
  \end{subfigure}
  \begin{subfigure}{0.25\linewidth}
    \includegraphics[width=1.0\linewidth]{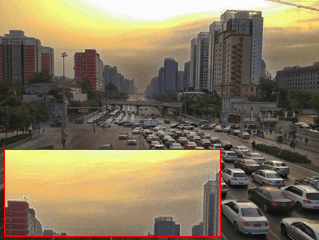}     

    \vspace{0.08cm}
    \includegraphics[width=1.01\linewidth]{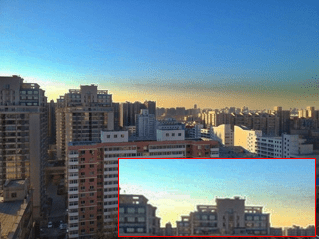}  
    \caption{$R(x)=1$}
    \label{fig10:Rx-b}
  \end{subfigure}
  \begin{subfigure}{0.25\linewidth}
    \includegraphics[width=1.0\linewidth]{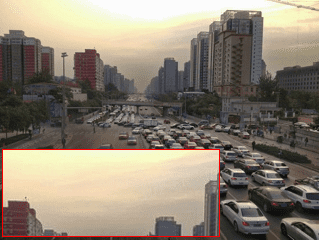}    

    \vspace{0.08cm}
    \includegraphics[width=1.01\linewidth]{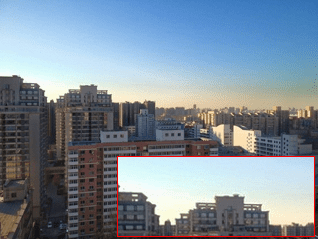}  
    \caption{$R(x)=2$}
    \label{fig10:Rx-c}
  \end{subfigure}
  \begin{subfigure}{0.25\linewidth}
    \includegraphics[width=1.0\linewidth]{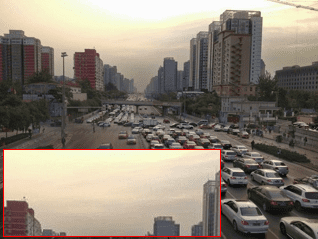}

    \vspace{0.08cm}
    \includegraphics[width=1.01\linewidth]{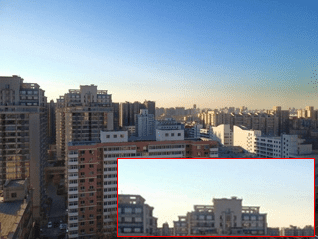}  
    \caption{$R(x)=3$}
    \label{fig10:Rx-d}
  \end{subfigure}
  }

  \vspace{0.2cm}
  
  \resizebox{0.50\textwidth}{!}{
  \begin{subfigure}{0.25\linewidth}
    \includegraphics[width=1.0\linewidth]{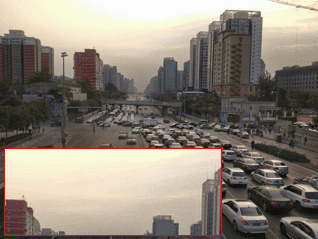}

    \vspace{0.08cm}
    \includegraphics[width=1.01\linewidth]{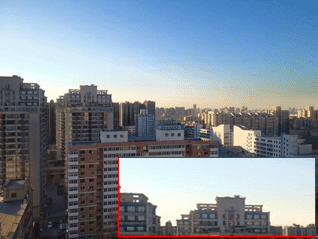}  
    \caption{$R(x)=4$}
    \label{fig10:Rx-e}
  \end{subfigure}
  \begin{subfigure}{0.25\linewidth}
    \includegraphics[width=1.0\linewidth]{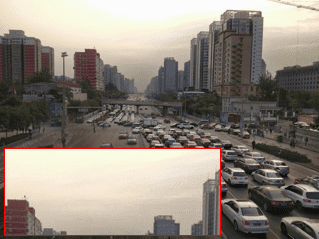}     

    \vspace{0.08cm}
    \includegraphics[width=1.01\linewidth]{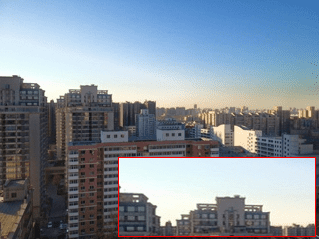}  
    \caption{$R(x)=5$}
    \label{fig10:Rx-f}
  \end{subfigure}
  \begin{subfigure}{0.25\linewidth}
    \includegraphics[width=1.0\linewidth]{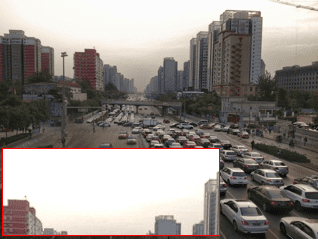}    

    \vspace{0.08cm}
    \includegraphics[width=1.01\linewidth]{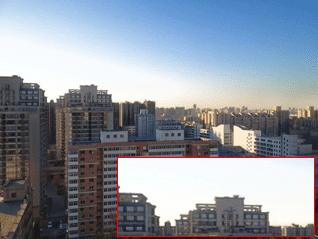}  
    \caption{$R(x)=10$}
    \label{fig10:Rx-g}
  \end{subfigure}
  \begin{subfigure}{0.25\linewidth}
    \includegraphics[width=1.0\linewidth]{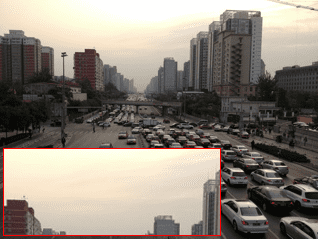}    

    \vspace{0.08cm}
    \includegraphics[width=1.01\linewidth]{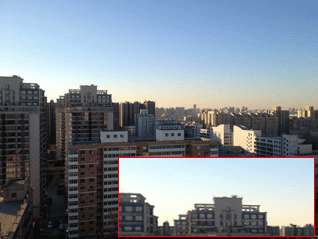}  
    \caption{Clean}
    \label{fig10:Rx-h}
  \end{subfigure}
  }
  \caption{Dehazing results with various configurations of $R(x)$.}
  \label{fig10:Rx}
\end{figure}

\begin{figure*}
  \centering  
  \resizebox{0.98\textwidth}{!}{
  \begin{subfigure}{0.120\linewidth}
    \includegraphics[width=1.01\linewidth]{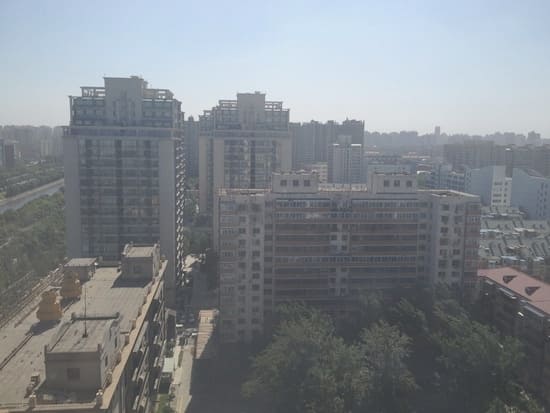}
    
    \vspace{0.08cm}
    \includegraphics[width=1.01\linewidth]{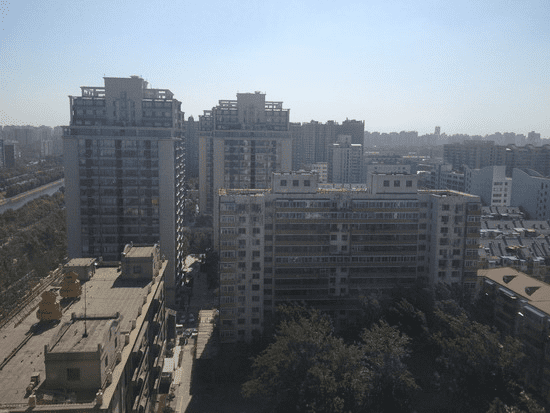}  

    \vspace{0.15cm}
    \includegraphics[width=1.01\linewidth]{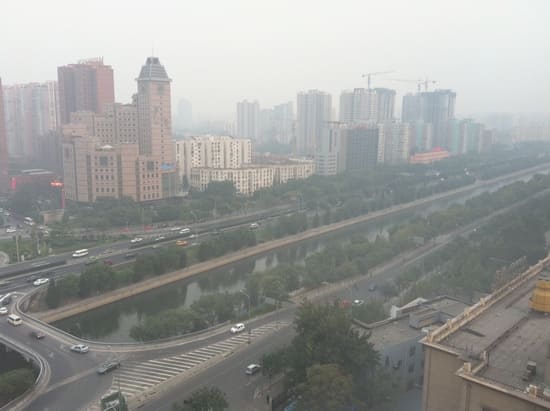}
    
    \vspace{0.08cm}
    \includegraphics[width=1.01\linewidth]{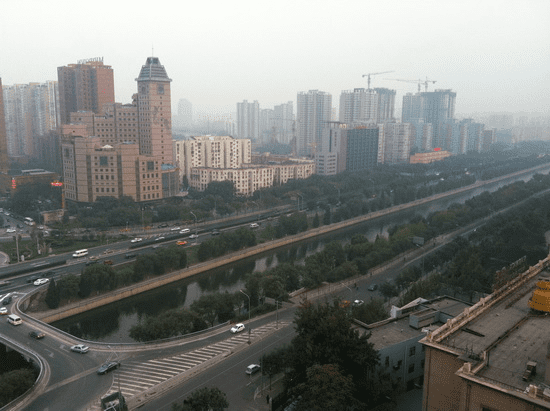}
    \caption{Hazy/clean}
    \label{fig11:SOTS-Outdoor-a}
  \end{subfigure}
  \hfill
  \begin{subfigure}{0.120\linewidth} 
    \includegraphics[width=1.01\linewidth]{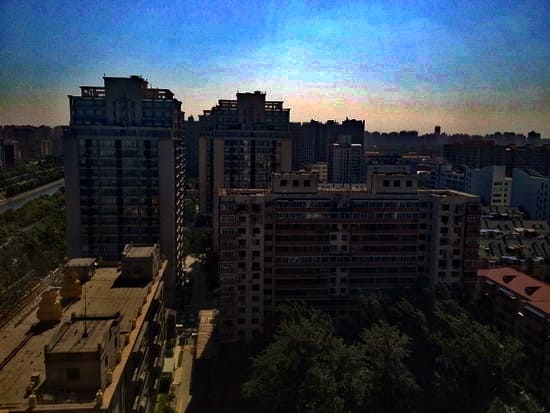}  

    \vspace{0.08cm}
    \includegraphics[width=1.01\linewidth]{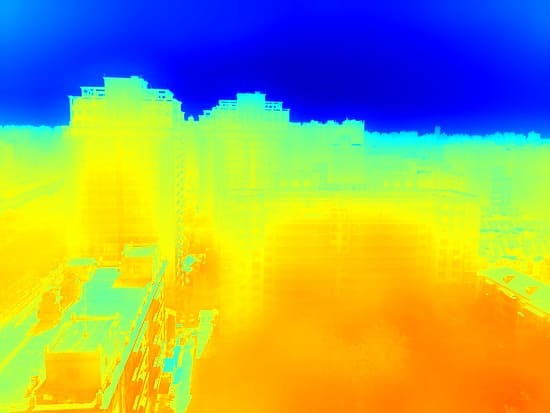}   

    \vspace{0.15cm}
    \includegraphics[width=1.01\linewidth]{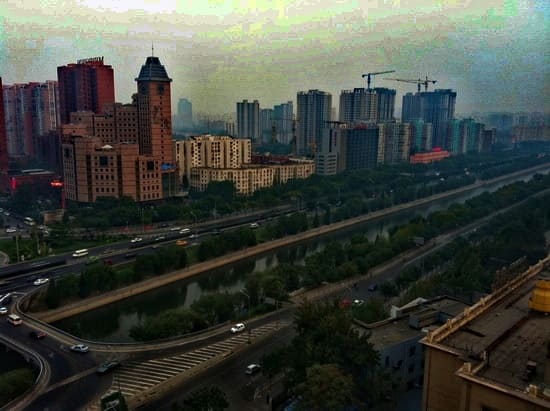}
    
    \vspace{0.08cm}
    \includegraphics[width=1.01\linewidth]{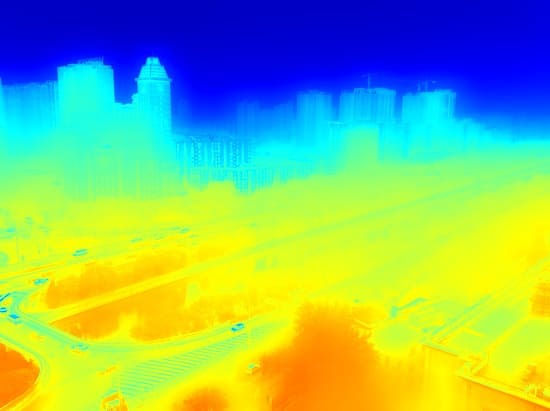}
    \caption{CEP}
    \label{fig11:SOTS-Outdoor-b}
  \end{subfigure}
  \hfill
  \begin{subfigure}{0.120\linewidth}
    \includegraphics[width=1.01\linewidth]{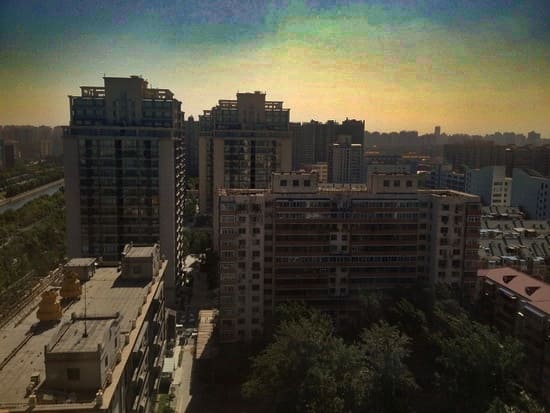}  

    \vspace{0.08cm}
    \includegraphics[width=1.01\linewidth]{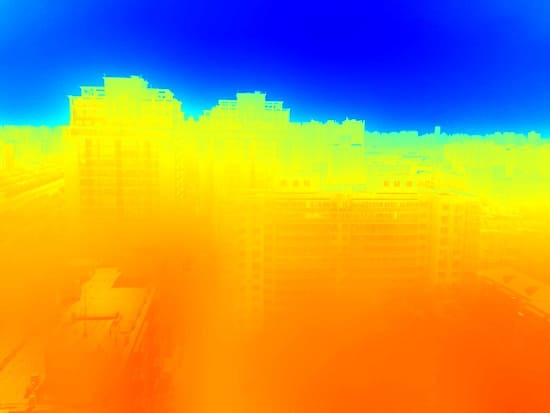}    

    \vspace{0.15cm}
    \includegraphics[width=1.01\linewidth]{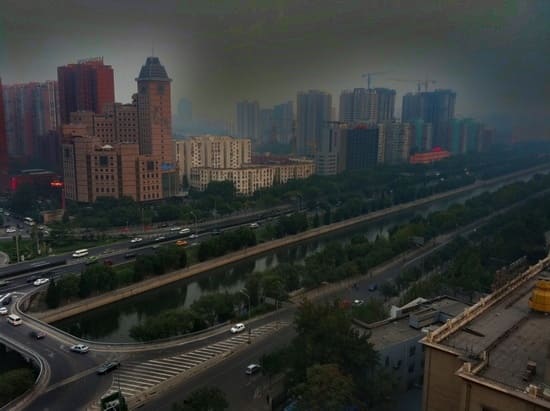}
    
    \vspace{0.08cm}
    \includegraphics[width=1.01\linewidth]{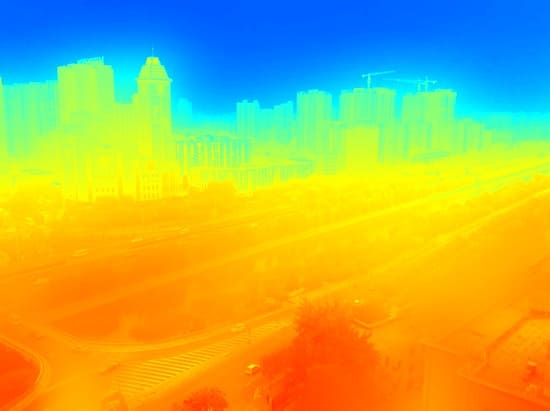}
    \caption{DCP}
    \label{fig11:SOTS-Outdoor-c}
  \end{subfigure}
  \hfill
  \begin{subfigure}{0.120\linewidth}
    \includegraphics[width=1.01\linewidth]{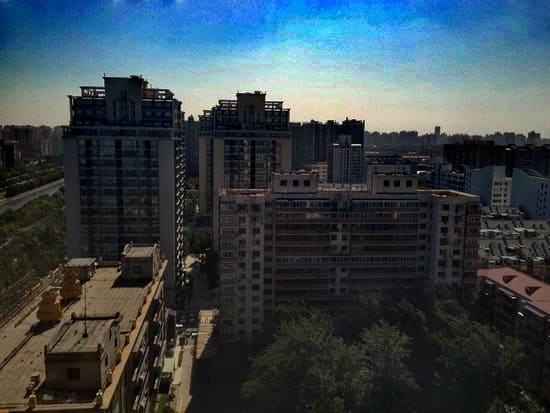}  

    \vspace{0.08cm}
    \includegraphics[width=1.01\linewidth]{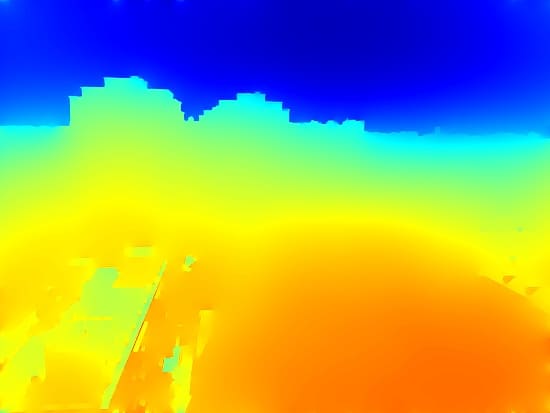}    

    \vspace{0.15cm}
    \includegraphics[width=1.01\linewidth]{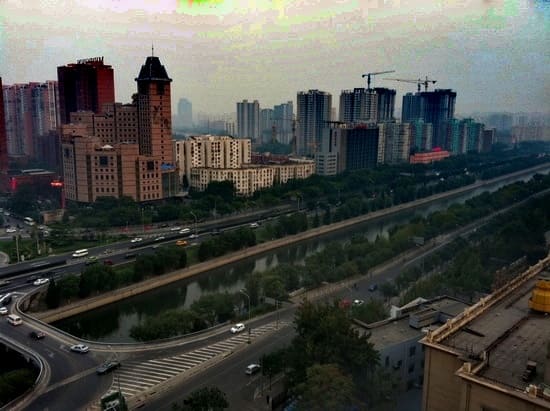}
    
    \vspace{0.08cm}
    \includegraphics[width=1.01\linewidth]{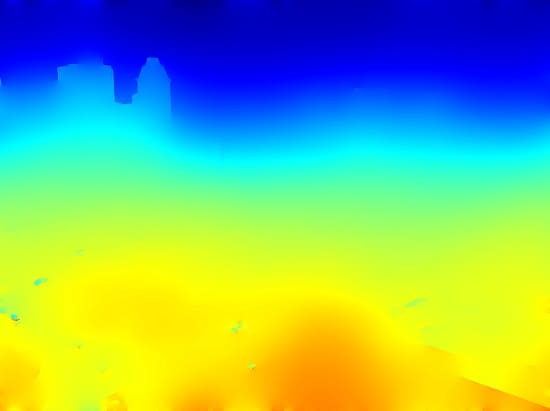}
    \caption{BCCR}
    \label{fig11:SOTS-Outdoor-d}
  \end{subfigure}
  \hfill
  \begin{subfigure}{0.120\linewidth}
    \includegraphics[width=1.01\linewidth]{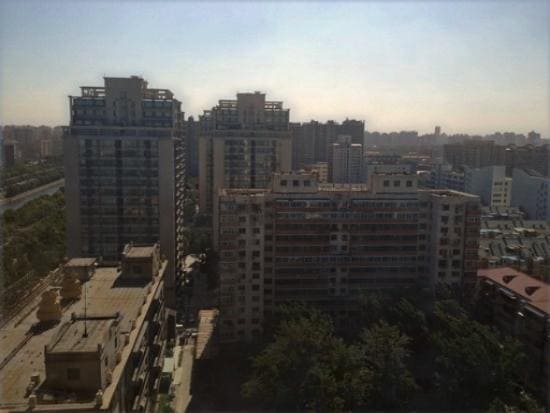}  

    \vspace{0.08cm}
    \includegraphics[width=1.01\linewidth]{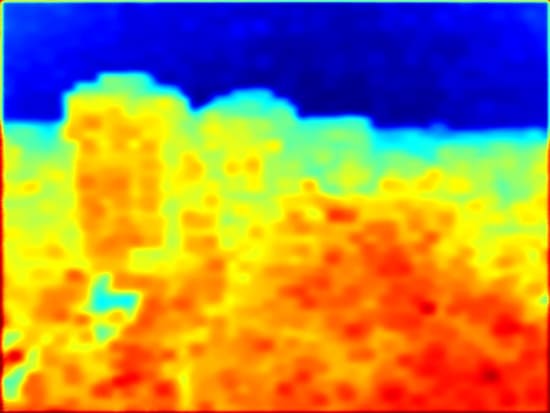}    

    \vspace{0.15cm}
    \includegraphics[width=1.01\linewidth]{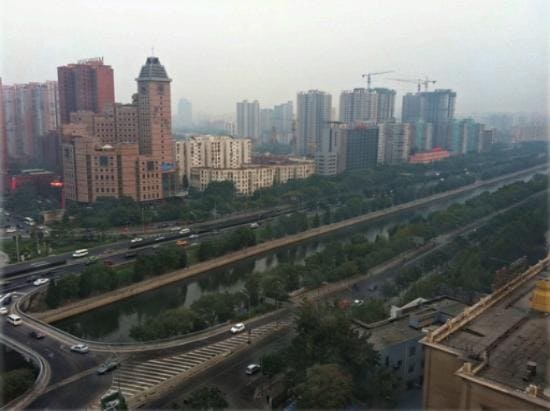}
    
    \vspace{0.08cm}
    \includegraphics[width=1.01\linewidth]{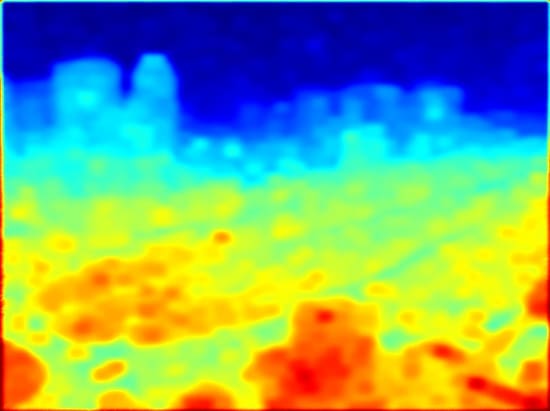}
    \caption{CAP}
    \label{fig11:SOTS-Outdoor-e}
  \end{subfigure}
  \hfill
  \begin{subfigure}{0.120\linewidth}
    \includegraphics[width=1.01\linewidth]{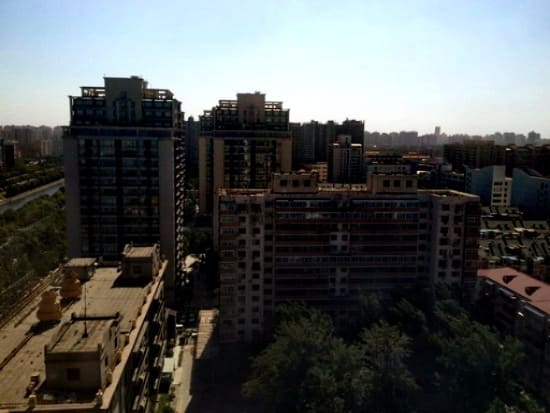}  

    \vspace{0.08cm}
    \includegraphics[width=1.01\linewidth]{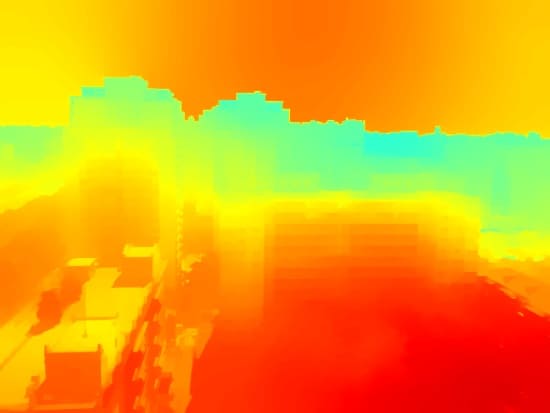}   

    \vspace{0.15cm}
    \includegraphics[width=1.01\linewidth]{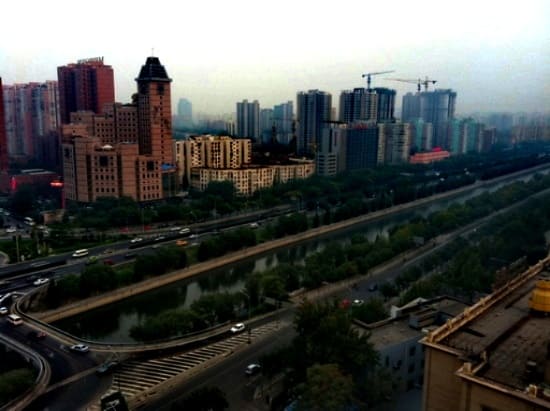}
    
    \vspace{0.08cm}
    \includegraphics[width=1.01\linewidth]{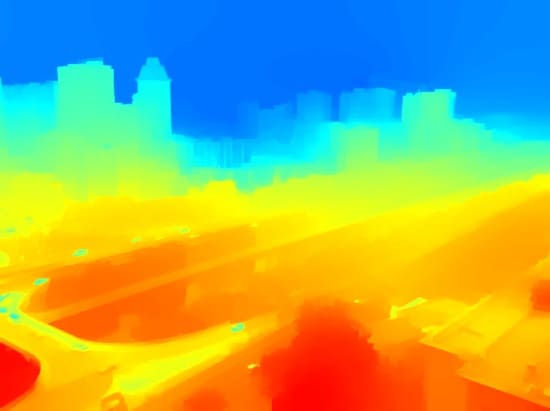} 
    \caption{NLID}
    \label{fig11:SOTS-Outdoor-f}
  \end{subfigure}
  \hfill
  \begin{subfigure}{0.120\linewidth}
    \includegraphics[width=1.01\linewidth]{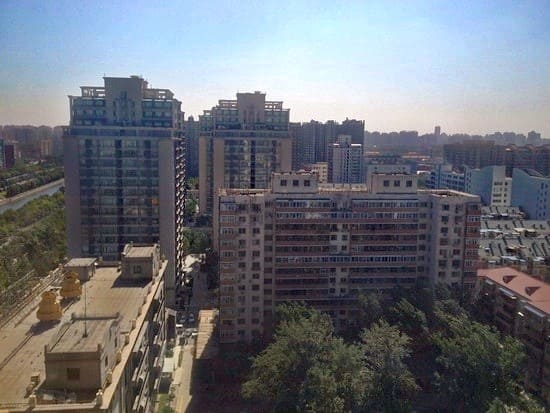}  

    \vspace{0.08cm}
    \includegraphics[width=1.01\linewidth]{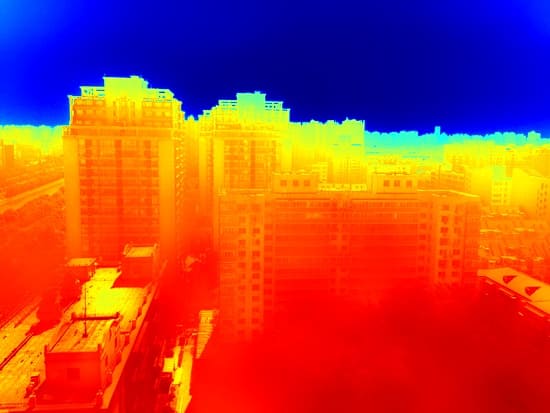}    

    \vspace{0.15cm}
    \includegraphics[width=1.01\linewidth]{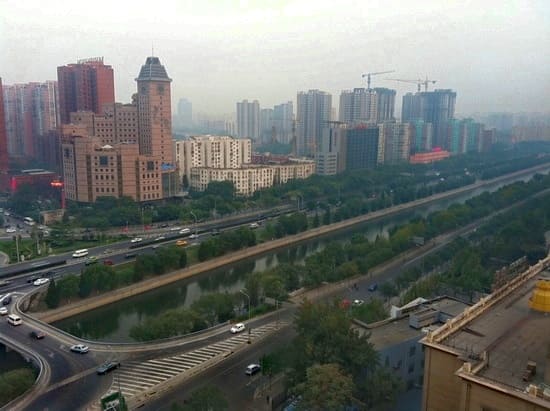}
    
    \vspace{0.08cm}
    \includegraphics[width=1.01\linewidth]{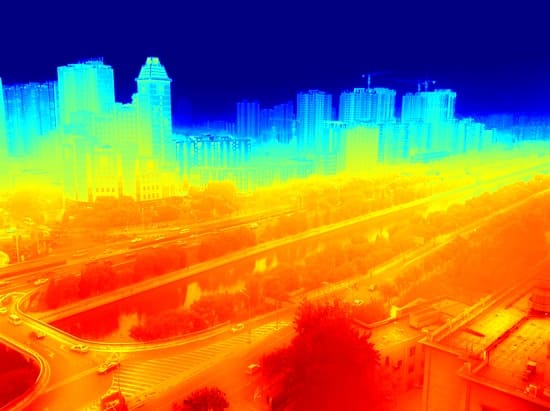}
    \caption{RSVT (ours)}
    \label{fig11:SOTS-Outdoor-g}
  \end{subfigure}
  }
  \caption{Typical dehazing results of various approaches on SOTS-Outdoor dataset: (a) hazy (top) and clean (bottom) images, (b-g) results by CEP \cite{bui2017single}, DCP \cite{he2010single}, BCCR \cite{meng2013efficient}, CAP \cite{zhu2015fast}, NLID \cite{berman2016non}, and the proposed RSVT method, respectively. The corresponding recovered transmission map for each method is shown below every output.}
  \label{fig11:SOTS-Outdoor}
\end{figure*}

\subsection{Experimental Settings}
\label{subsec:settings}

The proposed method has been compared with various prevailing prior-based dehazing approaches such as CEP \cite{bui2017single}, DCP \cite{he2010single}, CAP \cite{zhu2015fast}, NLID \cite{berman2016non}, CLAHE \cite{pizer1987adaptive}, and BCCR \cite{meng2013efficient} as well as several existing deep learning-based methods including AOD-Net \cite{li2017aod}, MSCNN \cite{ren2016single}, DehazeNet \cite{cai2016dehazenet}, GFN \cite{ren2018gated}, GCANet \cite{chen2019gated}, and YOLY \cite{li2021you}. The comparisons were conducted on four data sets including: two widely used synthetic datasets, SOTS-Outdoor and HSTS-Synthetic, and two realistic hazy image sets, HSTS-Realistic and RESIDE-Natural; all these data sets are part of the RESIDE benchmark \cite{li2018benchmarking}. The SOTS-Outdoor and HSTS-Synthetic datasets consist of 500 pairs and 10 pairs of synthetic hazy-clean outdoor images, respectively. On the other hand, the HSTS-Realistic dataset includes 10 real-world hazy images, while the RESIDE-Natural image set contains 20 randomly sampled images with natural haze from the RESIDE database \cite{li2018benchmarking}. The quantitative performances on paired image data were measured using Peak Signal-to-Noise Ratio (PSNR) and Structural Similarity Index Measure (SSIM), while the dehazing performances on natural images were assessed using Naturalness Image Quality Evaluator (NIQE) \cite{mittal2012making} and Blind/Referenceless Image Spatial Quality Evaluator (BRISQUE) \cite{mittal2012no}. The experiments were conducted on an Intel(R) Core(TM) i5-8600K CPU @ 3.60GHz.

\begin{table}
  \caption{Ablation study on the values of $R(x)$. The best and second-best results are indicated in \textbf{bold} and \textcolor{blue}{blue}, respectively.}
  \centering
  \resizebox{0.492\textwidth}{!}{
  \begin{tabular}{cccccccc}
    \toprule
    \multirow{2}{*}{Dataset} & \multirow{2}{*}{Metric} &  \multicolumn{6}{c}{$R(x)$}  \\
    \cmidrule{3-8}
     & & 1 & 2 & 3 & 4 & 5 & 10 \\
    \midrule
    SOTS & PSNR & 17.68 & 21.09 & \textbf{22.31} & \textcolor{blue}{21.98} & 21.06 & 18.35 \\
    Outdoor & SSIM & 0.7894 & \textcolor{blue}{0.8835} & \textbf{0.9031} & 0.8668 & 0.8691 & 0.8115 \\
    \midrule
    HSTS & PSNR & 16.59 & \textcolor{blue}{21.22} & \textbf{21.53} & 21.06 & 20.45 & 16.82 \\
    Synthetic & SSIM & 0.7740 & \textcolor{blue}{0.8798} & \textbf{0.8868} & 0.8785 & 0.8325 & 0.7853 \\
    \bottomrule
  \end{tabular}
  }
  \label{tab01:Rx}
\end{table}

\begin{table}
  \centering
  \caption{Quantitative performances of various dehazing methods on SOTS-Outdoor and HSTS-Synthetic datasets. The best and second-best results are indicated in \textbf{bold} and \textcolor{blue}{blue}, respectively.}
  \resizebox{0.455\textwidth}{!}{
  \begin{tabular}{cccccc}
    \toprule
    \multirow{2}{*}{Type} & \multirow{2}{*}{Method} & \multicolumn{2}{c}{SOTS-Outdoor} & \multicolumn{2}{c}{HSTS-Synthetic} \\
    \cmidrule{3-6}
           & & PSNR & SSIM & PSNR & SSIM \\
    \midrule
    \multirow{6}{*}{Prior-based} & CEP \cite{bui2017single} & 13.44 & 0.7053 & 14.02 & 0.7270 \\
    & DCP \cite{he2010single} & 16.62 & 0.8179 & 17.01 & 0.8030 \\
    & CLAHE \cite{pizer1987adaptive} & 16.38 & 0.8072 & 15.55 & 0.7886 \\
    & BCCR \cite{meng2013efficient} & 16.88 & 0.7913 & 15.21 & 0.7470 \\
    & NLID \cite{berman2016non} & 17.29 & 0.7489 & 17.62 & 0.7980 \\
    & CAP \cite{zhu2015fast} & 19.05 & 0.8364 & 21.57 & 0.8730 \\
    \midrule
    \multirow{6}{*}{CNN-based} & AOD-Net \cite{li2017aod} & 20.29 & \textcolor{blue}{0.8770} & 20.55 & \textcolor{blue}{0.8970} \\
    & MSCNN \cite{ren2016single} & 19.48 & 0.8390 & 18.64 & 0.8170 \\
    & DehazeNet \cite{cai2016dehazenet} & \textbf{22.46} & 0.8510 & \textbf{24.48} & \textbf{0.9150} \\
    & GFN \cite{ren2018gated} & 21.55 & 0.8440 & \textcolor{blue}{22.06} & 0.8470 \\
    & GCANet \cite{chen2019gated} & 21.66 & 0.8670 & 21.37 & 0.8740 \\
    & YOLY \cite{li2021you} & 20.39 & 0.8890 & 21.02 & 0.9050 \\
    \midrule
    Prior-based & RSVT (ours) & \textcolor{blue}{22.31} & \textbf{0.9031} & 21.53 & 0.8868 \\
    \bottomrule
  \end{tabular}}
  \label{tab02:synthetic-results}
\end{table}

\begin{figure*}
  \centering  
  \resizebox{0.98\textwidth}{!}{
  \begin{subfigure}{0.120\linewidth}
    \includegraphics[width=1.01\linewidth]{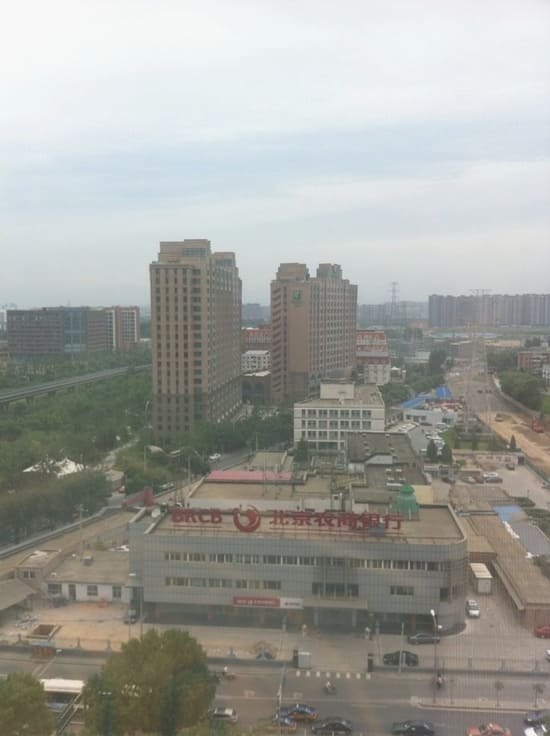}
    
    \vspace{0.08cm}
    \includegraphics[width=1.01\linewidth]{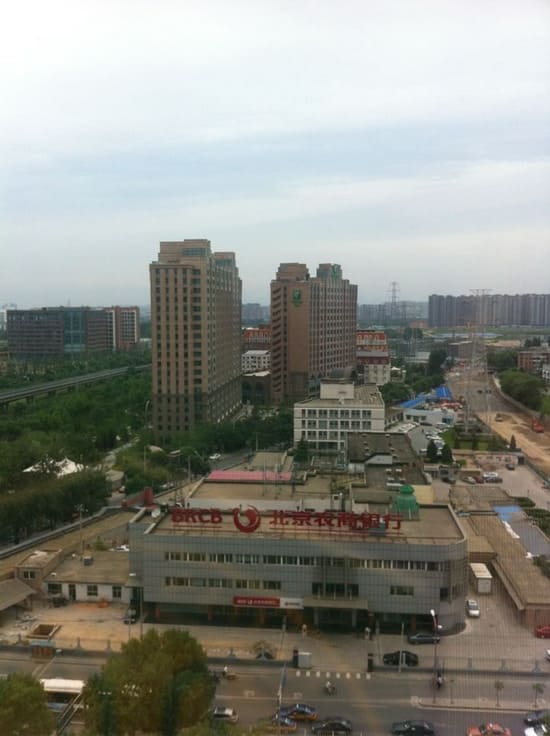}  

    \vspace{0.15cm}
    \includegraphics[width=1.01\linewidth]{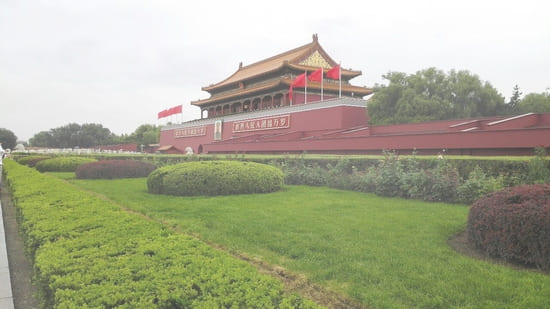}
    
    \vspace{0.08cm}
    \includegraphics[width=1.01\linewidth]{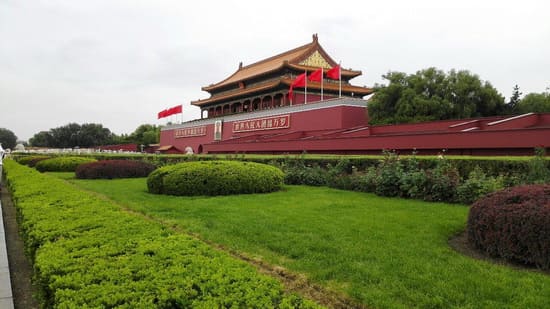}
    \caption{Hazy/clean}
    \label{fig12:HSTS-Synthetic-a}
  \end{subfigure}
  \hfill
  \begin{subfigure}{0.120\linewidth} 
    \includegraphics[width=1.01\linewidth]{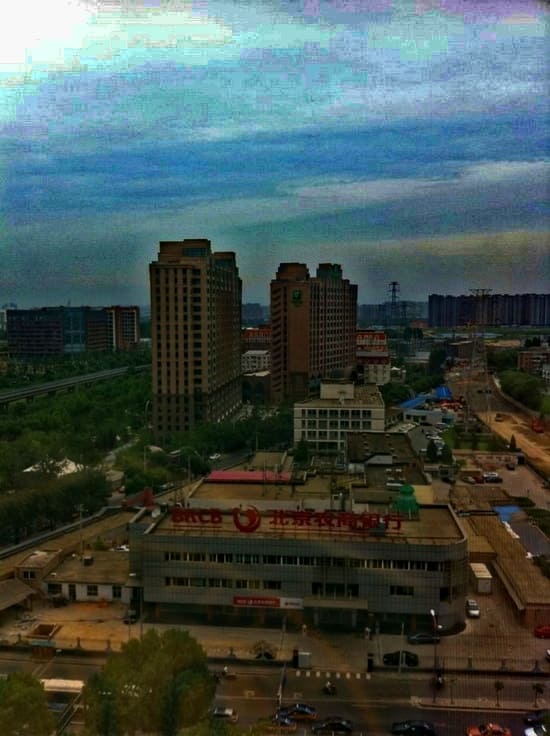}  

    \vspace{0.08cm}
    \includegraphics[width=1.01\linewidth]{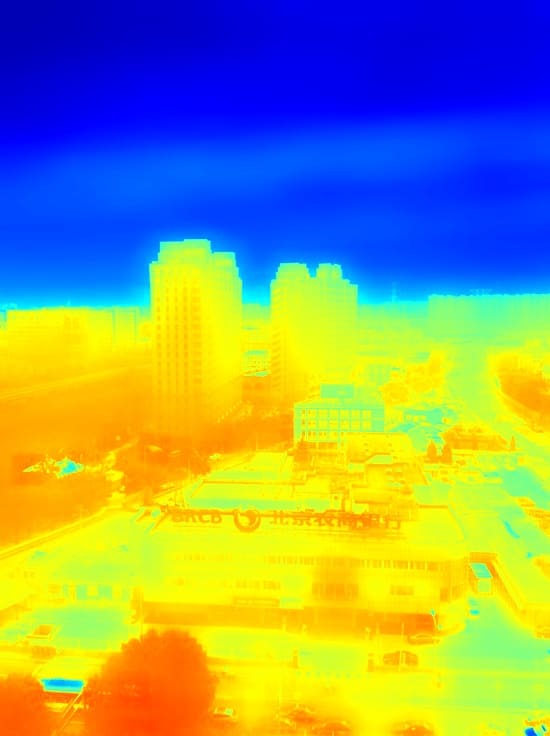}   

    \vspace{0.15cm}
    \includegraphics[width=1.01\linewidth]{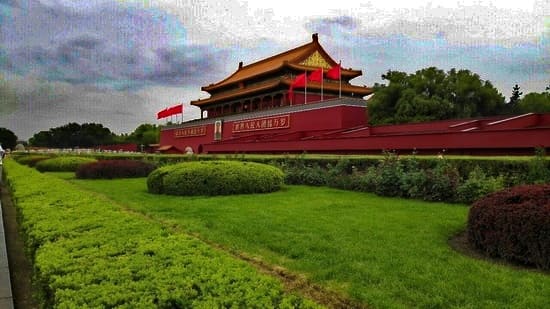}
    
    \vspace{0.08cm}
    \includegraphics[width=1.01\linewidth]{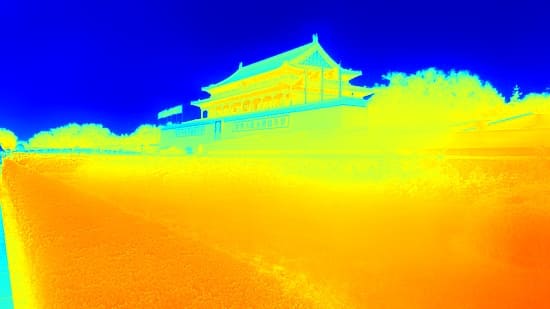}
    \caption{CEP}
    \label{fig12:HSTS-Synthetic-b}
  \end{subfigure}
  \hfill
  \begin{subfigure}{0.120\linewidth}
    \includegraphics[width=1.01\linewidth]{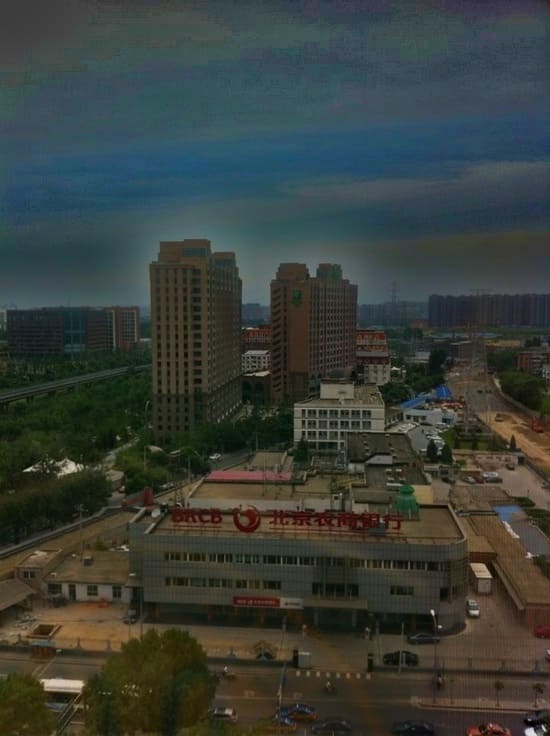}  

    \vspace{0.08cm}
    \includegraphics[width=1.01\linewidth]{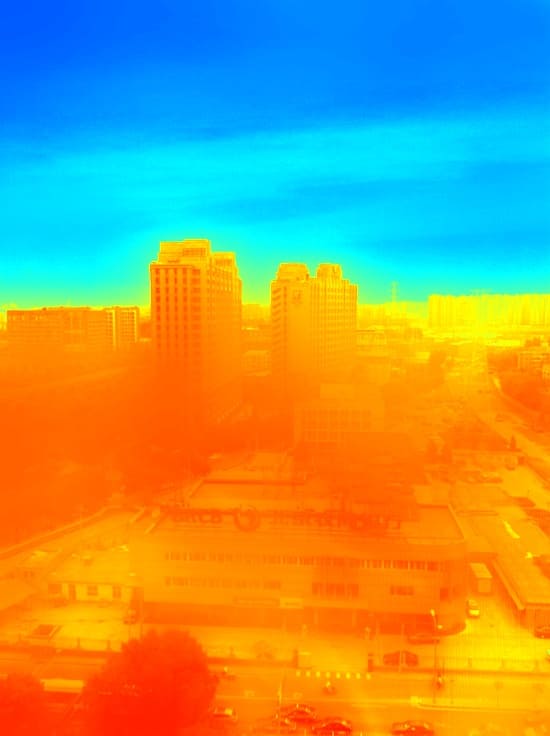}    

    \vspace{0.15cm}
    \includegraphics[width=1.01\linewidth]{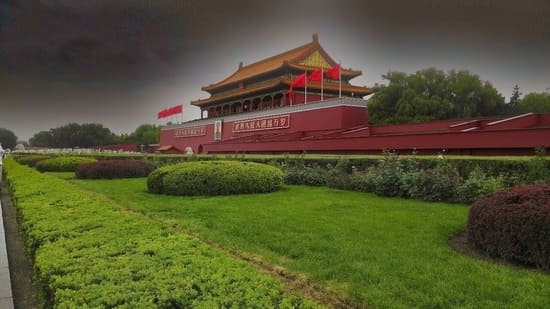}
    
    \vspace{0.08cm}
    \includegraphics[width=1.01\linewidth]{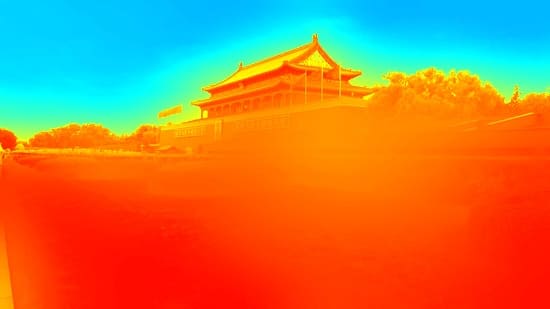}
    \caption{DCP}
    \label{fig12:HSTS-Synthetic-c}
  \end{subfigure}
  \hfill
  \begin{subfigure}{0.120\linewidth}
    \includegraphics[width=1.01\linewidth]{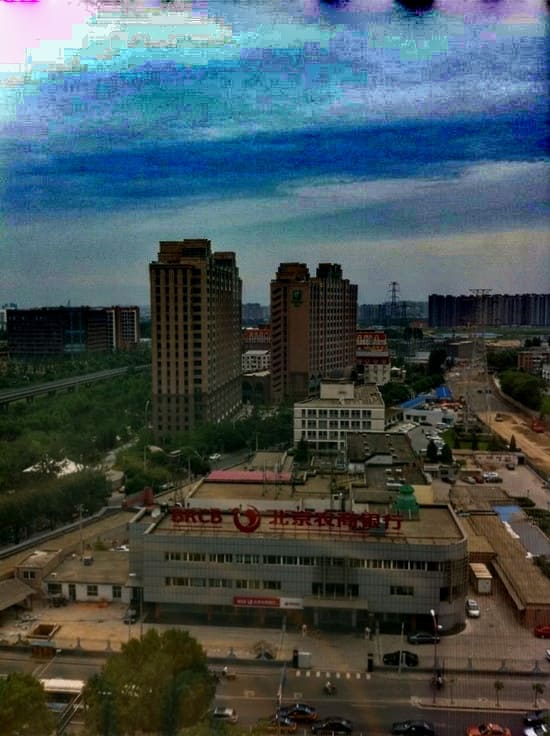}  

    \vspace{0.08cm}
    \includegraphics[width=1.01\linewidth]{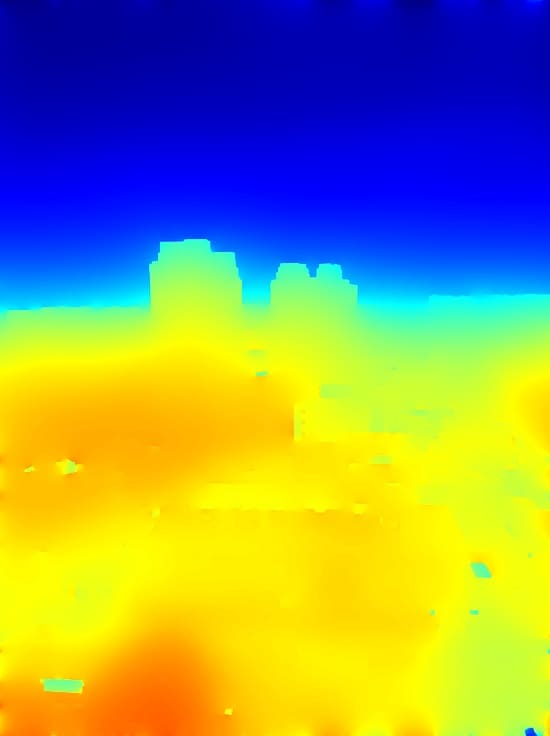}    

    \vspace{0.15cm}
    \includegraphics[width=1.01\linewidth]{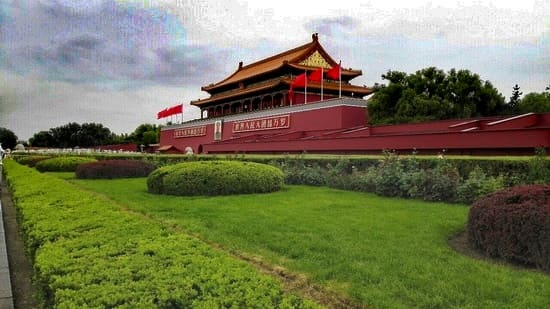}
    
    \vspace{0.08cm}
    \includegraphics[width=1.01\linewidth]{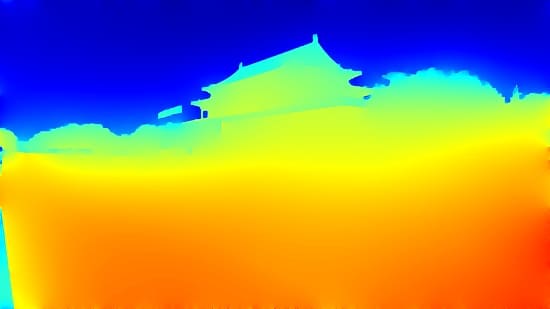}
    \caption{BCCR}
    \label{fig12:HSTS-Synthetic-d}
  \end{subfigure}
  \hfill
  \begin{subfigure}{0.120\linewidth}
    \includegraphics[width=1.01\linewidth]{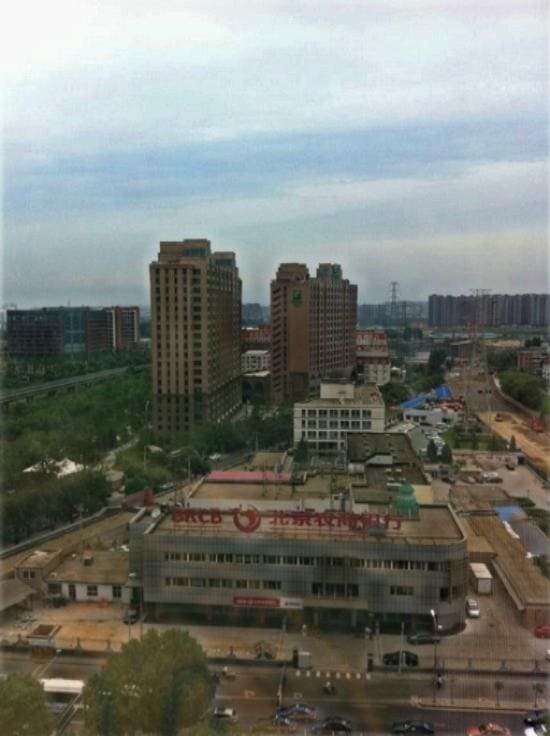}  

    \vspace{0.08cm}
    \includegraphics[width=1.01\linewidth]{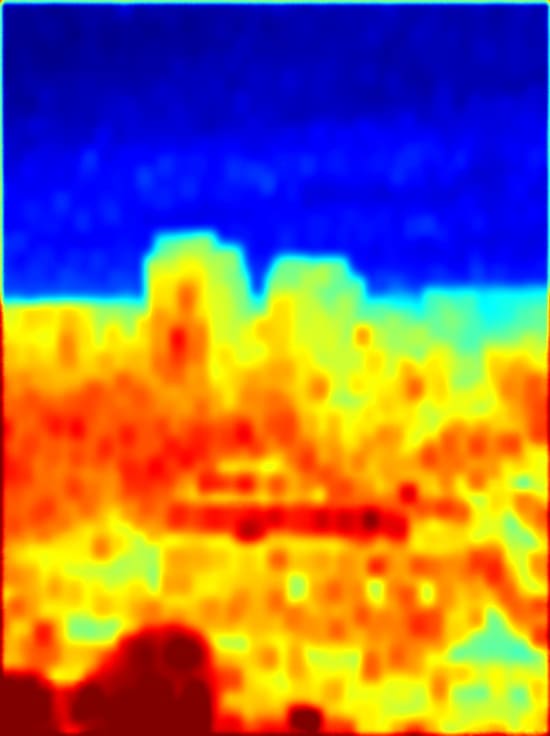}    

    \vspace{0.15cm}
    \includegraphics[width=1.01\linewidth]{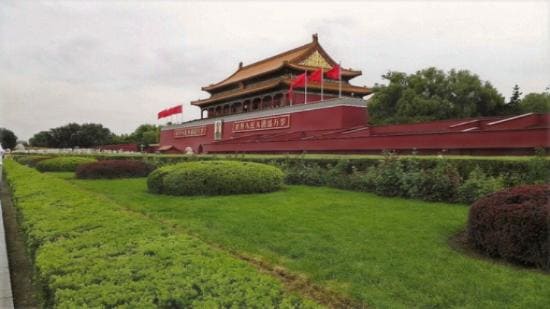}
    
    \vspace{0.08cm}
    \includegraphics[width=1.01\linewidth]{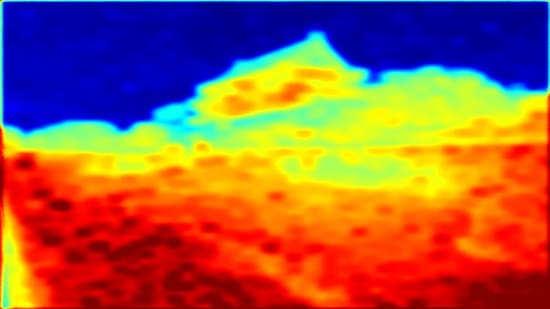}
    \caption{CAP}
    \label{fig12:HSTS-Synthetic-e}
  \end{subfigure}
  \hfill
  \begin{subfigure}{0.120\linewidth}
    \includegraphics[width=1.01\linewidth]{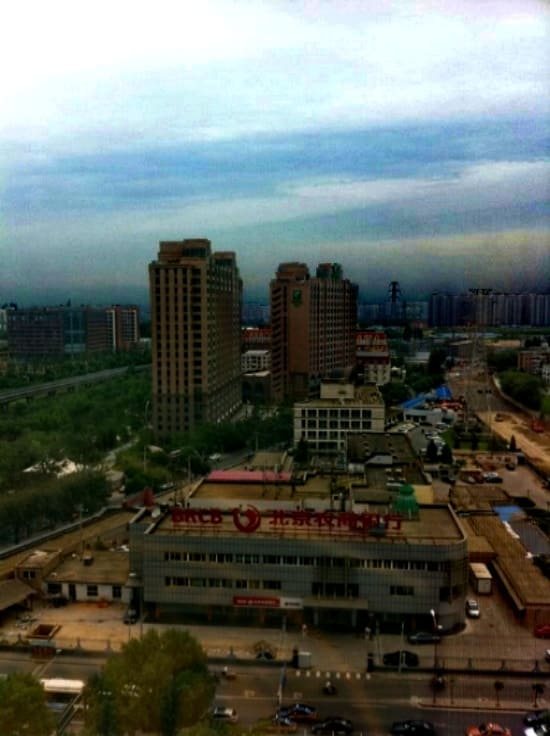}  

    \vspace{0.08cm}
    \includegraphics[width=1.01\linewidth]{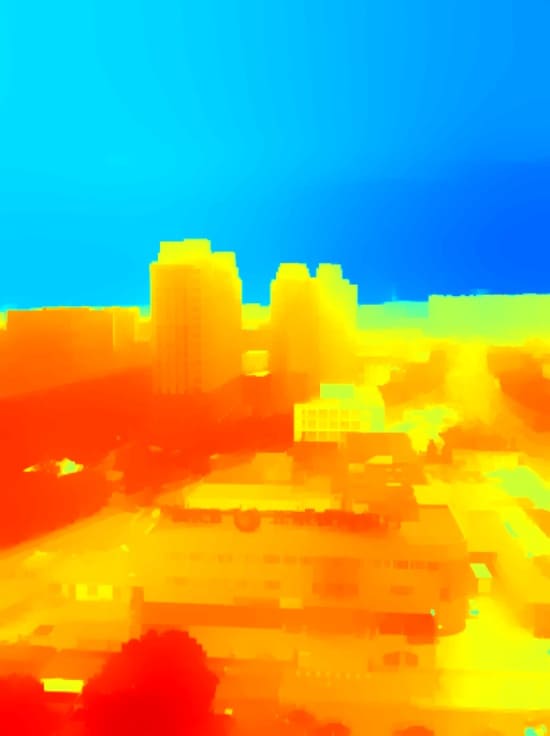}   

    \vspace{0.15cm}
    \includegraphics[width=1.01\linewidth]{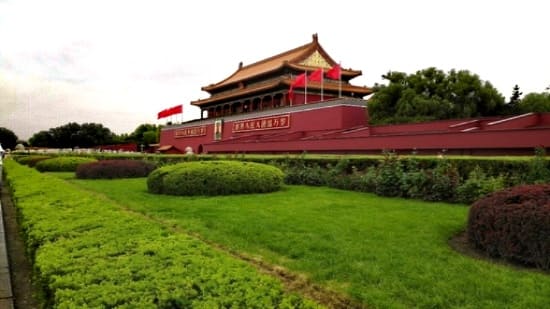}
    
    \vspace{0.08cm}
    \includegraphics[width=1.01\linewidth]{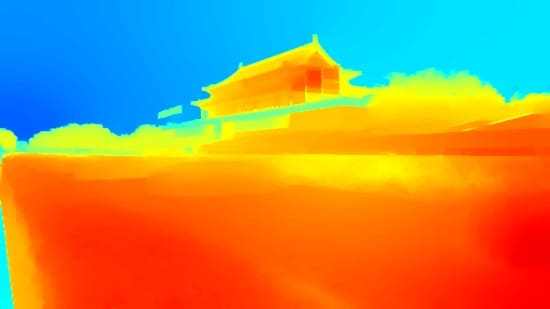} 
    \caption{NLID}
    \label{fig12:HSTS-Synthetic-f}
  \end{subfigure}
  \hfill
  \begin{subfigure}{0.120\linewidth}
    \includegraphics[width=1.01\linewidth]{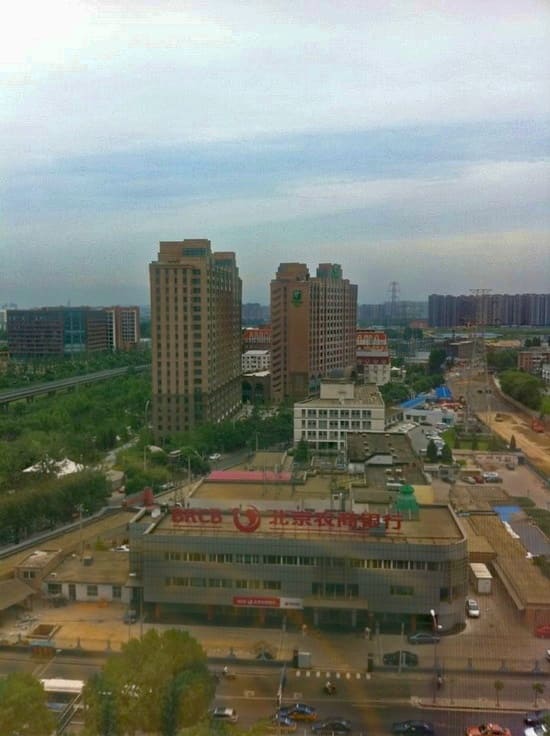}  

    \vspace{0.08cm}
    \includegraphics[width=1.01\linewidth]{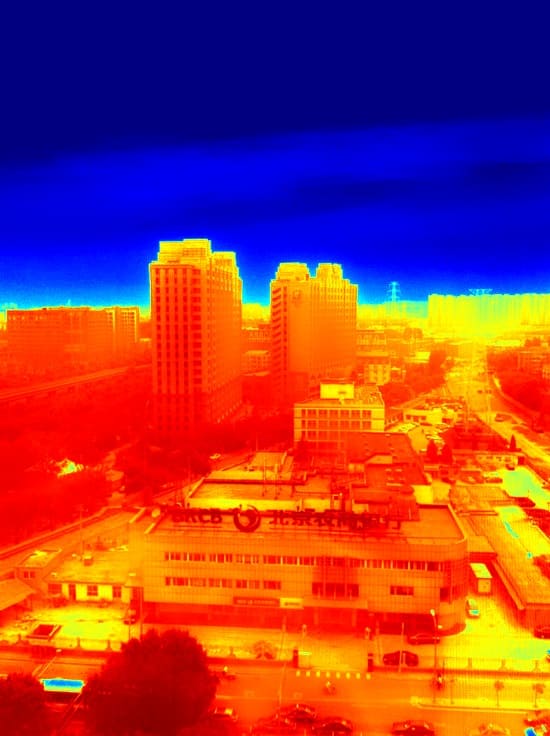}    

    \vspace{0.15cm}
    \includegraphics[width=1.01\linewidth]{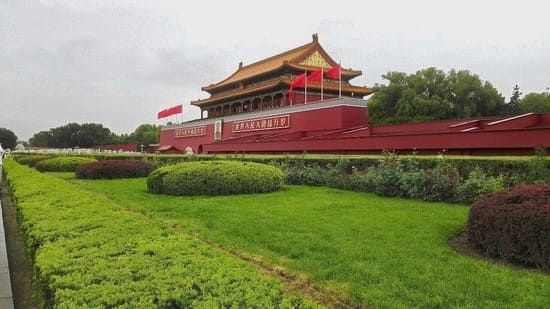}
    
    \vspace{0.08cm}
    \includegraphics[width=1.01\linewidth]{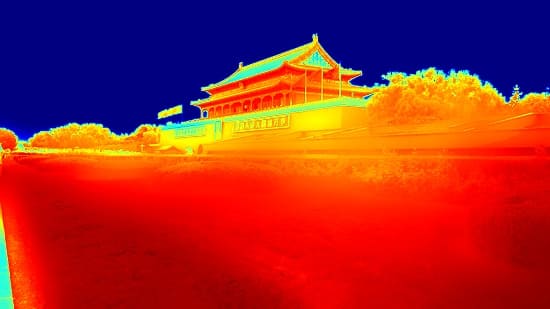}
    \caption{RSVT (ours)}
    \label{fig12:HSTS-Synthetic-g}
  \end{subfigure}
  }
  \caption{Typical dehazing results of various approaches on HSTS-Synthetic dataset: (a) hazy (top) and clean (bottom) images, (b-g) results by CEP \cite{bui2017single}, DCP \cite{he2010single}, BCCR \cite{meng2013efficient}, CAP \cite{zhu2015fast}, NLID \cite{berman2016non}, and the proposed RSVT method, respectively. The corresponding recovered transmission map for each method is shown below every output.}
  \label{fig12:HSTS-Synthetic}
  
\end{figure*}

\subsection{Evaluation of Scene Segmentation}
\label{subsec:softseg}

In order to validate the effectiveness of the proposed segmentation approach, a qualitative analysis is conducted to compare the segmentation masks generated by our approach and those obtained from other segmentation methods, such as the region-growing method \cite{pal1993review}, GMM-based segmentation \cite{titterington1985statistical}, and U\textsuperscript{2}-Net \cite{qin2020u2}. Note that the region-growing and GMM-based approaches have been adopted in several existing image decomposition-based dehazing schemes like \cite{wang2017dehazing, salazar2020fast, yu2016image}, while U\textsuperscript{2}-Net represents a typical robust neural network-based image enhancement method for images with the appearance of the sky \cite{liu2017single, liba2020sky}. Based on the visual comparisons presented in Fig. \ref{fig08:softseg-masks}, it is evident that our proposed method is capable of generating smooth soft segmentation masks with a natural transition from foreground to background and can avoid generating undesirable halos and artifacts at the boundary regions. Notably, our approach outperforms both the region-growing and GMM-based segmentation methods, and can be considered competitive when compared with the deep learning method, U\textsuperscript{2}-Net, while it is worth mentioning that our approach does not necessitate a training process.

Furthermore, the dehazing results obtained by combining our proposed RSVT prior with the above-mentioned segmentation methods are also examined and shown in Fig. \ref{fig08:softseg-masks}. It can be observed that our proposed approach can yield visually compelling restored images, whereas the other methods under comparison fail to produce satisfactory dehazing outputs in distant and dim areas where there is no clear boundary between the foreground and background regions. Further visual dehazing outcomes produced by the proposed framework along with the corresponding segmentation masks are presented in Fig. \ref{fig09:visual-results}.

\subsection{Selection of $R(x)$}
\label{subsec:parameter}

According to the constraint presented in Section \ref{subsec:3rd-assumption}, it is assumed that $R(x)$ could be set as a constant for any input image such that the intensities of the restored image are minimally affected when $R(x)$ varies within the range of its highest frequencies, $[2.0,5.0]$. To verify this assumption, we examine the impact of different values of $R(x)$ on the visibility of the outputs when $R(x)$ is set to in-range and out-of-range values, $R(x)=\{1,2,3,4,5,10\}$, where $R(x)=1$ and $R(x)=10$ are typical out-of-range values.

\begin{table}
  \centering
  \caption{Quantitative comparisons of various dehazing methods on HSTS-Realistic and RESIDE-Natural datasets. The best and second-best results are indicated in \textbf{bold} and \textcolor{blue}{blue}, respectively.}
  \resizebox{0.49\textwidth}{!}{
  \begin{tabular}{cccccc}
    \toprule
    \multirow{2}{*}{Type} & \multirow{2}{*}{Method} & \multicolumn{2}{c}{HSTS-Realistic} & \multicolumn{2}{c}{RESIDE-Natural} \\
    \cmidrule{3-6}
           & & NIQE & BRISQUE & NIQE & BRISQUE \\
    \midrule
    \multirow{6}{*}{Prior-based} & CEP \cite{bui2017single} & 13.583 & 49.785 & 12.260 & 30.419 \\
    & DCP \cite{he2010single} & 12.318 & 52.234 & 14.181 & 30.947 \\
    & CLAHE \cite{pizer1987adaptive} & 13.450 & 50.272 & 13.446 & 23.911 \\
    & BCCR \cite{meng2013efficient} & 13.713 & 49.261 & 14.052 & 34.072 \\
    & NLID \cite{berman2016non} & 14.146 & 62.126 & 13.493 & 41.265 \\
    & CAP \cite{zhu2015fast} & 13.899 & \textcolor{blue}{38.305} & 12.900 & 33.001 \\
    \midrule
    \multirow{4}{*}{CNN-based} & AOD-Net \cite{li2017aod} & \textbf{12.554} & 41.200 & \textcolor{blue}{11.971} & 24.658 \\
    & GFN \cite{ren2018gated} & 14.612 & 48.178 & 12.970 & 24.983 \\
    & GCANet \cite{chen2019gated} & 15.303 & 58.829 & 12.406 & \textcolor{blue}{23.360} \\
    & YOLY \cite{li2021you} & 14.542 & 46.535 & 12.854 & 24.235 \\
    \midrule
    Prior-based & RSVT (ours) & \textcolor{blue}{13.414} & \textbf{36.711} & \textbf{11.780} & \textbf{20.659} \\
    \bottomrule
  \end{tabular}}
  \label{tab03:natural-results}
\end{table}

The results of the quantitative evaluation are summarized in Table \ref{tab01:Rx}, which shows that the best performance is achieved when $R(x)$ is set closer to the highest-frequency value, specifically $R(x) = 3$. Also, the other in-range values of $R(x)$ such as $R(x)=\{2, 4, 5\}$ also yield competitive results despite of slight drops in performance. On the contrary, the use of out-of-range values, $R(x)=\{1,10\}$, results in a degradation in the dehazing performance. To visually verify these measurements, qualitative restoration results are demonstrated in Fig. \ref{fig10:Rx}. As can be seen Fig. Fig. \ref{fig10:Rx}, $R(x) \in [2.0,5.0]$ can result in competitive visual dehazing outcomes whereas $R(x)=\{1,10\}$ may lead to color distortion or over-saturation. Based on the obtained distribution shown in Fig. \ref{fig03:statistics-c} and the quantitative evaluation summarized in Table \ref{tab01:Rx}, we set $R(x)=3$ as the primary configuration in the proposed framework.

\begin{figure*}
  \centering  
  \resizebox{0.98\textwidth}{!}{
  \begin{subfigure}{0.120\linewidth}
    \includegraphics[width=1.01\linewidth]{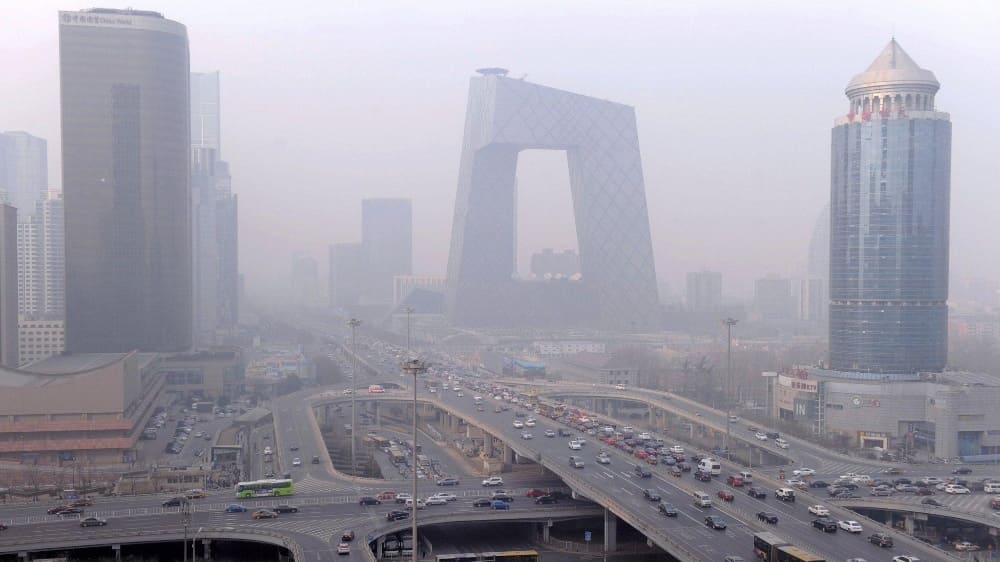}
    
    \vspace{0.08cm}
    \includegraphics[width=1.01\linewidth]{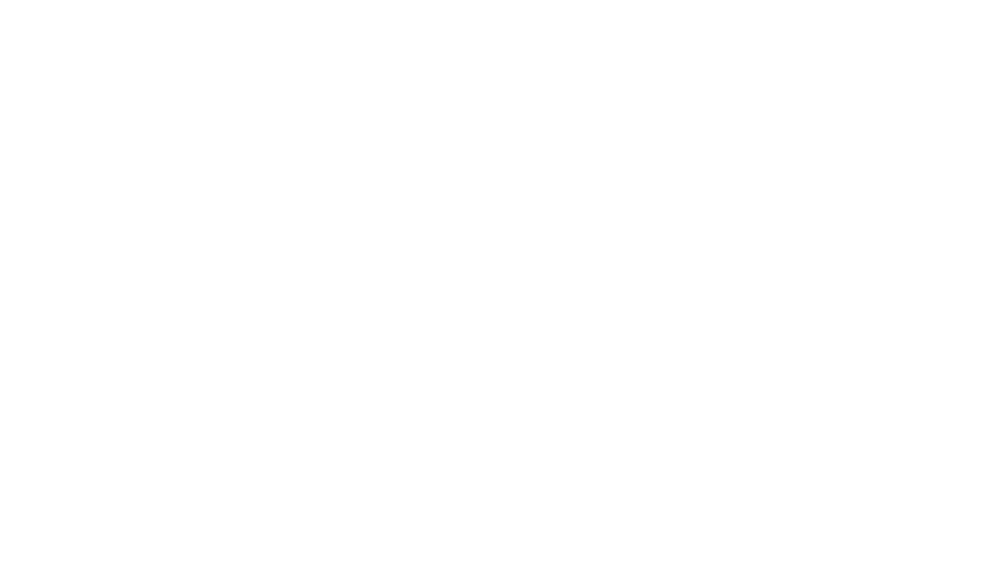}  

    \vspace{0.15cm}
    \includegraphics[width=1.01\linewidth]{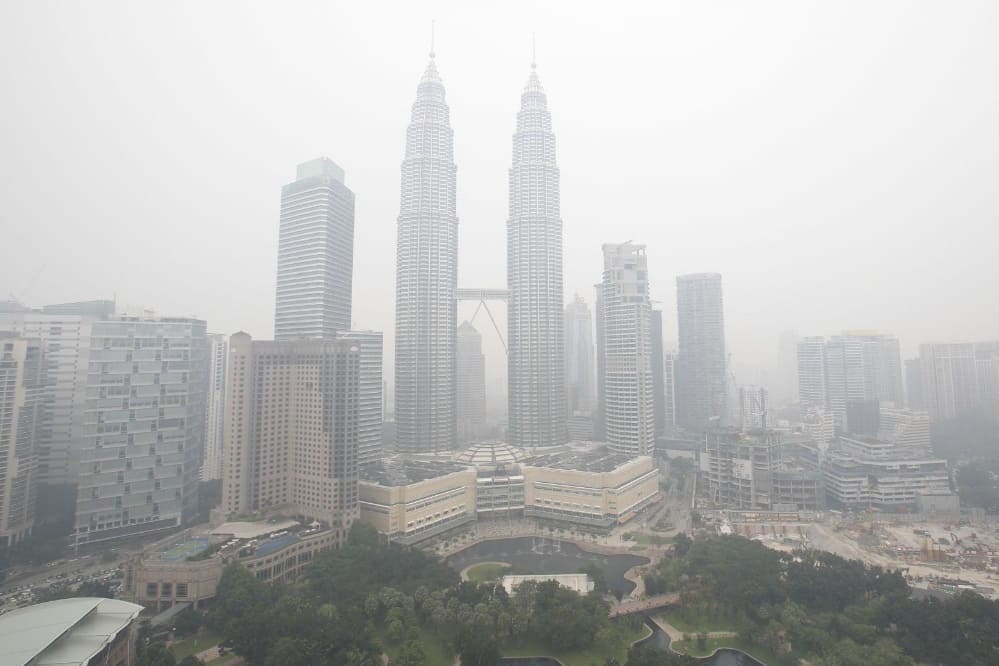}
    
    \vspace{0.08cm}
    \includegraphics[width=1.01\linewidth]{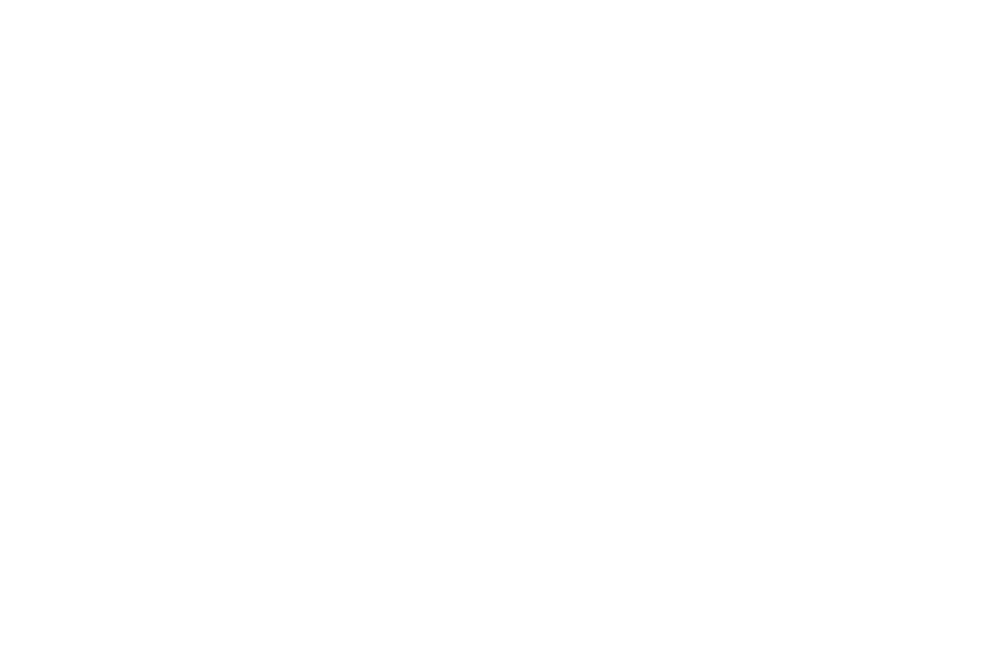}
    \caption{Hazy}
    \label{fig13:HSTS-Realistic-a}
  \end{subfigure}
  \hfill
  \begin{subfigure}{0.120\linewidth} 
    \includegraphics[width=1.01\linewidth]{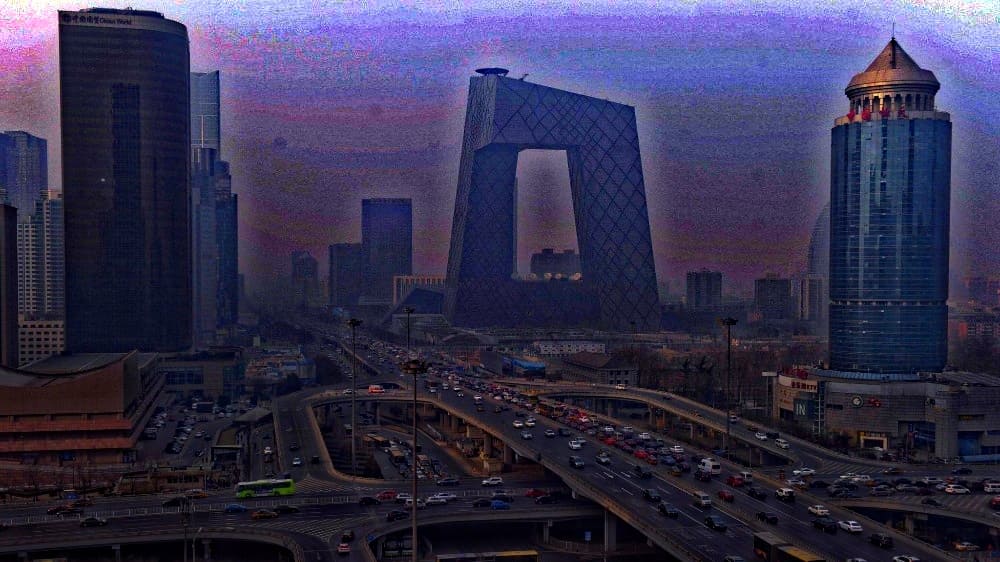}  

    \vspace{0.08cm}
    \includegraphics[width=1.01\linewidth]{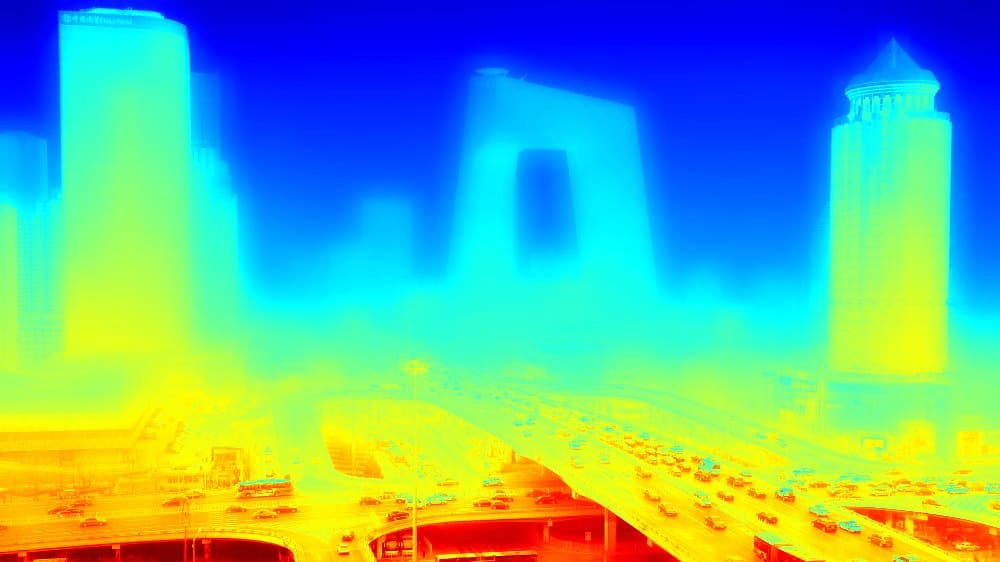}   

    \vspace{0.15cm}
    \includegraphics[width=1.01\linewidth]{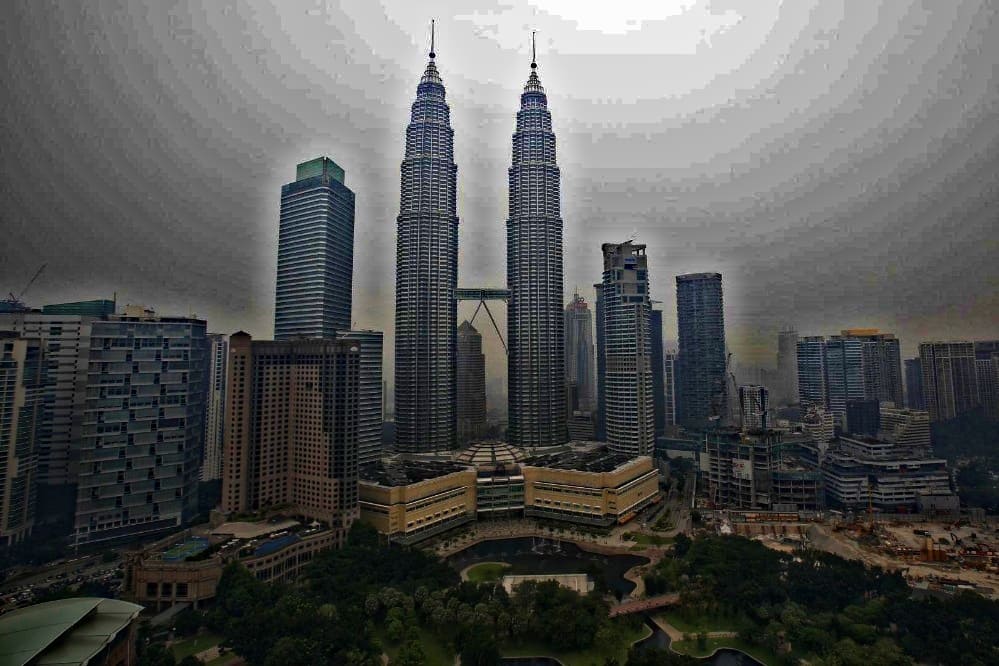}
    
    \vspace{0.08cm}
    \includegraphics[width=1.01\linewidth]{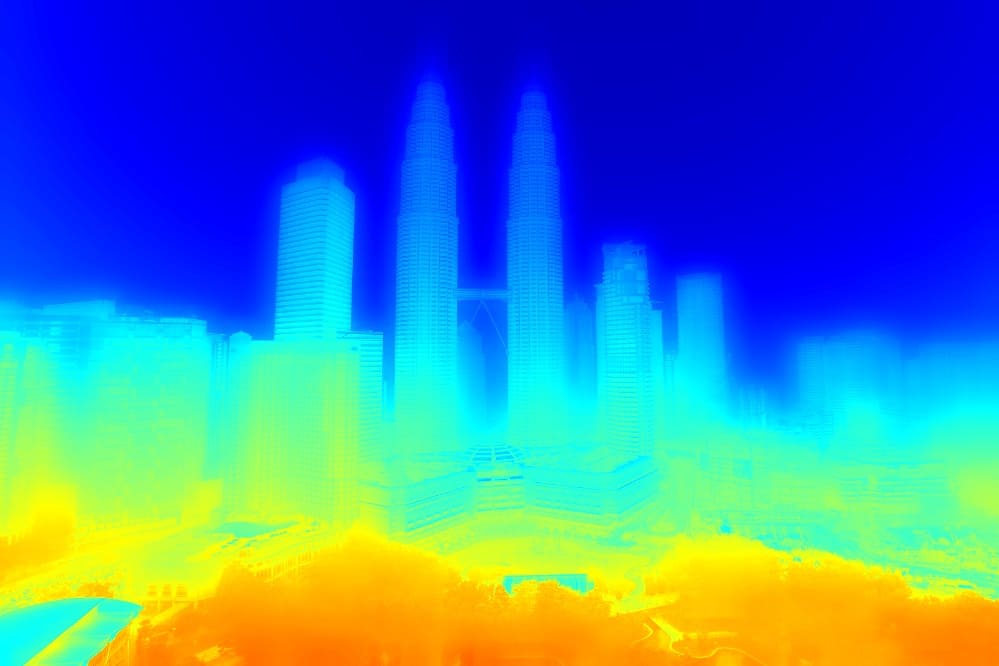}
    \caption{CEP}
    \label{fig13:HSTS-Realistic-b}
  \end{subfigure}
  \hfill
  \begin{subfigure}{0.120\linewidth}
    \includegraphics[width=1.01\linewidth]{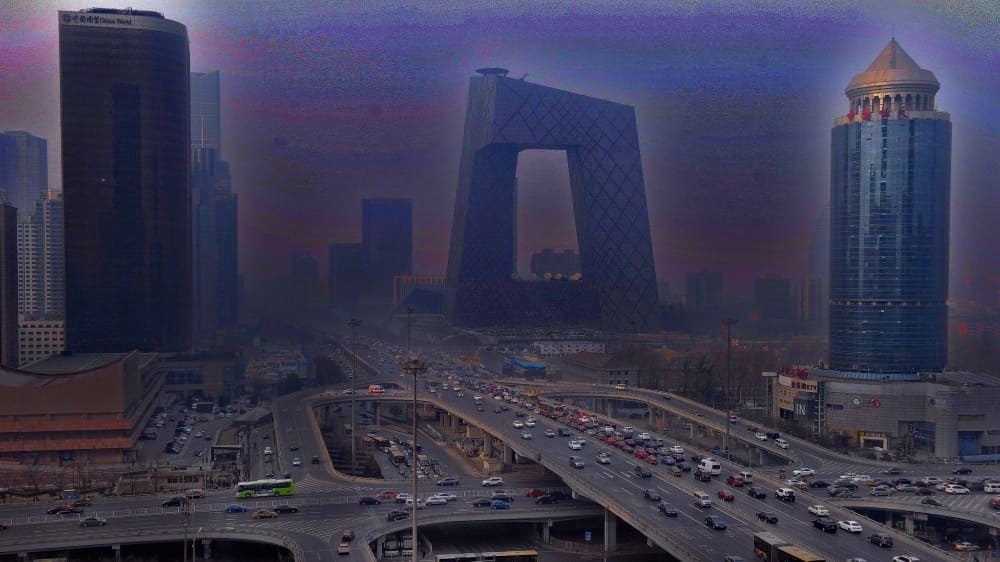}  

    \vspace{0.08cm}
    \includegraphics[width=1.01\linewidth]{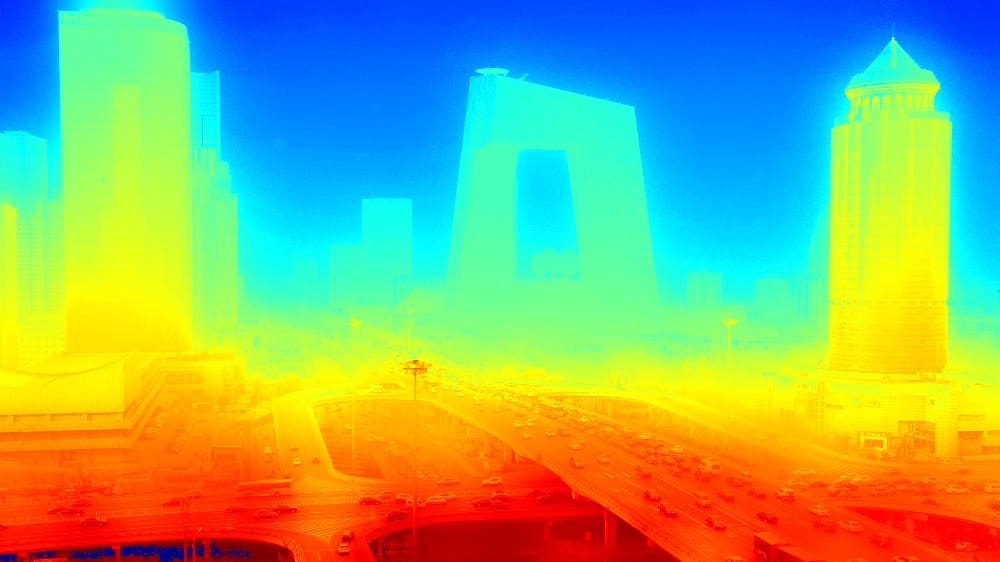}    

    \vspace{0.15cm}
    \includegraphics[width=1.01\linewidth]{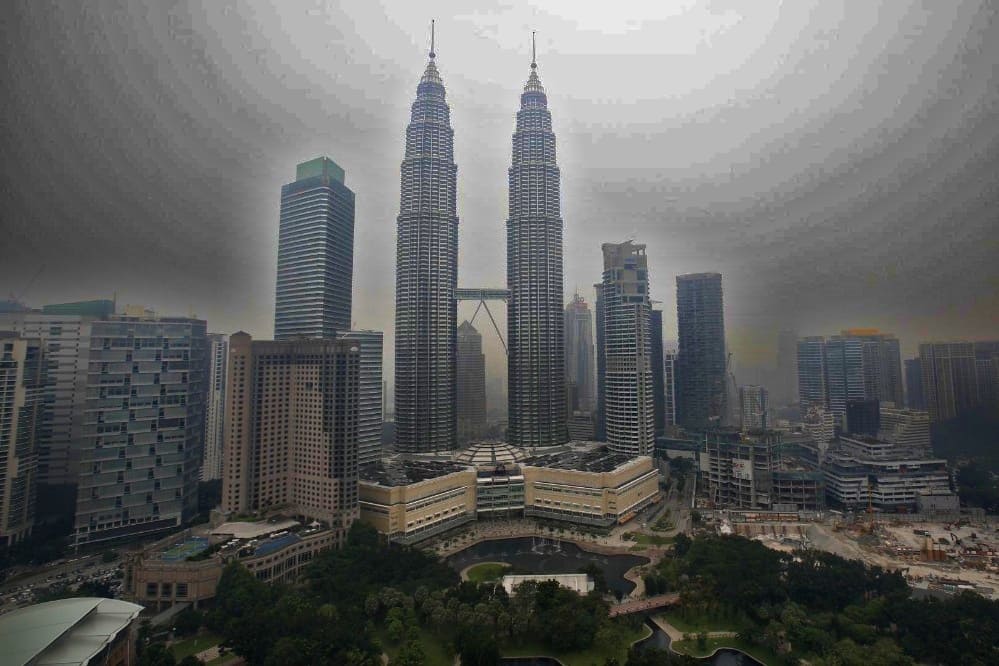}
    
    \vspace{0.08cm}
    \includegraphics[width=1.01\linewidth]{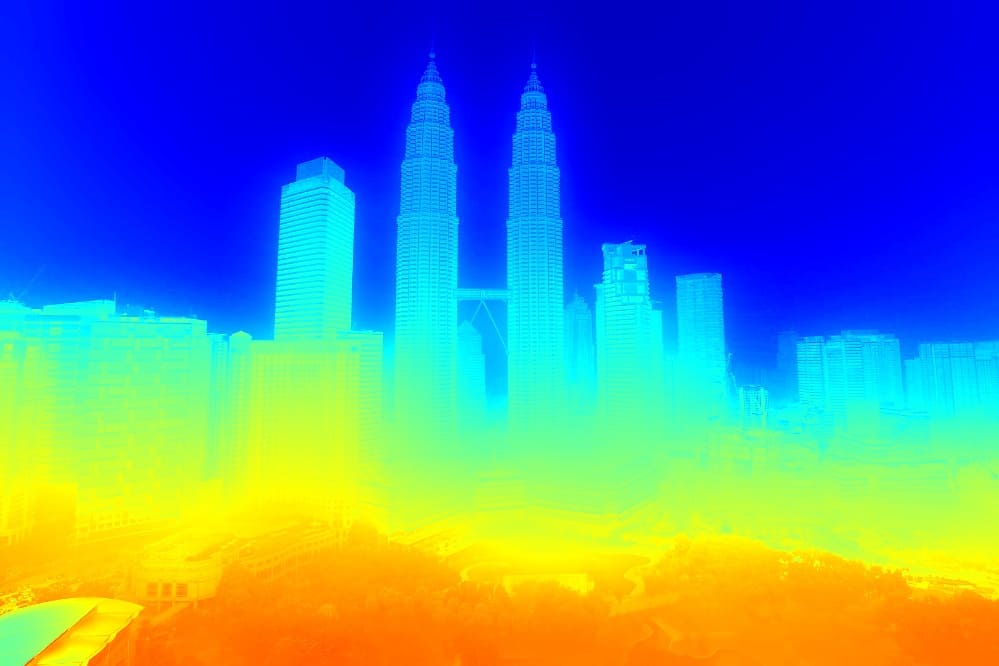}
    \caption{DCP}
    \label{fig13:HSTS-Realistic-c}
  \end{subfigure}
  \hfill
  \begin{subfigure}{0.120\linewidth}
    \includegraphics[width=1.01\linewidth]{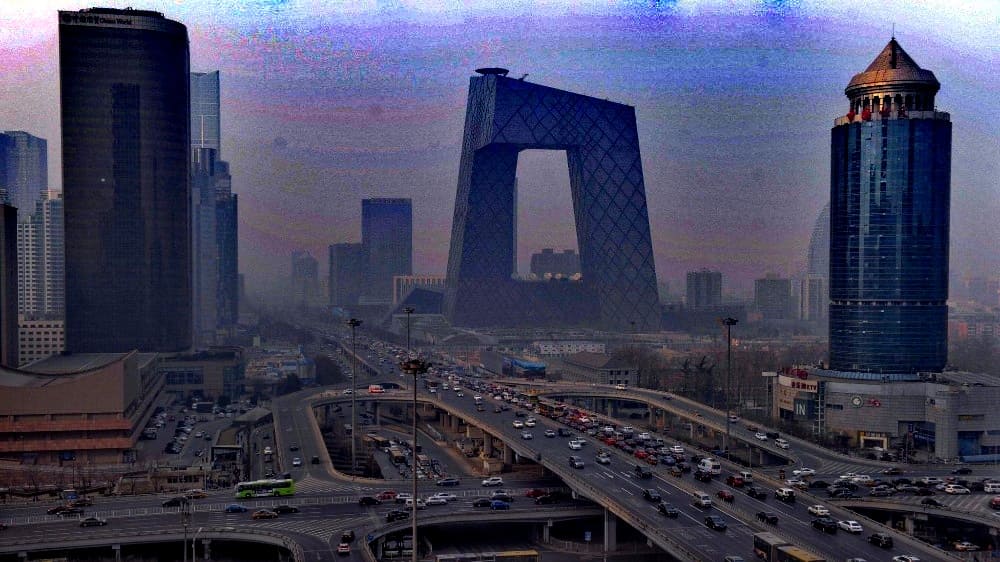}  

    \vspace{0.08cm}
    \includegraphics[width=1.01\linewidth]{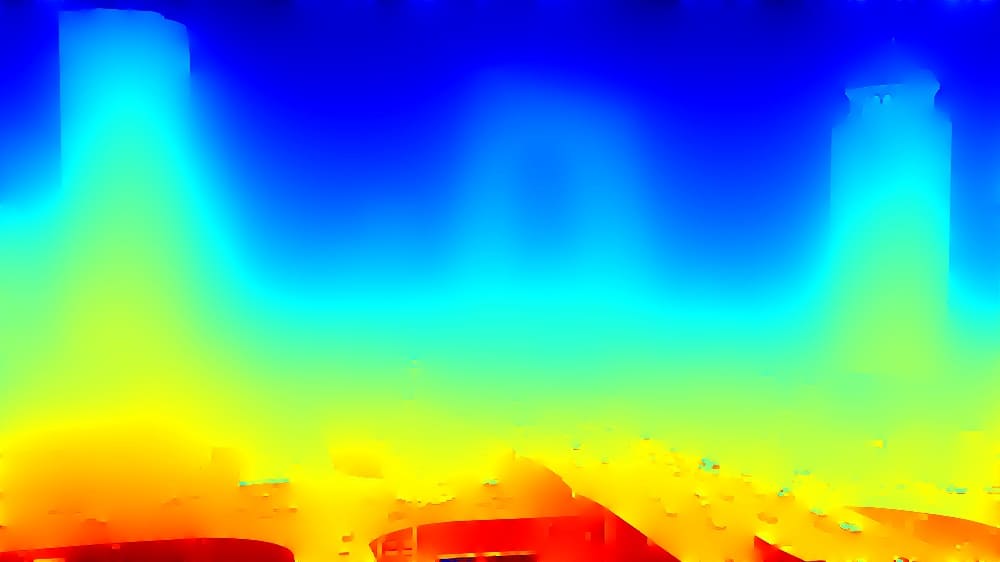}    

    \vspace{0.15cm}
    \includegraphics[width=1.01\linewidth]{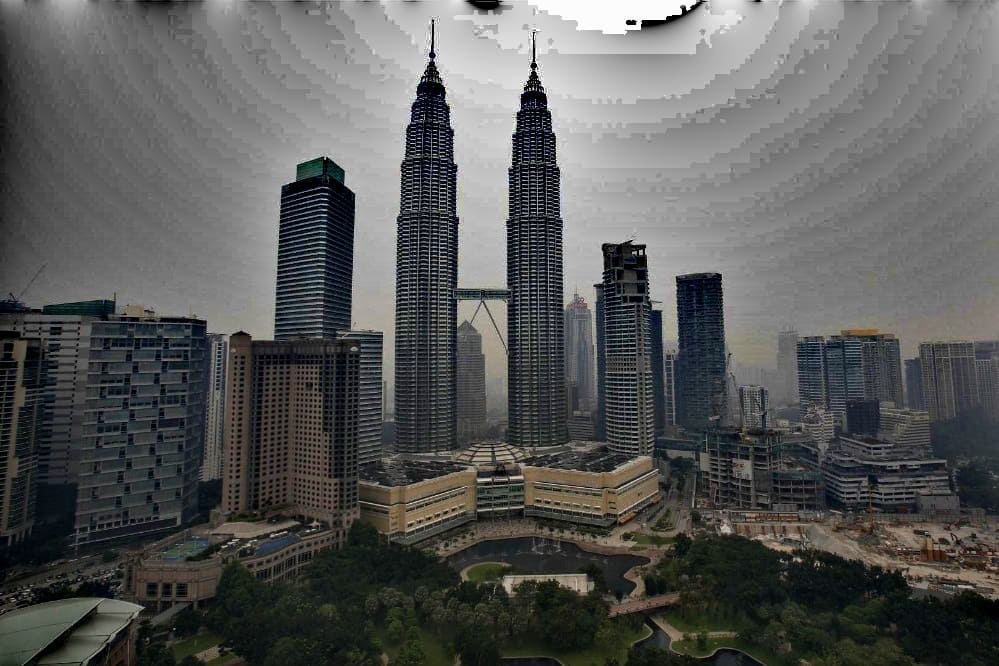}
    
    \vspace{0.08cm}
    \includegraphics[width=1.01\linewidth]{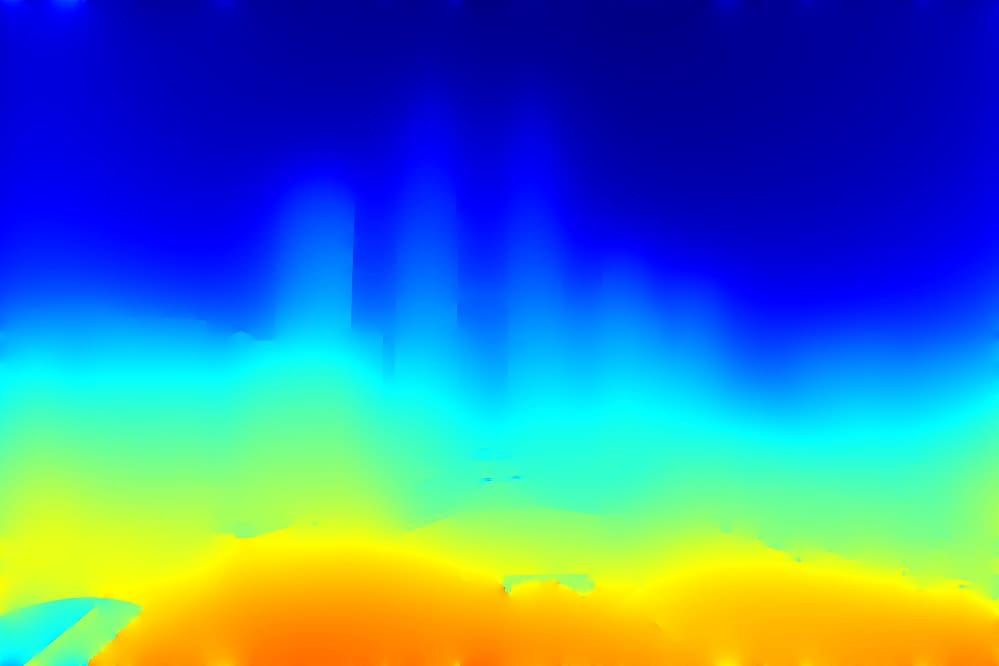}
    \caption{BCCR}
    \label{fig13:HSTS-Realistic-d}
  \end{subfigure}
  \hfill
  \begin{subfigure}{0.120\linewidth}
    \includegraphics[width=1.01\linewidth]{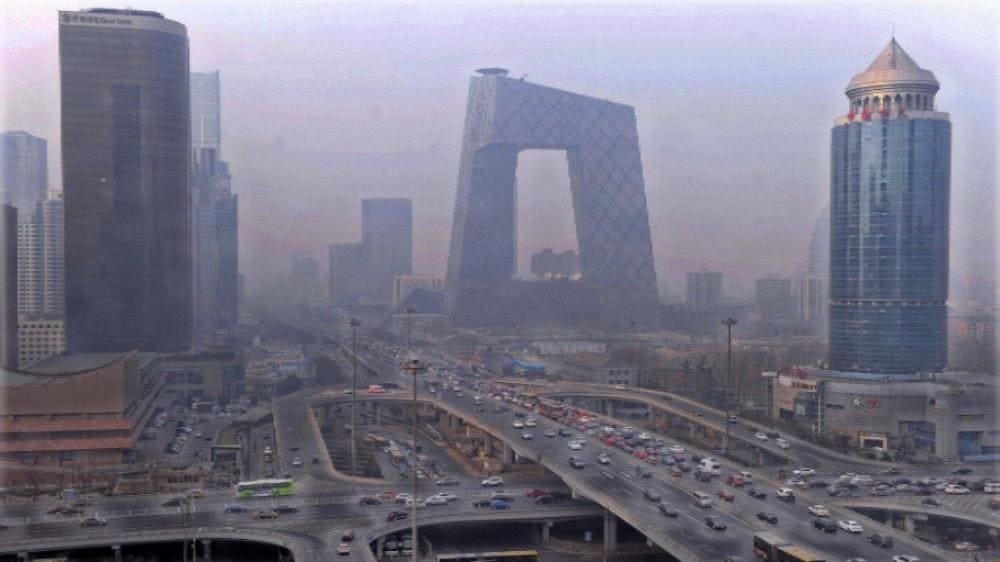}  

    \vspace{0.08cm}
    \includegraphics[width=1.01\linewidth]{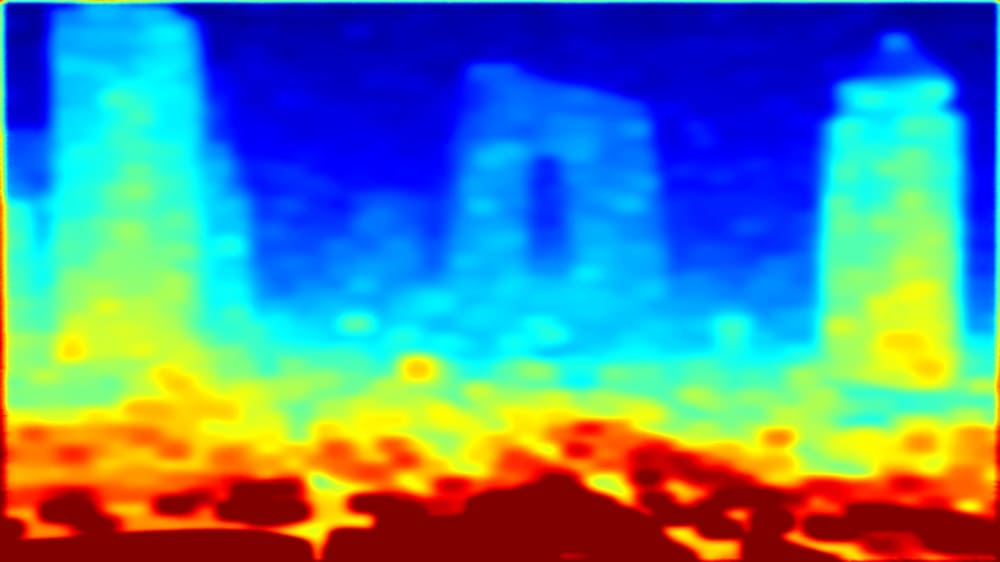}    

    \vspace{0.15cm}
    \includegraphics[width=1.01\linewidth]{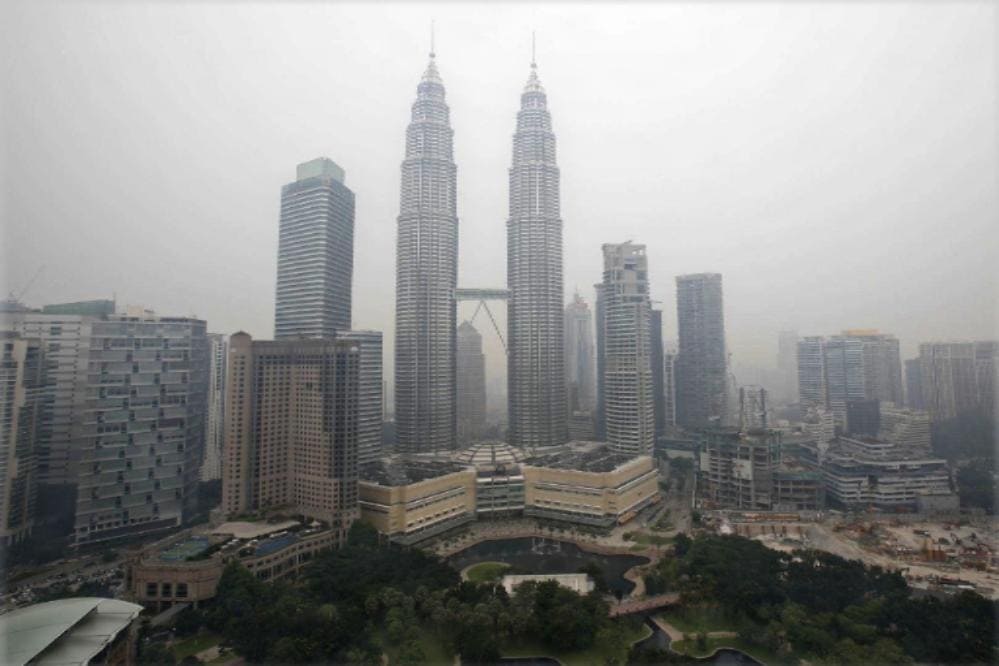}
    
    \vspace{0.08cm}
    \includegraphics[width=1.01\linewidth]{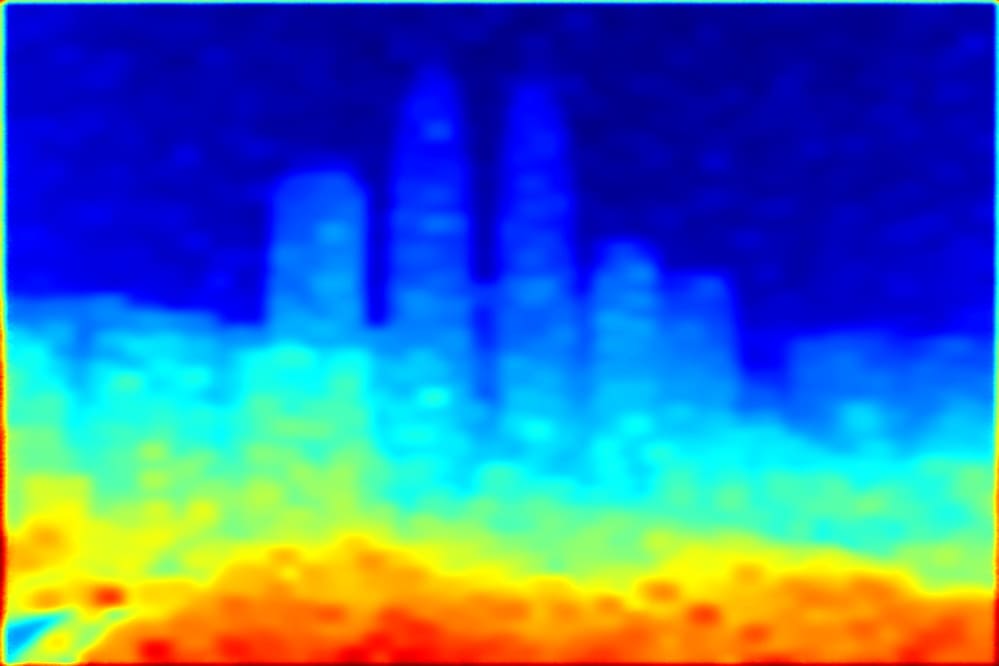}
    \caption{CAP}
    \label{fig13:HSTS-Realistic-e}
  \end{subfigure}
  \hfill
  \begin{subfigure}{0.120\linewidth}
    \includegraphics[width=1.01\linewidth]{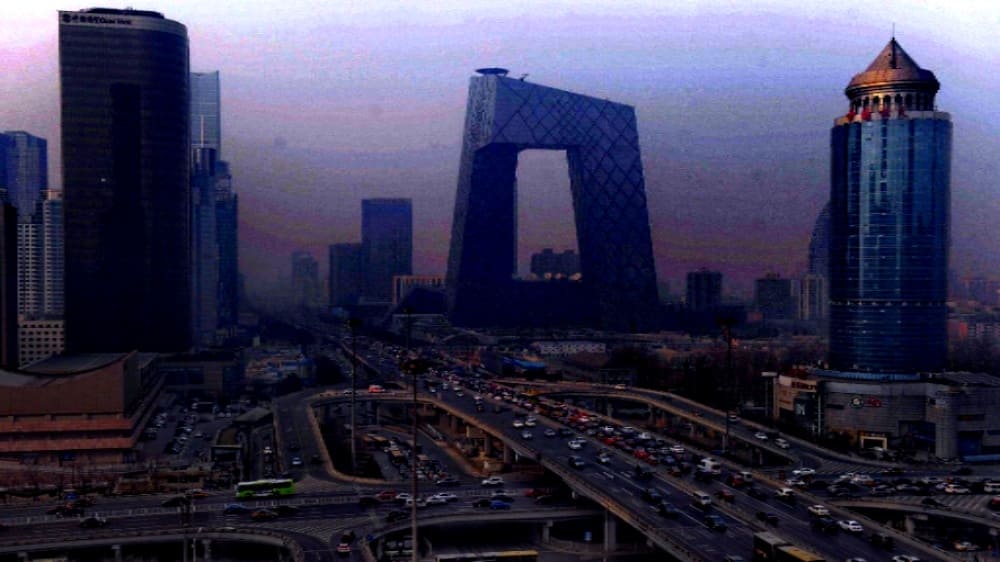}  

    \vspace{0.08cm}
    \includegraphics[width=1.01\linewidth]{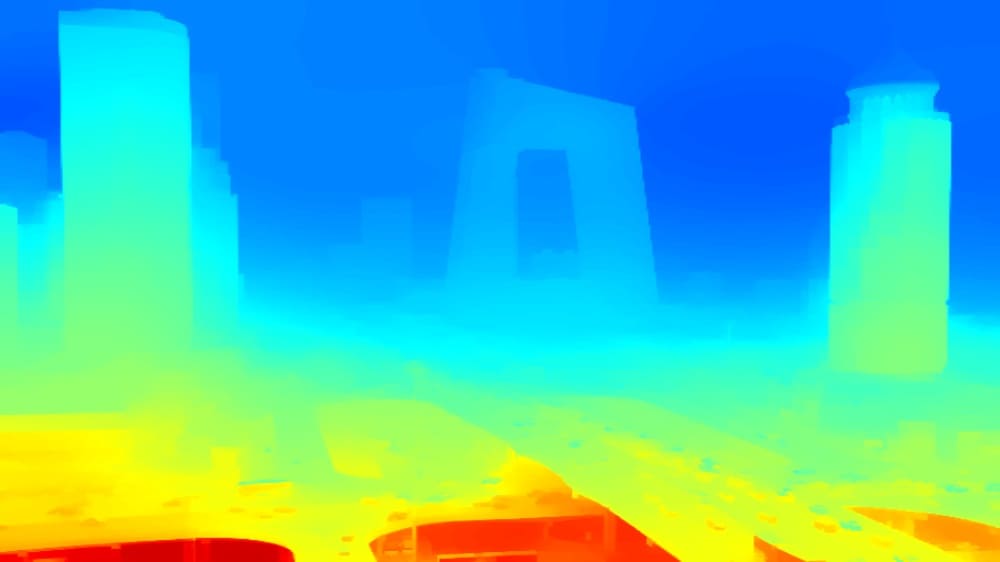}   

    \vspace{0.15cm}
    \includegraphics[width=1.01\linewidth]{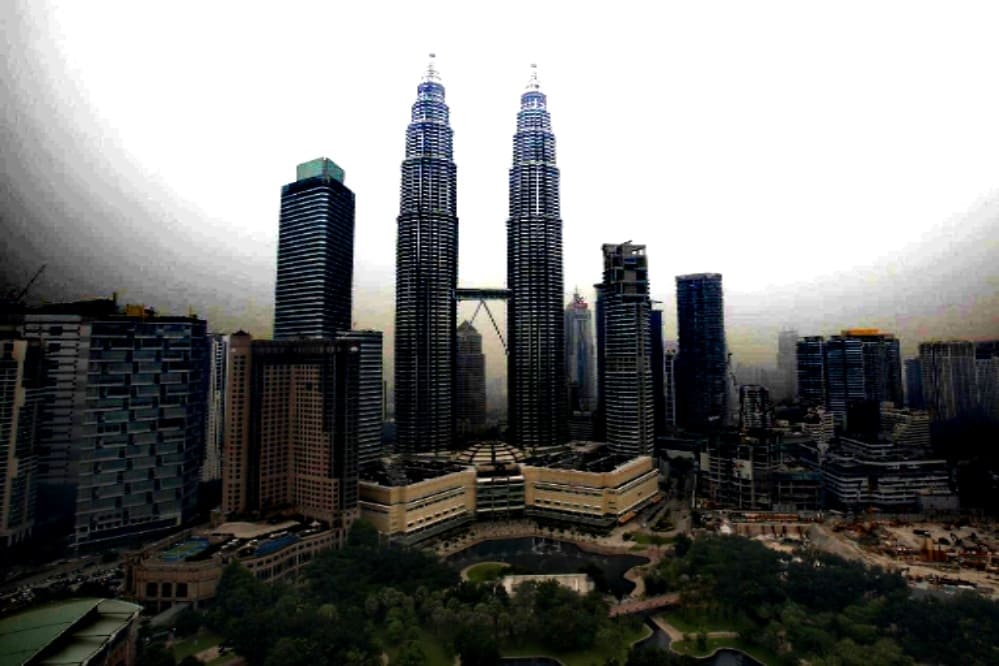}
    
    \vspace{0.08cm}
    \includegraphics[width=1.01\linewidth]{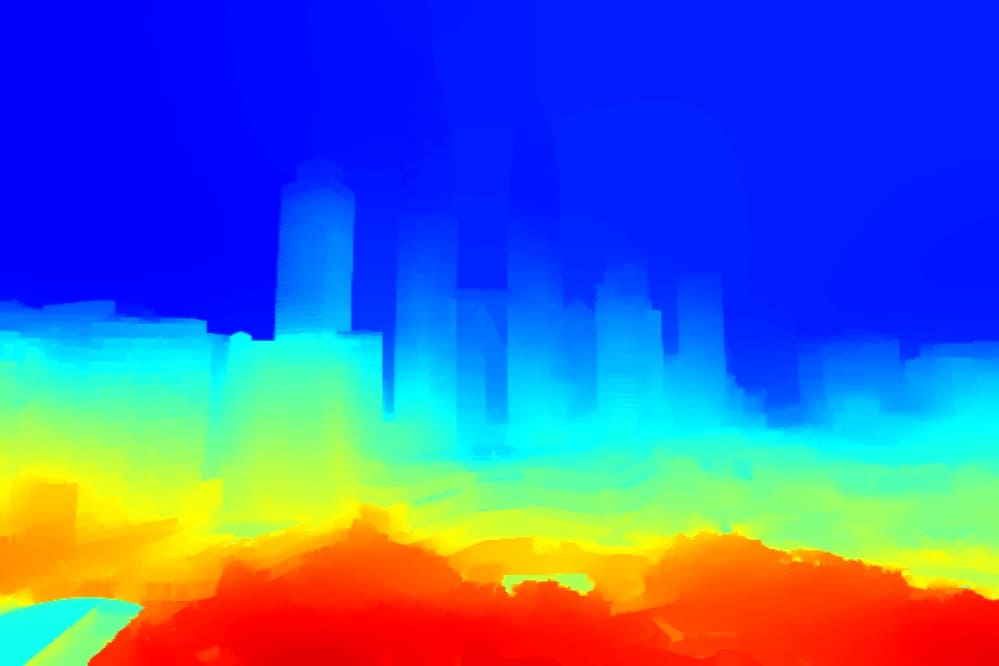} 
    \caption{NLID}
    \label{fig13:HSTS-Realistic-f}
  \end{subfigure}
  \hfill
  \begin{subfigure}{0.120\linewidth}
    \includegraphics[width=1.01\linewidth]{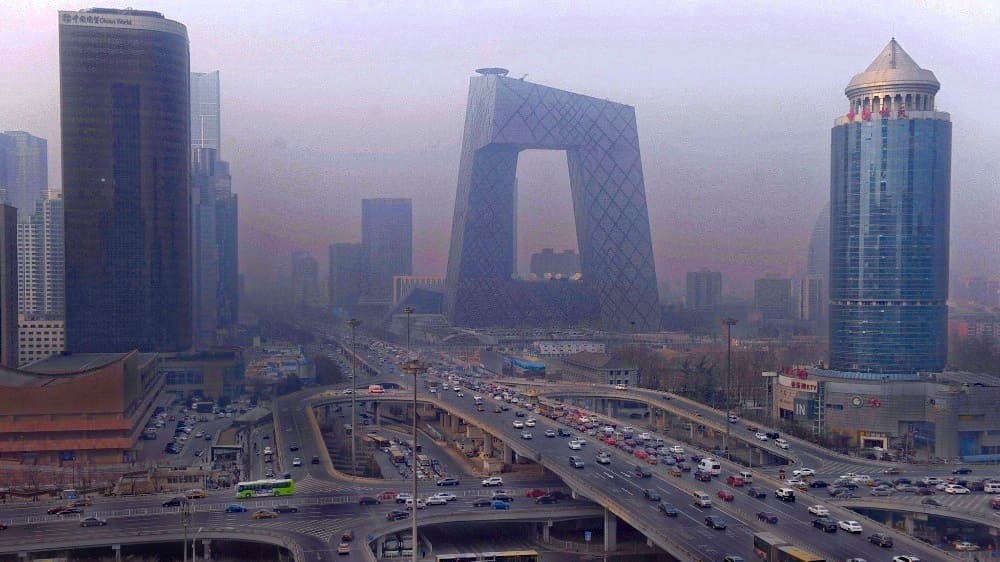}  

    \vspace{0.08cm}
    \includegraphics[width=1.01\linewidth]{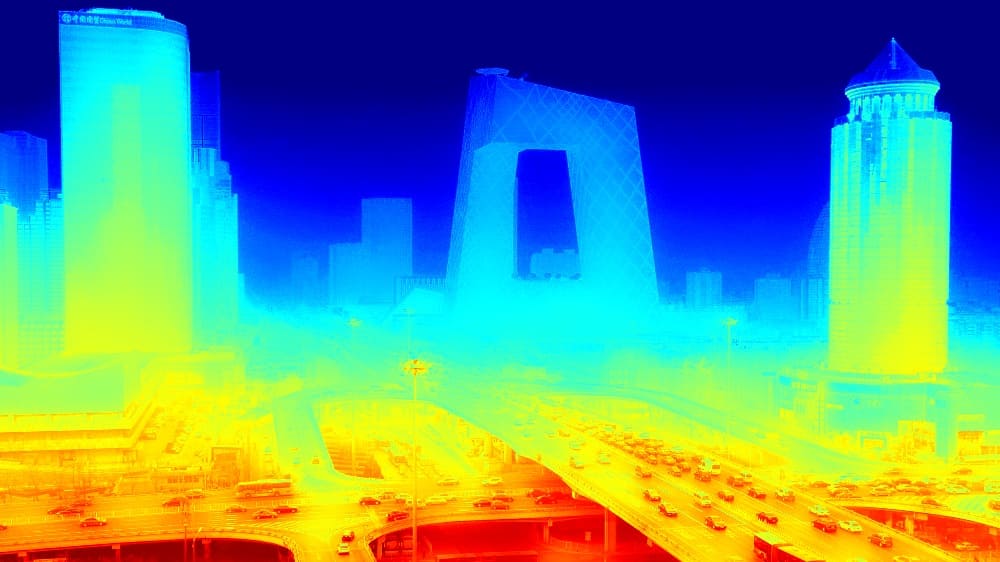}    

    \vspace{0.15cm}
    \includegraphics[width=1.01\linewidth]{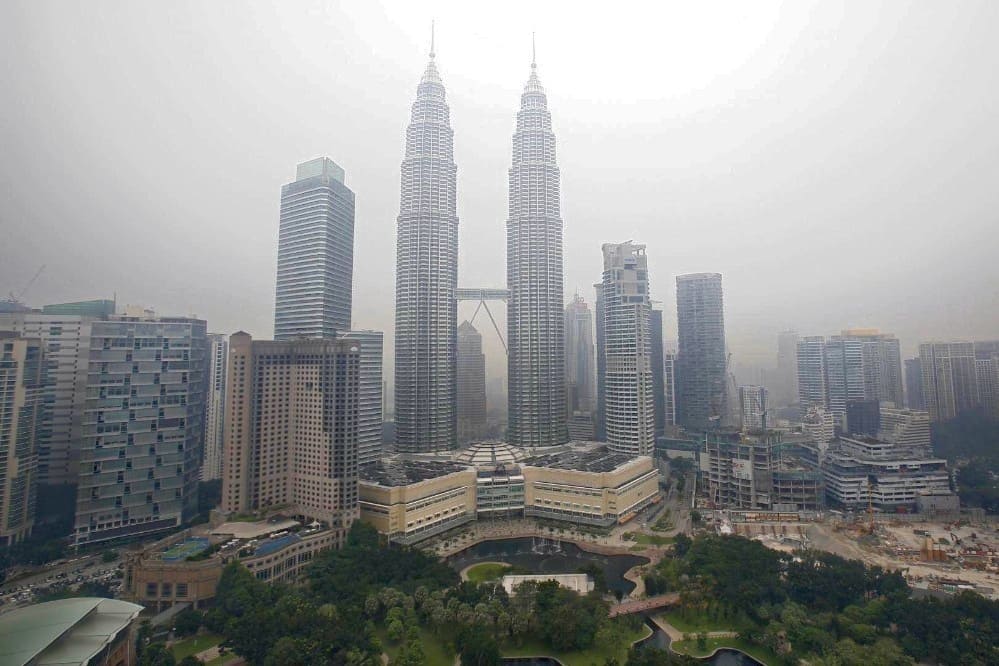}
    
    \vspace{0.08cm}
    \includegraphics[width=1.01\linewidth]{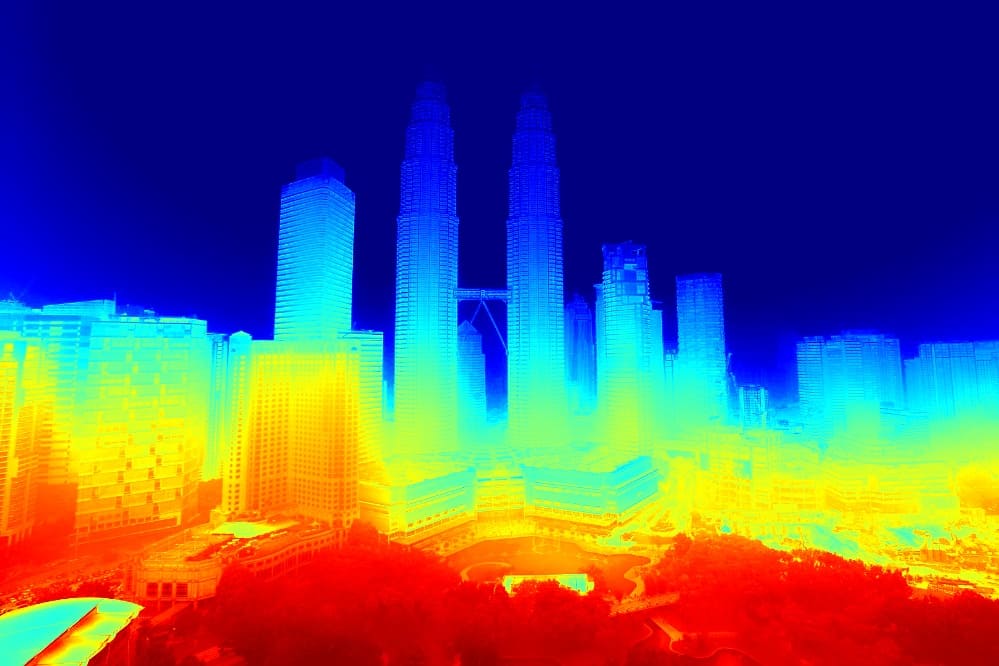}
    \caption{RSVT (ours)}
    \label{fig13:HSTS-Realistic-g}
  \end{subfigure}
  }
  \caption{Typical dehazing results of various approaches on HSTS-Realistic dataset: (a) hazy image, (b-g) results by CEP \cite{bui2017single}, DCP \cite{he2010single}, BCCR \cite{meng2013efficient}, CAP \cite{zhu2015fast}, NLID \cite{berman2016non}, and the proposed RSVT method, respectively. The corresponding recovered transmission map for each method is shown below every output.}
  \label{fig13:HSTS-Realistic}

\end{figure*}

\begin{figure*}
  \centering  
  \resizebox{0.98\textwidth}{!}{
  \begin{subfigure}{0.120\linewidth}
    \includegraphics[width=1.01\linewidth]{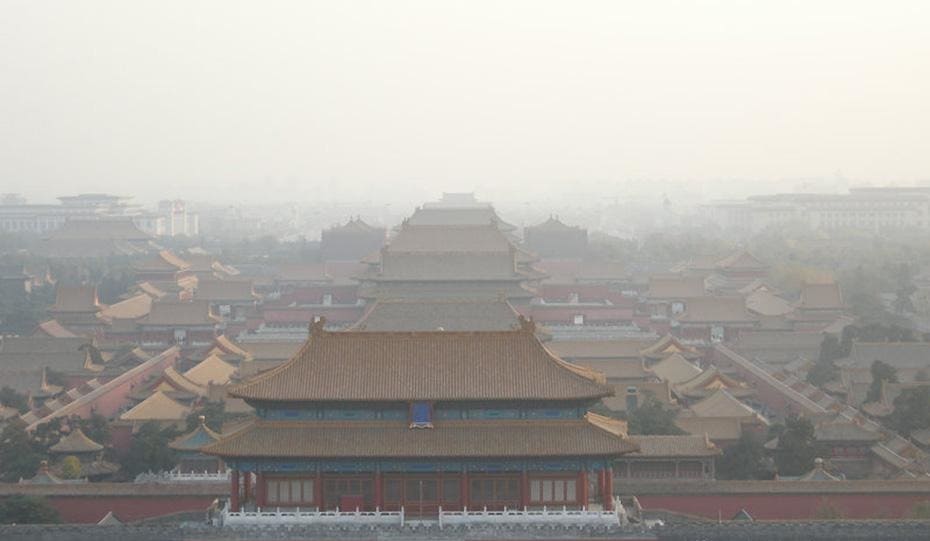}
    
    \vspace{0.08cm}
    \includegraphics[width=1.01\linewidth]{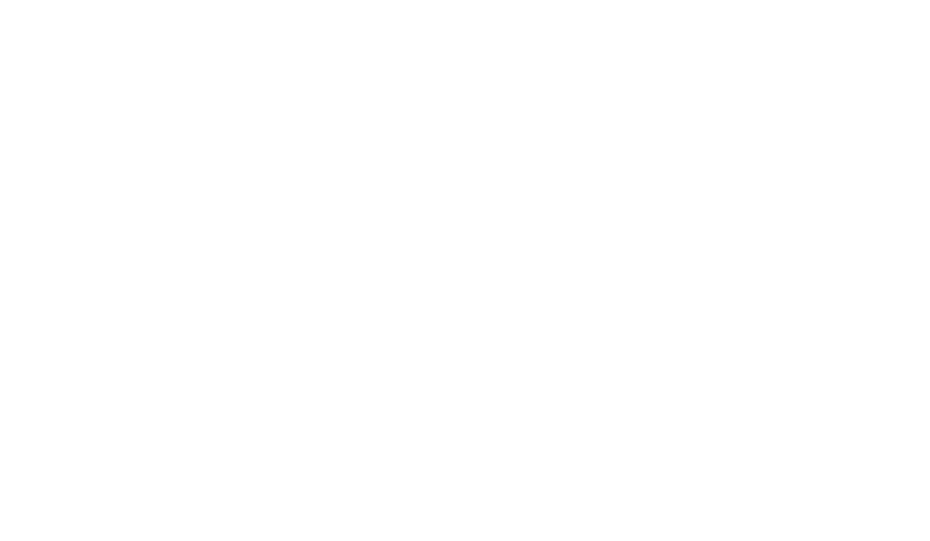}  

    \vspace{0.15cm}
    \includegraphics[width=1.01\linewidth]{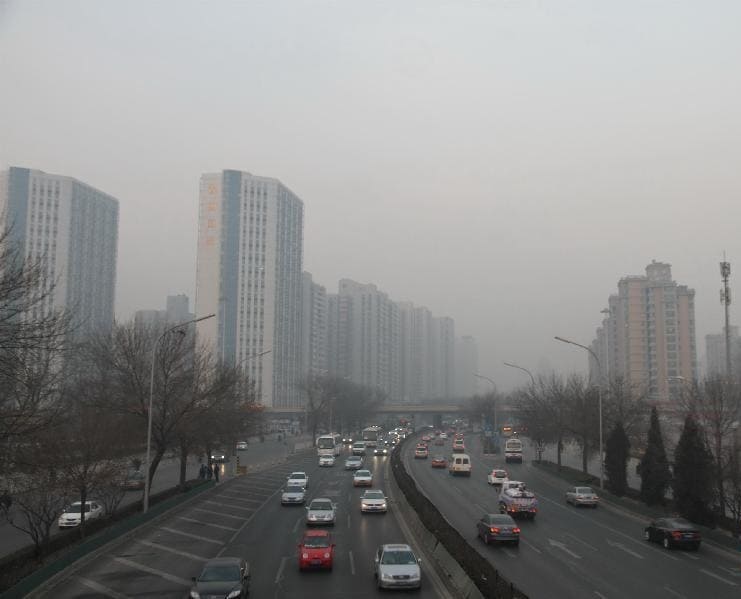}
    
    \vspace{0.08cm}
    \includegraphics[width=1.01\linewidth]{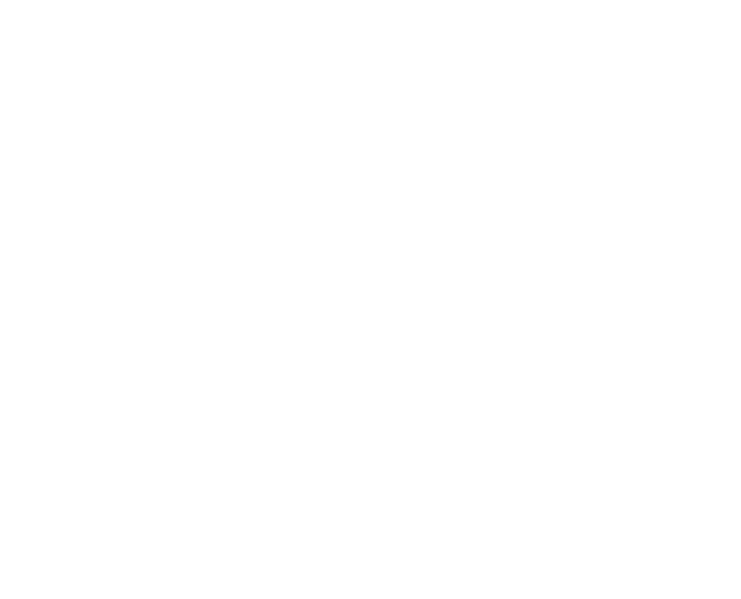}
    \caption{Hazy}
    \label{fig14:RESIDE-Natural-a}
  \end{subfigure}
  \hfill
  \begin{subfigure}{0.120\linewidth} 
    \includegraphics[width=1.01\linewidth]{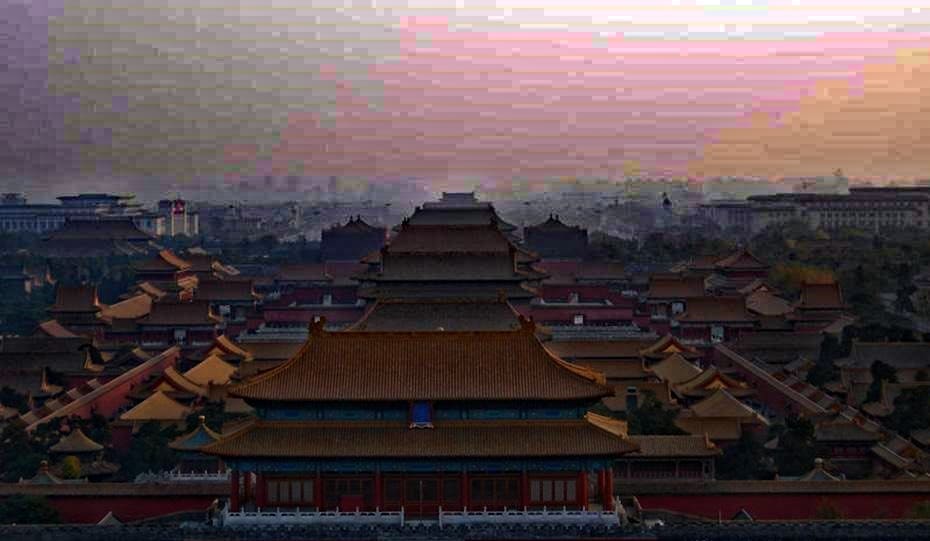}  

    \vspace{0.08cm}
    \includegraphics[width=1.01\linewidth]{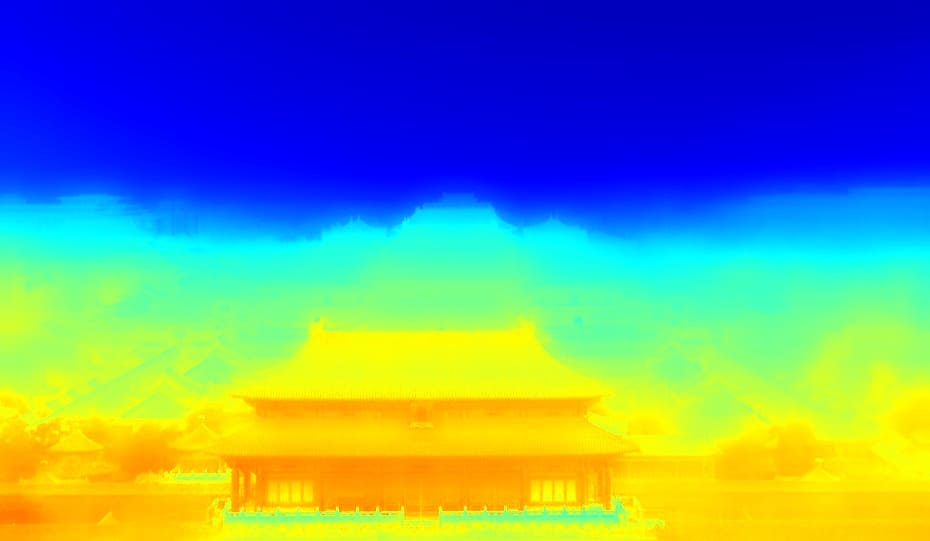}   

    \vspace{0.15cm}
    \includegraphics[width=1.01\linewidth]{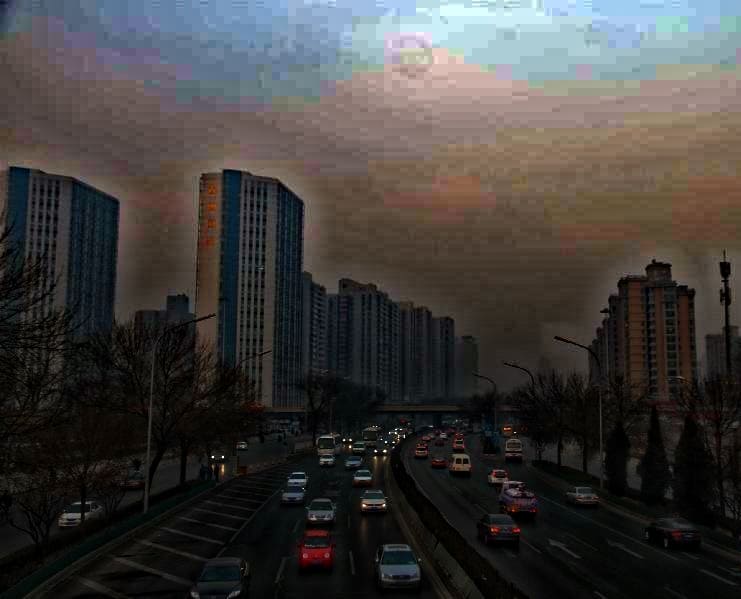}
    
    \vspace{0.08cm}
    \includegraphics[width=1.01\linewidth]{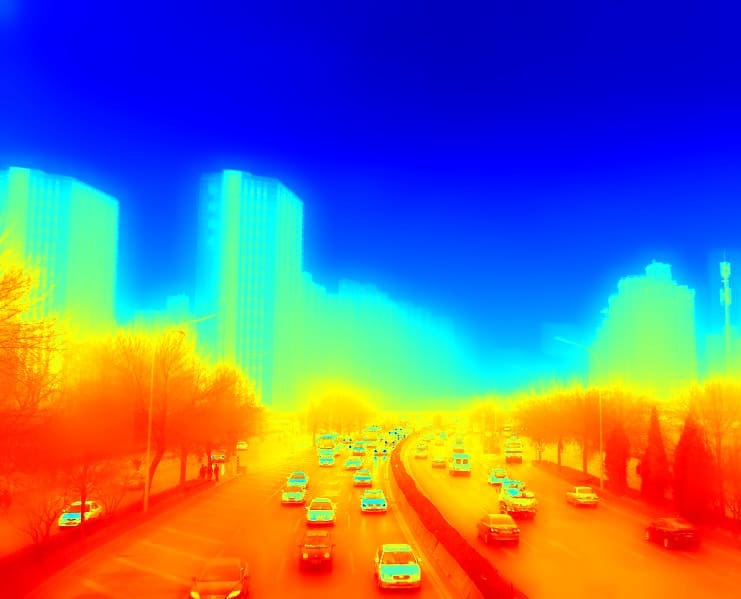}
    \caption{CEP}
    \label{fig14:RESIDE-Natural-b}
  \end{subfigure}
  \hfill
  \begin{subfigure}{0.120\linewidth}
    \includegraphics[width=1.01\linewidth]{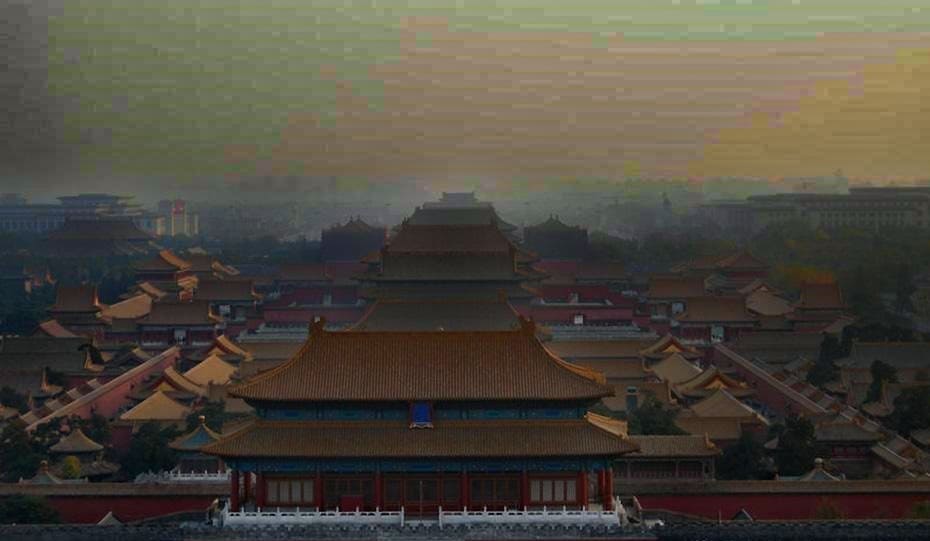}  

    \vspace{0.08cm}
    \includegraphics[width=1.01\linewidth]{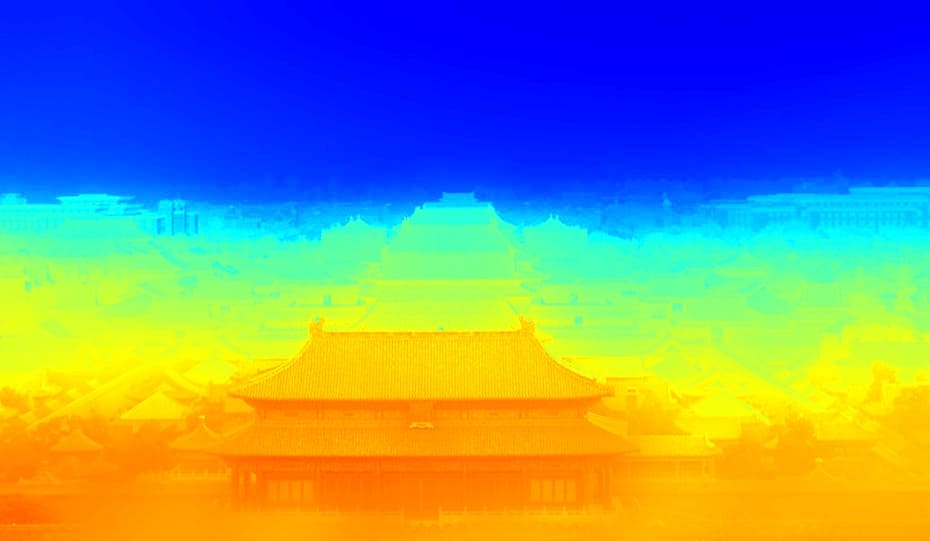}    

    \vspace{0.15cm}
    \includegraphics[width=1.01\linewidth]{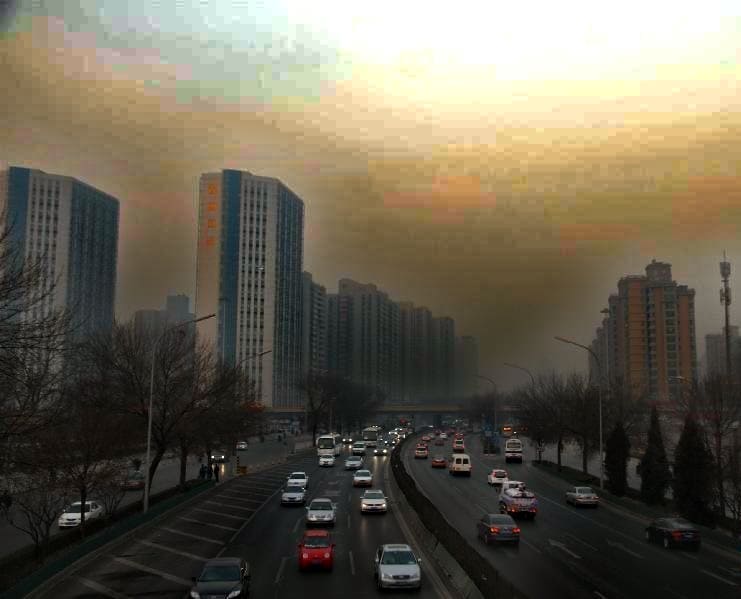}
    
    \vspace{0.08cm}
    \includegraphics[width=1.01\linewidth]{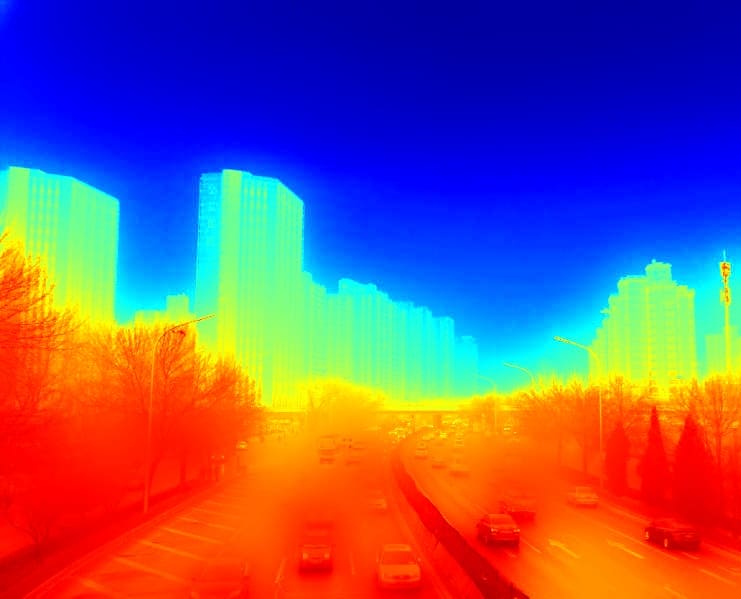}
    \caption{DCP}
    \label{fig14:RESIDE-Natural-c}
  \end{subfigure}
  \hfill
  \begin{subfigure}{0.120\linewidth}
    \includegraphics[width=1.01\linewidth]{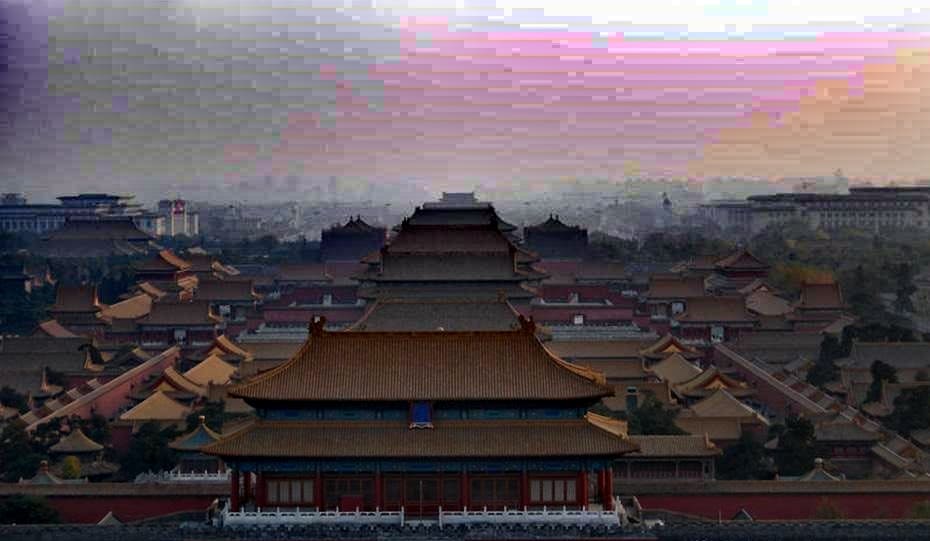}  

    \vspace{0.08cm}
    \includegraphics[width=1.01\linewidth]{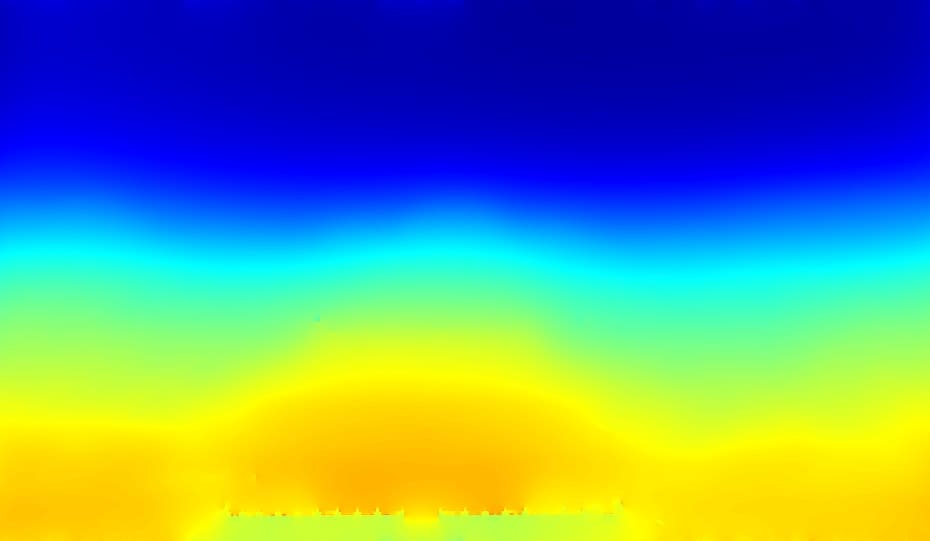}    

    \vspace{0.15cm}
    \includegraphics[width=1.01\linewidth]{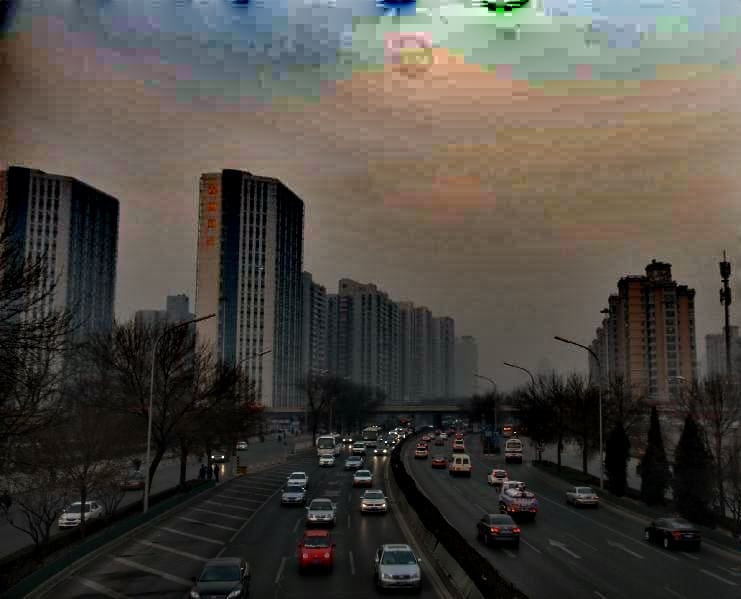}
    
    \vspace{0.08cm}
    \includegraphics[width=1.01\linewidth]{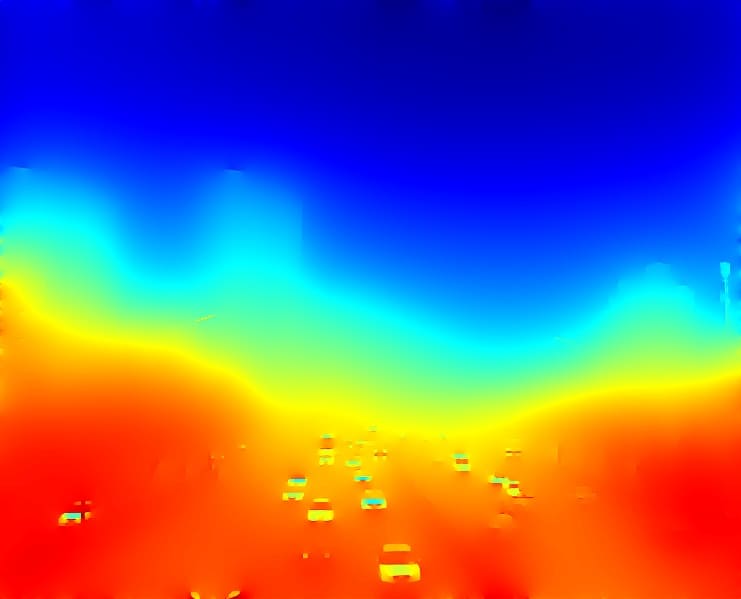}
    \caption{BCCR}
    \label{fig14:RESIDE-Natural-d}
  \end{subfigure}
  \hfill
  \begin{subfigure}{0.120\linewidth}
    \includegraphics[width=1.01\linewidth]{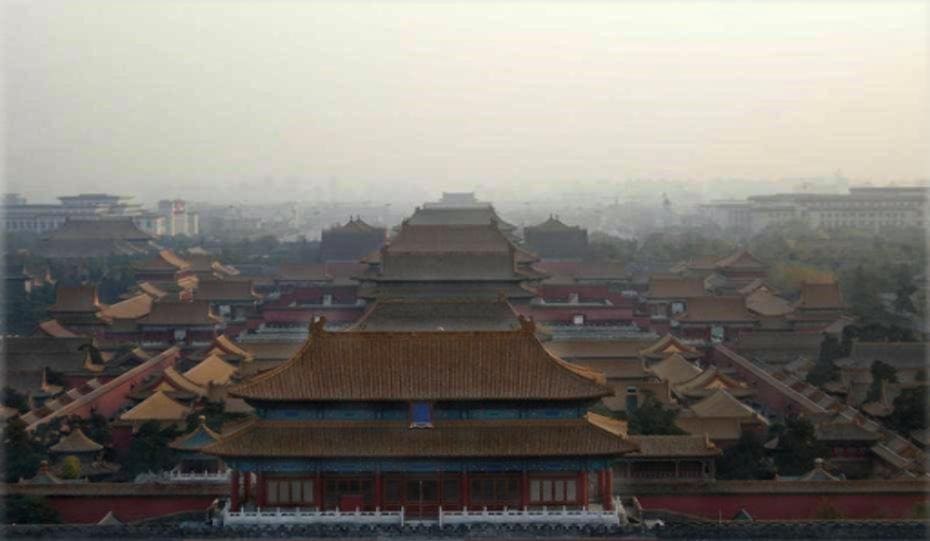}  

    \vspace{0.08cm}
    \includegraphics[width=1.01\linewidth]{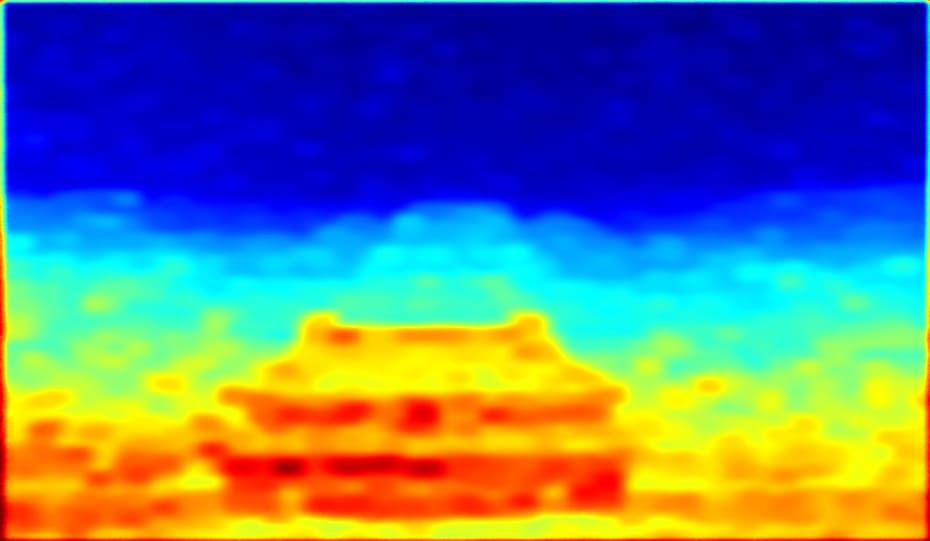}    

    \vspace{0.15cm}
    \includegraphics[width=1.01\linewidth]{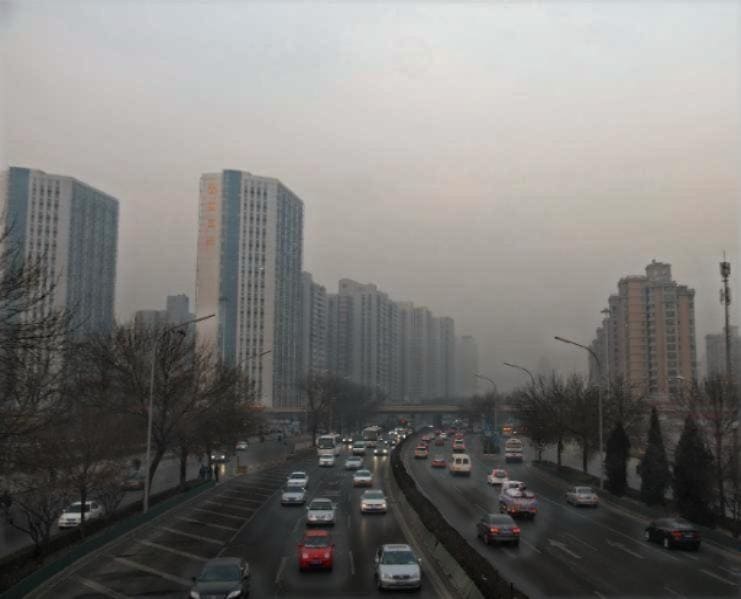}
    
    \vspace{0.08cm}
    \includegraphics[width=1.01\linewidth]{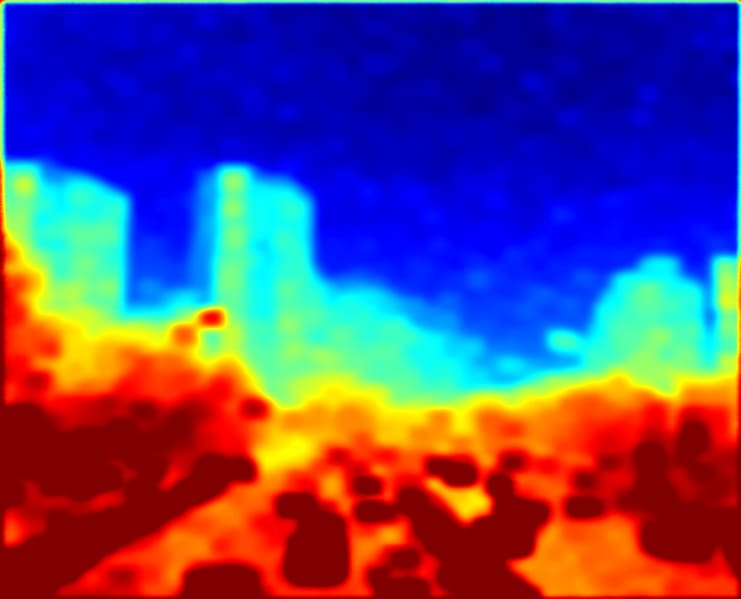}
    \caption{CAP}
    \label{fig14:RESIDE-Natural-e}
  \end{subfigure}
  \hfill
  \begin{subfigure}{0.120\linewidth}
    \includegraphics[width=1.01\linewidth]{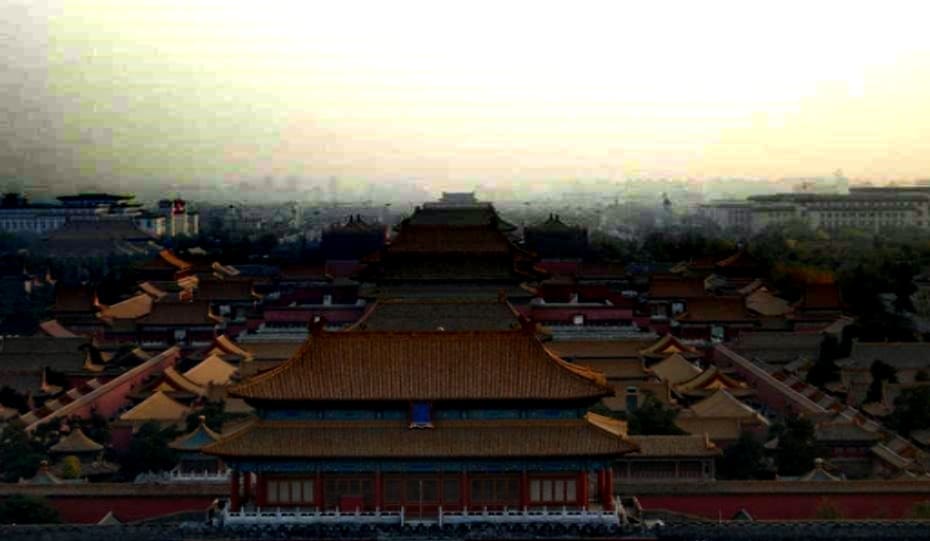}  

    \vspace{0.08cm}
    \includegraphics[width=1.01\linewidth]{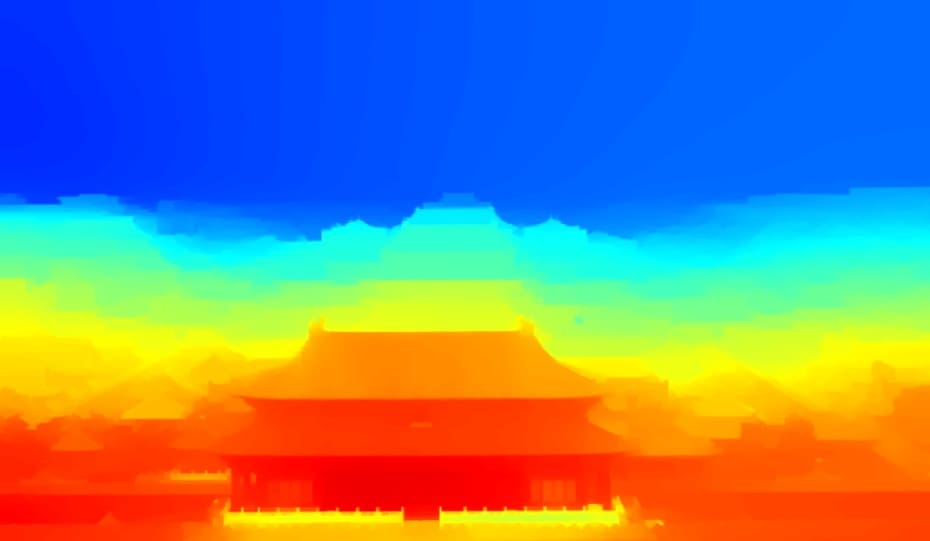}   

    \vspace{0.15cm}
    \includegraphics[width=1.01\linewidth]{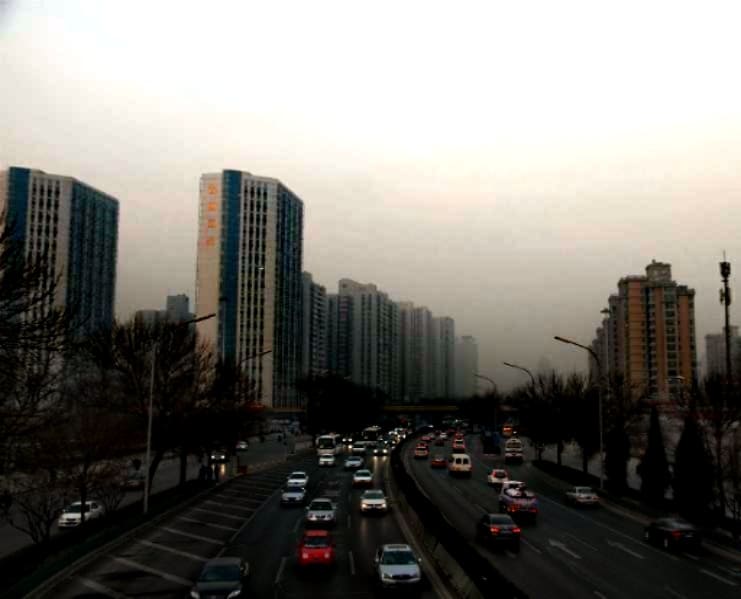}
    
    \vspace{0.08cm}
    \includegraphics[width=1.01\linewidth]{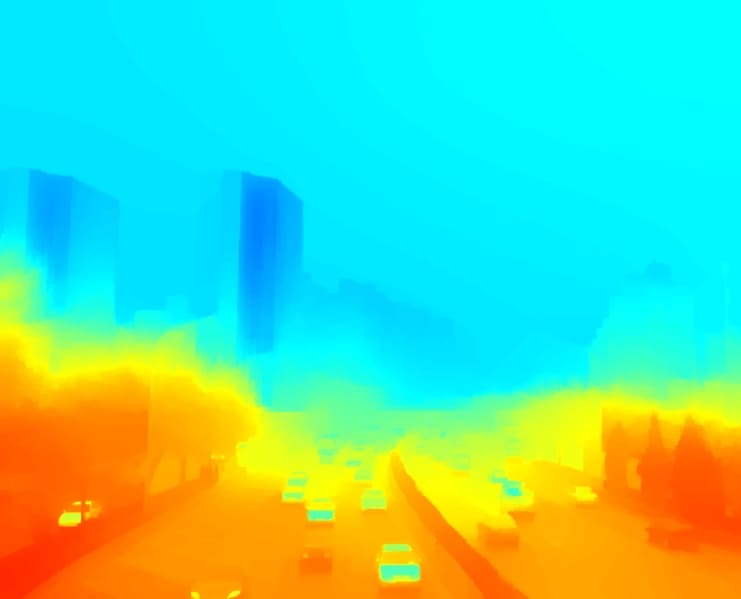} 
    \caption{NLID}
    \label{fig14:RESIDE-Natural-f}
  \end{subfigure}
  \hfill
  \begin{subfigure}{0.120\linewidth}
    \includegraphics[width=1.01\linewidth]{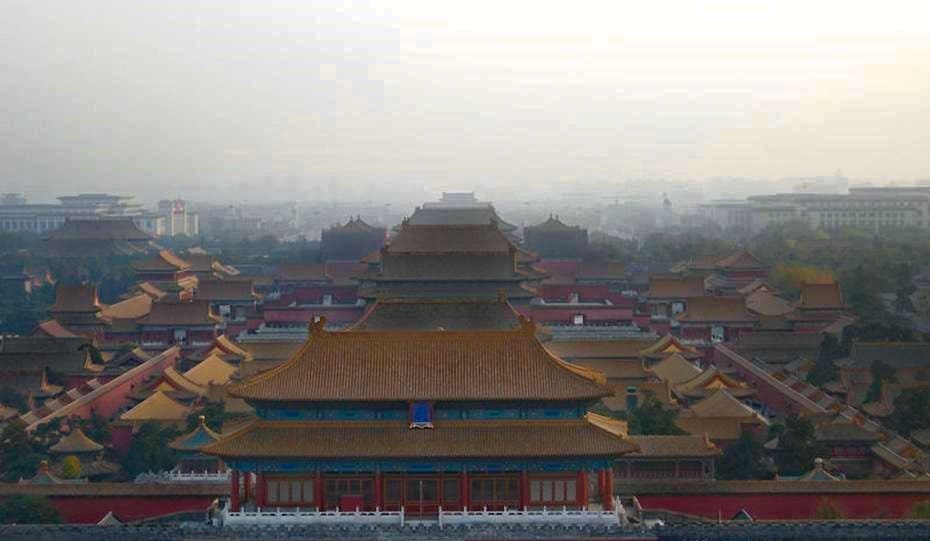}  

    \vspace{0.08cm}
    \includegraphics[width=1.01\linewidth]{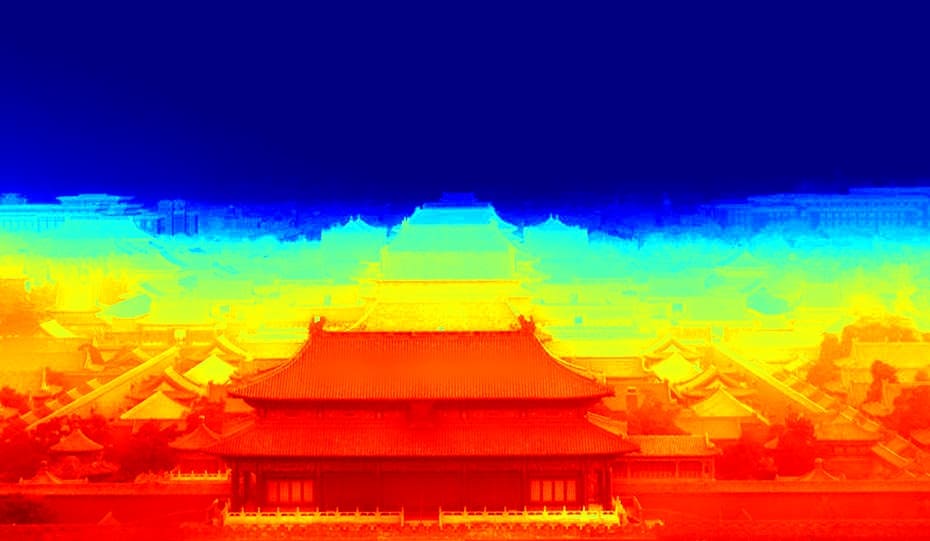}    

    \vspace{0.15cm}
    \includegraphics[width=1.01\linewidth]{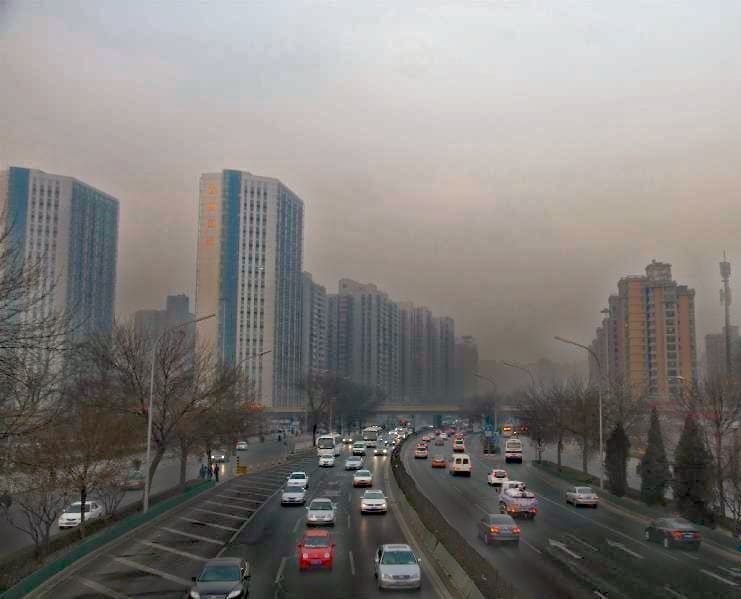}
    
    \vspace{0.08cm}
    \includegraphics[width=1.01\linewidth]{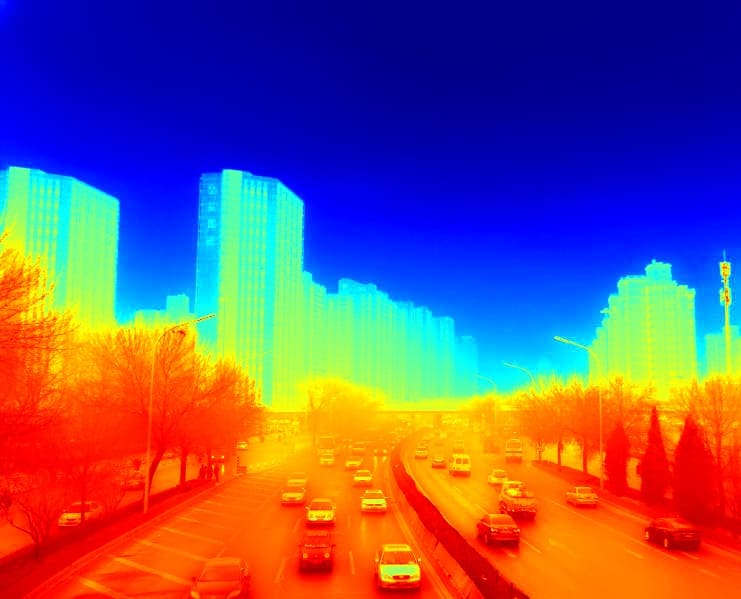}
    \caption{RSVT (ours)}
    \label{fig14:RESIDE-Natural-g}
  \end{subfigure}
  }
  \caption{Typical dehazing results of various approaches on RESIDE-Natural dataset: (a) hazy image, (b-g) results by CEP \cite{bui2017single}, DCP \cite{he2010single}, BCCR \cite{meng2013efficient}, CAP \cite{zhu2015fast}, NLID \cite{berman2016non}, and the proposed RSVT method, respectively. The corresponding recovered transmission map for each method is shown below every output.}
  \label{fig14:RESIDE-Natural}
  
\end{figure*}

\subsection{Dehazing Performance Comparison}
\label{subsec:comparison}

Table \ref{tab02:synthetic-results} summarizes the quantitative results obtained from various dehazing approaches on two synthetic hazy datasets, SOTS-Outdoor and HSTS-Synthetic. As reported in Table \ref{tab02:synthetic-results}, the proposed RSVT framework surpasses all the other existing prior-based methods under comparison by a significant margin in terms of both PSNR and SSIM measures. Moreover, our proposed approach can yield competitive performance when compared against various existing deep learning-based models such as GCANet and YOLY, particularly in terms of the SSIM criteria. On the other hand, further quantitative comparisons on natural hazy image datasets, HSTS-Realistic and RESIDE-Natural, are also summarized in Table \ref{tab03:natural-results}. It can be observed from Table \ref{tab03:natural-results} that the proposed method can be considered competitive against some of the leading methods in both prior-based and deep learning-based categories.

\begin{figure}
  \centering  
  \resizebox{0.49\textwidth}{!}{
  \begin{subfigure}{0.30\linewidth}
    \includegraphics[width=1.0\linewidth]{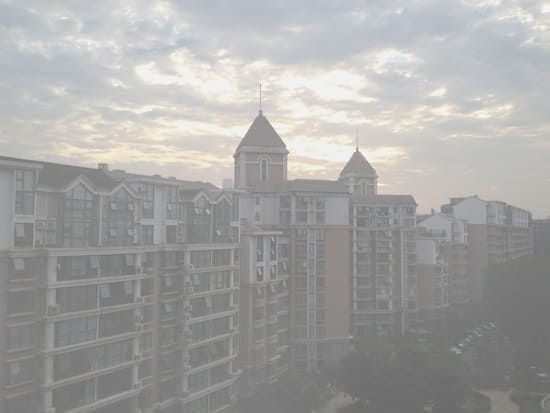} 

    \vspace{0.05cm}
    \includegraphics[width=1.0\linewidth]{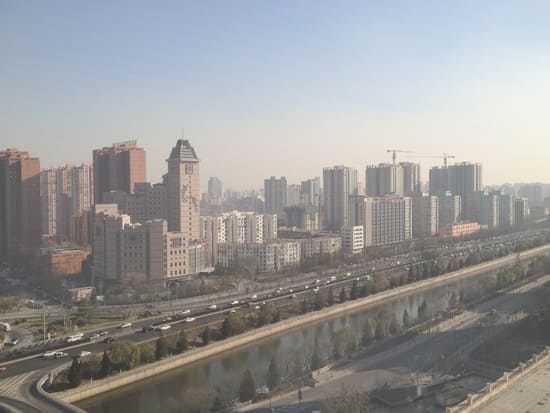} 
    \caption{Hazy}
    \label{fig15:compare-DL-Synt-a}
  \end{subfigure}
  \begin{subfigure}{0.30\linewidth}
    \includegraphics[width=1.0\linewidth]{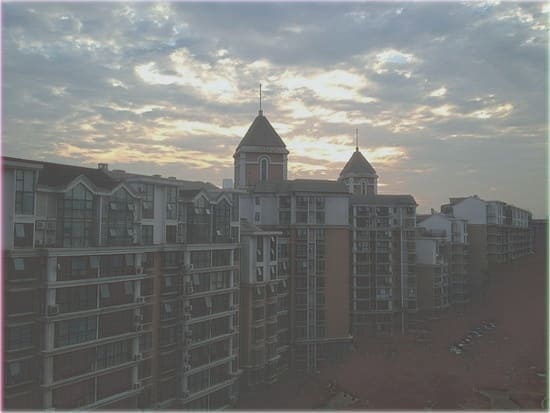}  

    \vspace{0.05cm}
    \includegraphics[width=1.0\linewidth]{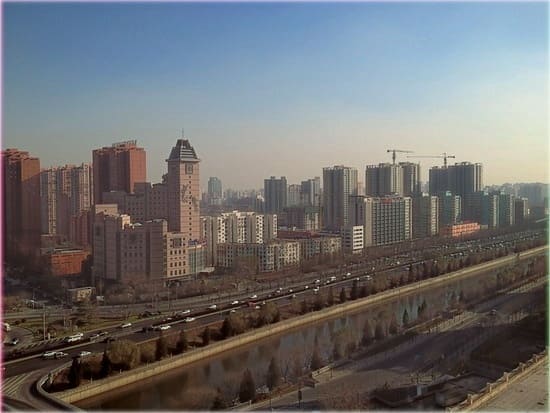}
    \caption{AOD-Net}
    \label{fig15:compare-DL-Synt-b}
  \end{subfigure}
  \begin{subfigure}{0.30\linewidth}
    \includegraphics[width=1.0\linewidth]{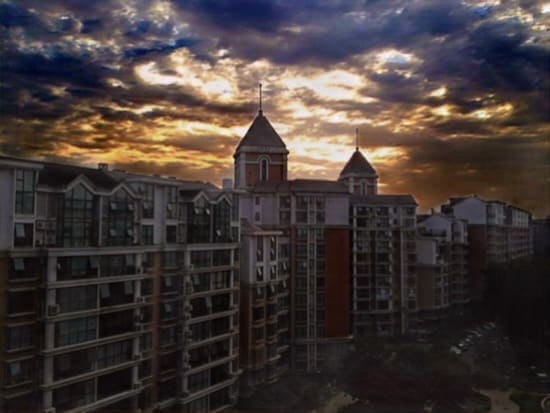} 

    \vspace{0.05cm}
    \includegraphics[width=1.0\linewidth]{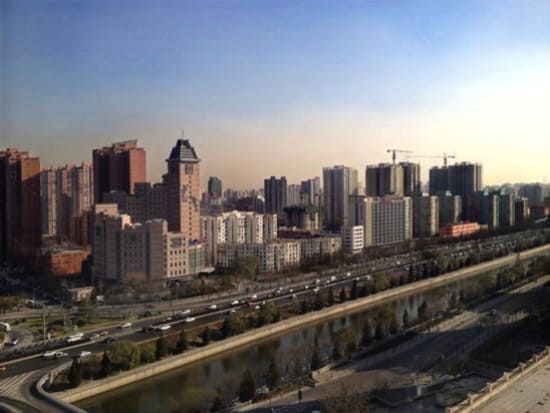} 
    \caption{GCANet}
    \label{fig15:compare-DL-Synt-c}
  \end{subfigure}
  }

  \vspace{0.2cm}
  
  \resizebox{0.49\textwidth}{!}{
  \begin{subfigure}{0.30\linewidth}
    \includegraphics[width=1.0\linewidth]{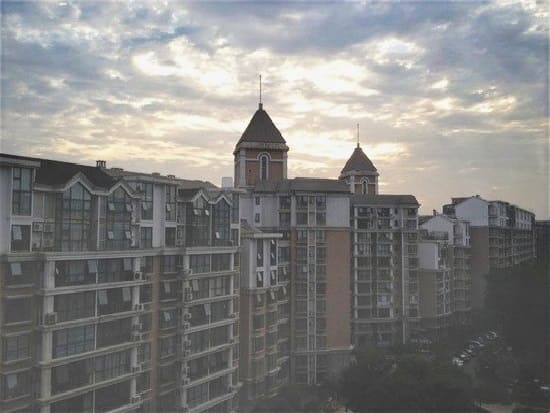} 

    \vspace{0.05cm}
    \includegraphics[width=1.0\linewidth]{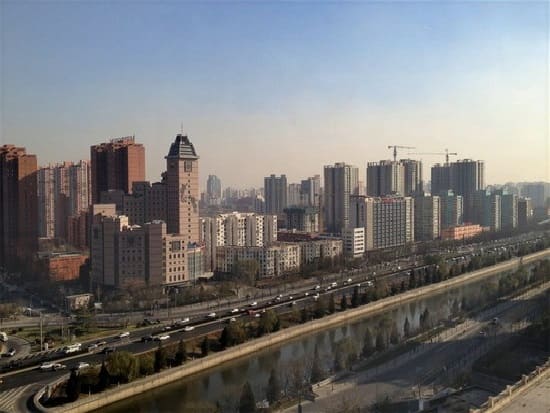}
    \caption{YOLY}
    \label{fig15:compare-DL-Synt-d}
  \end{subfigure}
  \begin{subfigure}{0.30\linewidth}
    \includegraphics[width=1.0\linewidth]{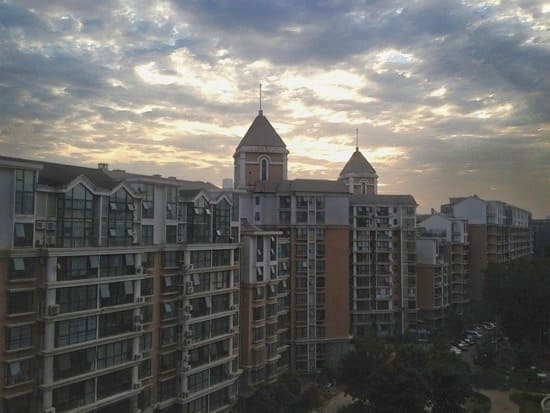} 

    \vspace{0.05cm}
    \includegraphics[width=1.0\linewidth]{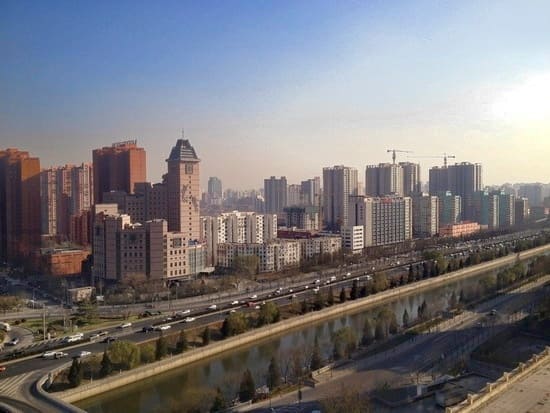}
    \caption{RSVT (ours)}
    \label{fig15:compare-DL-Synt-e}
  \end{subfigure}
  \begin{subfigure}{0.30\linewidth}
    \includegraphics[width=1.0\linewidth]{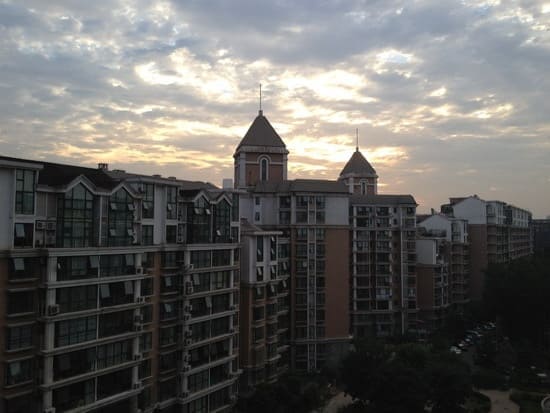} 

    \vspace{0.05cm}
    \includegraphics[width=1.0\linewidth]{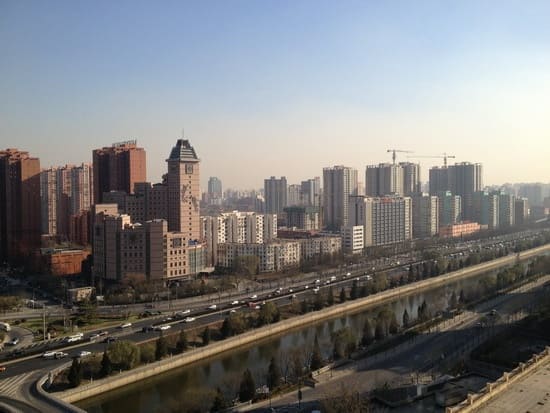}
    \caption{Clean}
    \label{fig15:compare-DL-Synt-f}
  \end{subfigure}
  }
  \caption{Dehazing results of the proposed RSVT framework in comparison to those of some deep learning-based methods on synthetic hazy images: (a) hazy image, (b,c,d,e) results by AOD-Net \cite{li2017aod}, GCANet \cite{chen2019gated}, YOLY \cite{li2021you}, and the proposed RSVT method, respectively, and (f) clean image.}
  \label{fig15:compare-DL-Synt}
  
\end{figure}

\begin{figure}
  \centering  
  \resizebox{0.49\textwidth}{!}{
  \begin{subfigure}{0.30\linewidth}
    \includegraphics[width=1.0\linewidth]{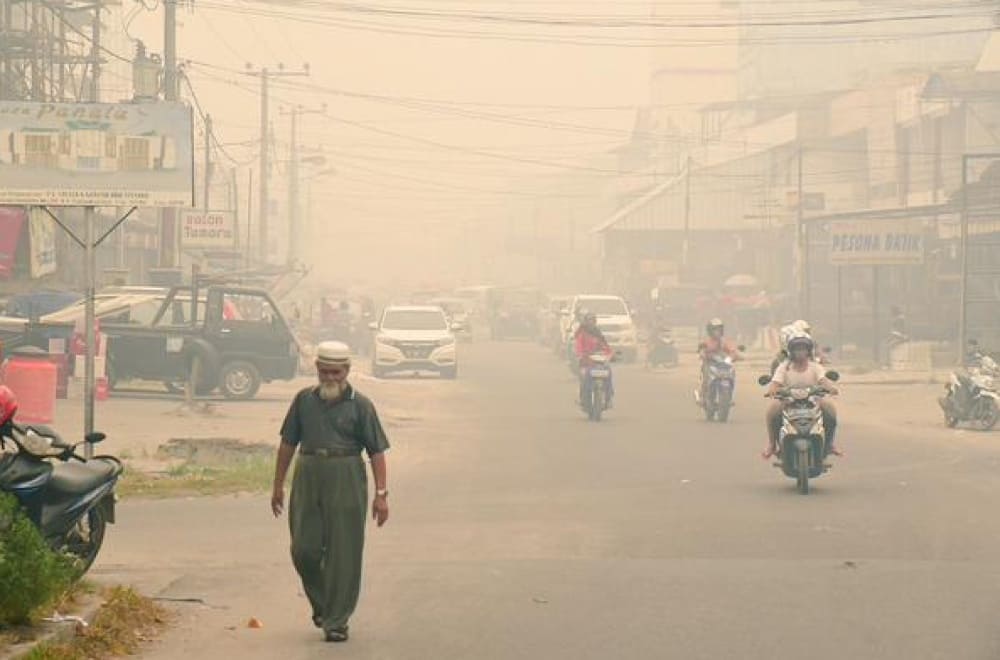} 

    \vspace{0.05cm}
    \includegraphics[width=1.0\linewidth]{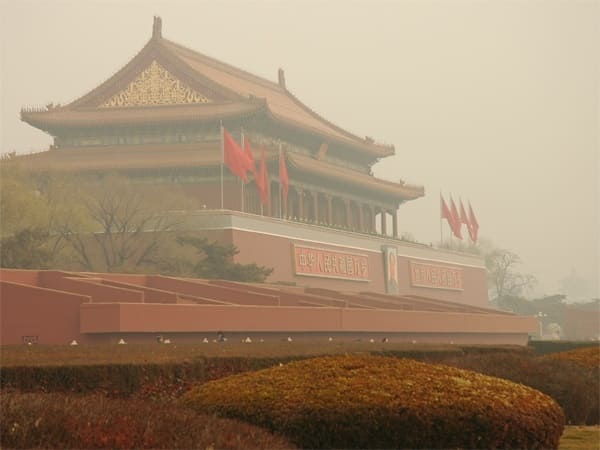} 
    \caption{Hazy}
    \label{fig16:compare-DL-Real-a0}
  \end{subfigure}
  \begin{subfigure}{0.30\linewidth}
    \includegraphics[width=1.0\linewidth]{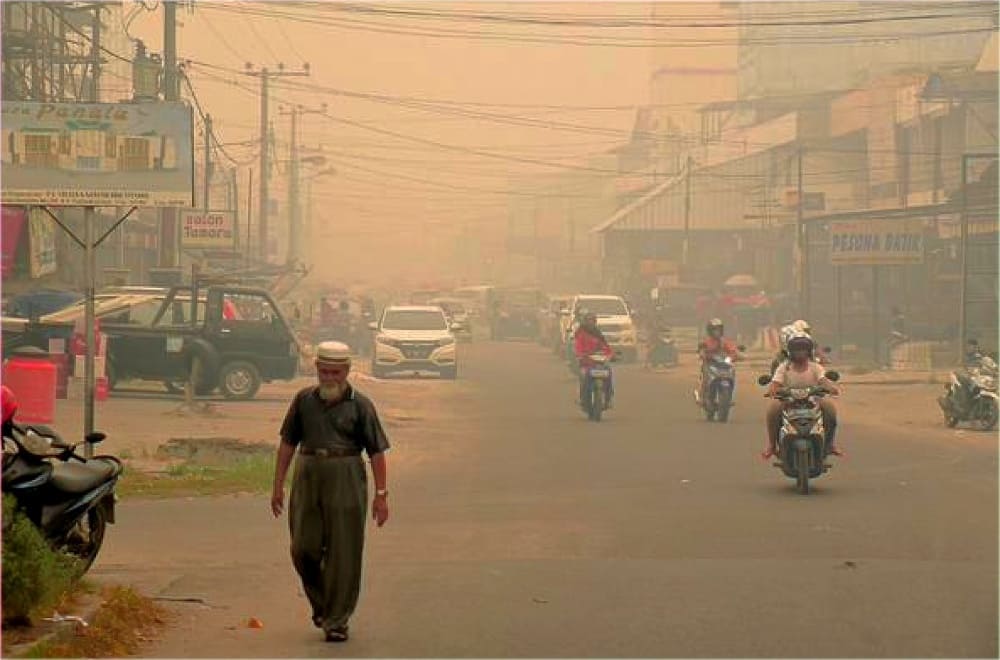}  

    \vspace{0.05cm}
    \includegraphics[width=1.0\linewidth]{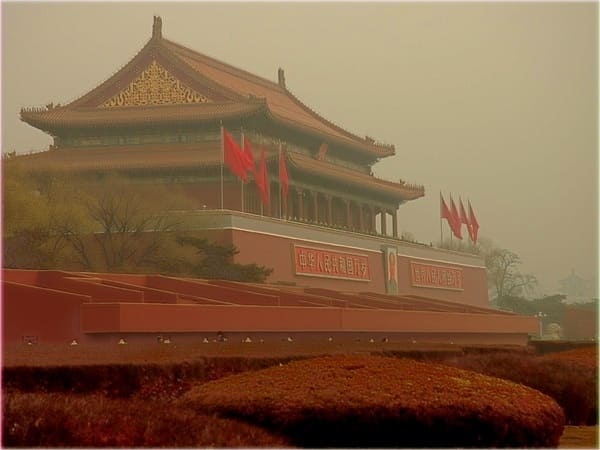}
    \caption{AOD-Net}
    \label{fig16:compare-DL-Real-a}
  \end{subfigure}
  \begin{subfigure}{0.30\linewidth}
    \includegraphics[width=1.0\linewidth]{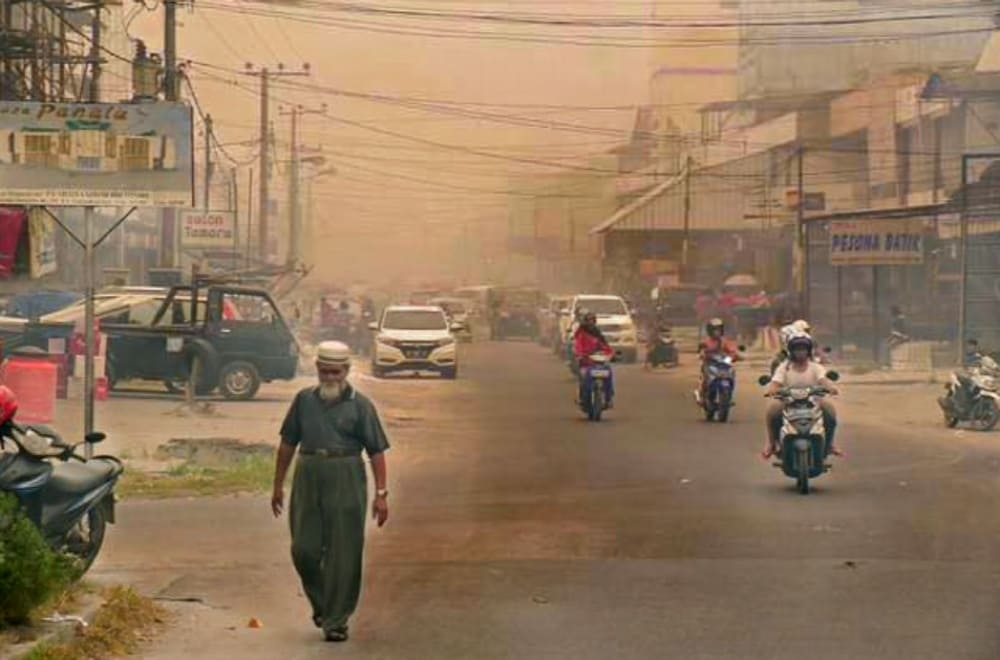} 

    \vspace{0.05cm}
    \includegraphics[width=1.0\linewidth]{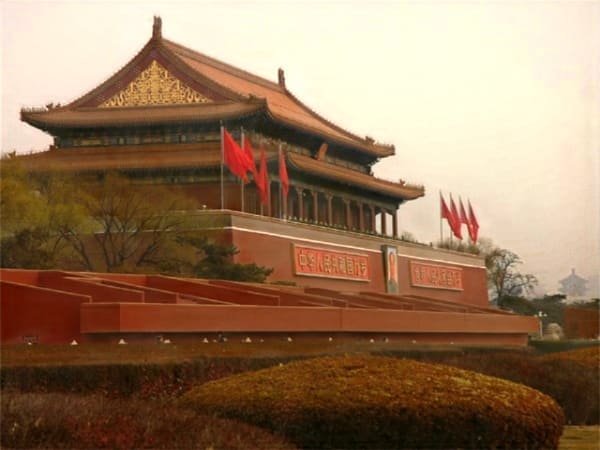} 
    \caption{GCANet}
    \label{fig16:compare-DL-Real-b}
  \end{subfigure}
  }

  \vspace{0.2cm}
  
  \resizebox{0.49\textwidth}{!}{
  \begin{subfigure}{0.30\linewidth}
    \includegraphics[width=1.0\linewidth]{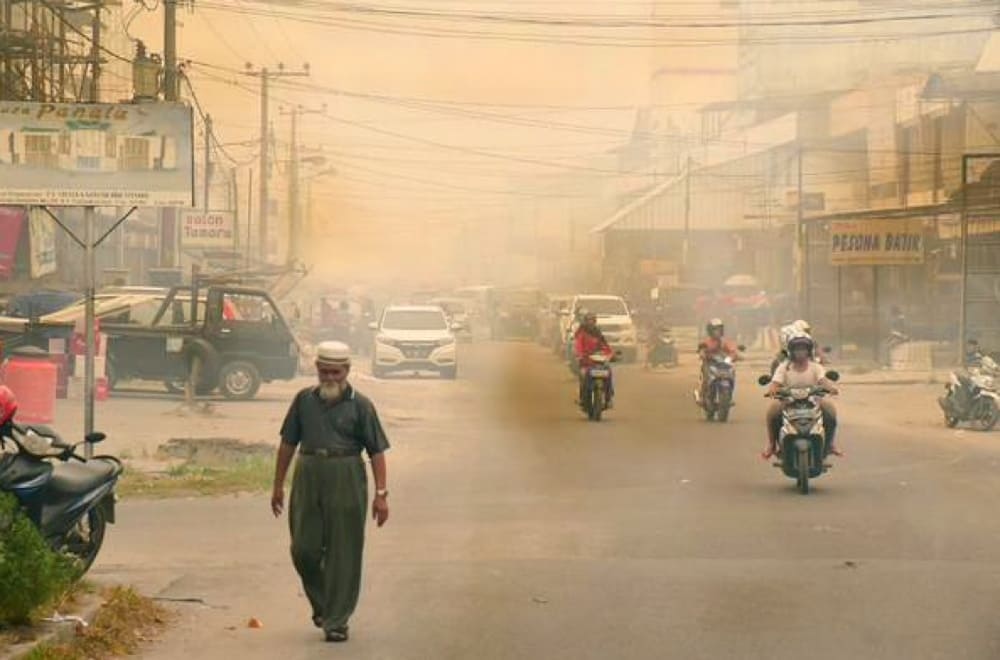} 

    \vspace{0.05cm}
    \includegraphics[width=1.0\linewidth]{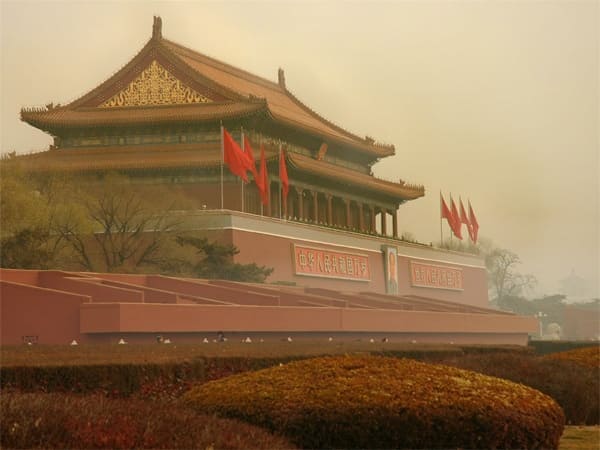}
    \caption{GFN}
    \label{fig16:compare-DL-Real-d}
  \end{subfigure}
  \begin{subfigure}{0.30\linewidth}
    \includegraphics[width=1.0\linewidth]{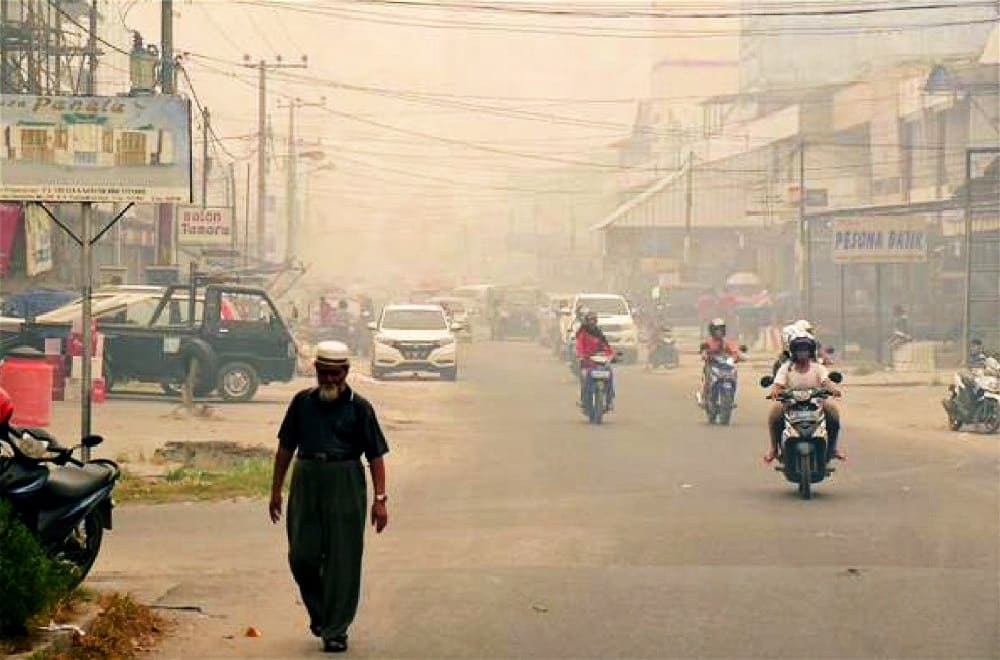} 

    \vspace{0.05cm}
    \includegraphics[width=1.0\linewidth]{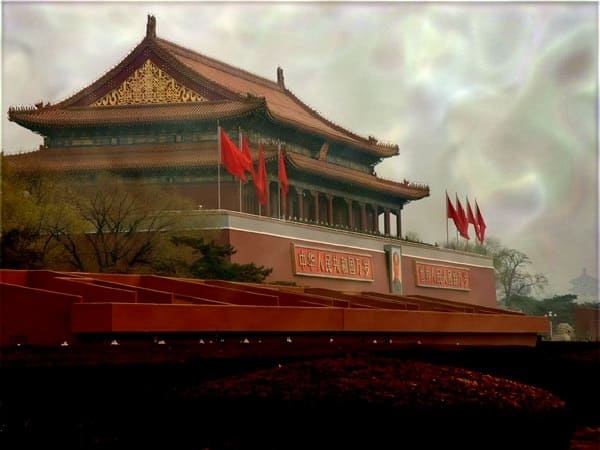}
    \caption{YOLY}
    \label{fig16:compare-DL-Real-e}
  \end{subfigure}
  \begin{subfigure}{0.30\linewidth}
    \includegraphics[width=1.0\linewidth]{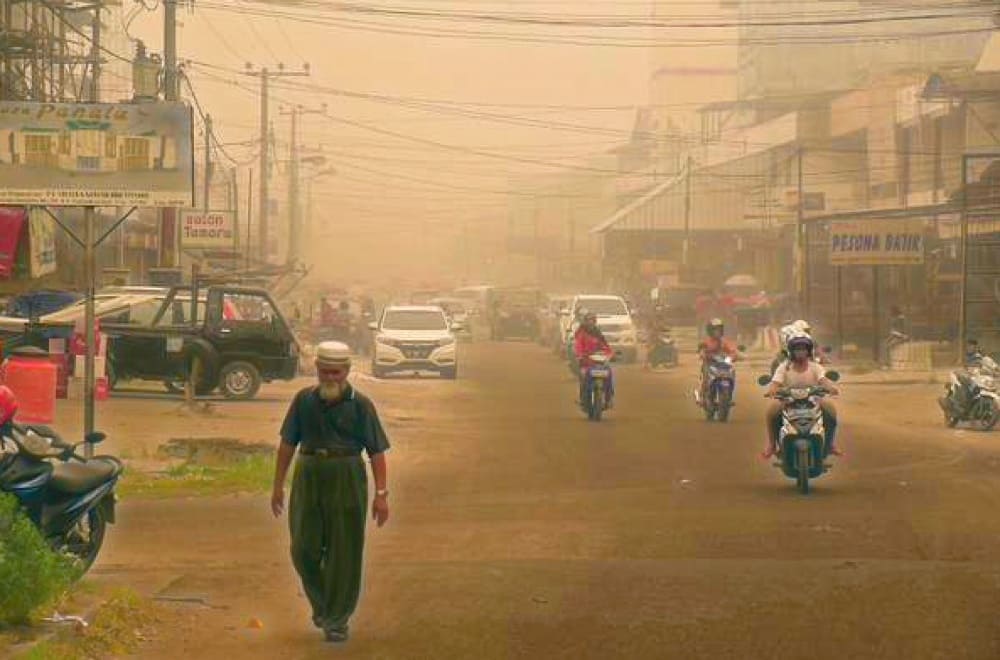} 

    \vspace{0.05cm}
    \includegraphics[width=1.0\linewidth]{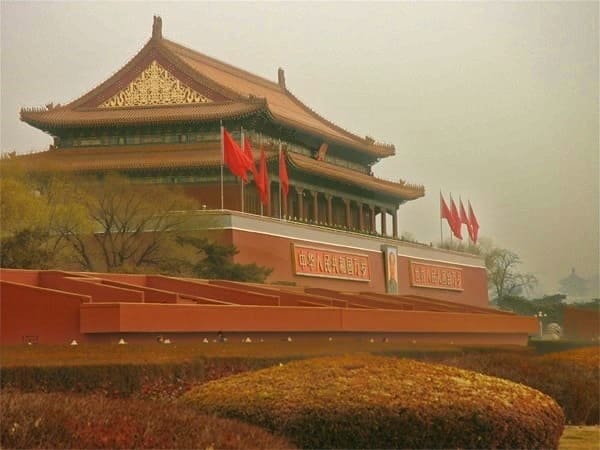}
    \caption{RSVT (ours)}
    \label{fig16:compare-DL-Real-f}
  \end{subfigure}
  }
  \caption{Dehazing results of the proposed RSVT framework in comparison to those of some deep learning-based methods on natural hazy images: (a) hazy, (b,c,d,e,f) results by AOD-Net \cite{li2017aod}, GCANet \cite{chen2019gated}, GFN \cite{ren2018gated}, YOLY \cite{li2021you}, and the proposed RSVT method, respectively.}
  \label{fig16:compare-DL-Real}
  
\end{figure}

Additionally, the comparisons in terms of visibility obtained by our proposed framework and the other prior-based methods on the considered datasets are also given in Fig.s \ref{fig11:SOTS-Outdoor}-\ref{fig14:RESIDE-Natural}. These visual results demonstrate that our proposed method is capable of producing visually appealing haze-free images without halo artifacts or color distortion. Notably, the sky regions restored by our proposed method can be considered significantly improved when compared to those produced by the other approaches. It implies that our proposed prior effectively addresses the color distortion problem in the bright regions, which is a common issue in most other prior-based algorithms. Moreover, the process of RSVT, which is independent of the haze imaging model, allows the proposed method to eliminate the influence of transmission lower bound for the sky regions. As a result, the soft segmentation mask can be utilized as weight to emphasize the transmission in the sky areas. The restored transmission maps shown in Fig.s \ref{fig11:SOTS-Outdoor}-\ref{fig14:RESIDE-Natural} indicate that the depth information provided by our proposed method, particularly in the sky regions, is more accurate when compared against those given by the other prior-based methods under consideration. In addition, typical comparisons between the outputs yielded by our method and various deep learning-based models are also depicted in Fig. \ref{fig15:compare-DL-Synt} and Fig. \ref{fig16:compare-DL-Real}. These comparisons clearly show that the proposed approach achieves competitive visual effects when compared with deep learning-based methods for both synthetic and natural hazy scenes.

\subsection{Execution Speed}

Efficiency is also considered a crucial factor when evaluating an image enhancement method. It partially reflects the suitability of the algorithm for real-world applications such as autonomous driving and surveillance \cite{tran2018vision, tran2019robust, tran2021enhancement, tran2024toward}. In order to assess the efficiency of our method, a comparative analysis with several typical dehazing approaches in terms of average execution time has been conducted, with the results reported in Table \ref{tab04:runtime}. As summarized in Table \ref{tab04:runtime}, the proposed dehazing framework takes approximately 0.231 and 0.364 seconds to process a $600 \times 450$ image and a $1280 \times 720$ image, respectively, with an Intel(R) Core(TM) i5-8600K CPU @ 3.60GHz. As compared against the other prior-based algorithms, even though the proposed approach may not be considered the fastest method, it can manage to achieve a more desirable trade-off between effectiveness and efficiency. Particularly, when compared with DCP or CEP, our method achieves notable improvement in effectiveness while only slightly increasing the processing time. In addition, our method can also be compared competitively with other lightweight deep learning-based approaches such as GCANet while noting that our proposed method does not require an extensive training process on any database. In conclusion, the proposed algorithm demonstrates remarkable efficiency, making it a feasible choice for real-world applications.


\begin{table}
  \caption{Average processing time of various dehazing methods (in seconds).}
  \centering
  \resizebox{0.40\textwidth}{!}{
  \begin{tabular}{cccc}
    \toprule
    \multirow{2}{*}{Method} & \multirow{2}{*}{Platform} & \multicolumn{2}{c}{Runtime (sec.)} \\
    \cmidrule{3-4}
           & & 600$\times$450 & 1280$\times$720 \\
    \midrule
    CEP \cite{bui2017single} & Python (CPU) & \textbf{0.101} & \textbf{0.182} \\
    DCP \cite{he2010single} & Python (CPU) & \textcolor{blue}{0.112} & \textcolor{blue}{0.204} \\
    BCCR \cite{meng2013efficient} & Python (CPU) & 0.445 & 0.865 \\
    NLID \cite{berman2016non} & Python (CPU) & 1.675 & 3.565 \\
    CAP \cite{zhu2015fast} & Python (CPU) & 0.689 & 1.243 \\
    AOD-Net \cite{li2017aod} & PyTorch (CPU) & 0.127 & 0.242 \\
    GCANet \cite{chen2019gated} & PyTorch (CPU) & 0.209 & 0.385 \\
    YOLY \cite{li2021you} & PyTorch (CPU) & 5.126 & 8.542 \\
    \midrule
    RSVT (ours) & Python (CPU) & 0.231 & 0.364 \\
    \bottomrule
  \end{tabular}}
  \label{tab04:runtime}
\end{table}

\section{Conclusions}
\label{sec:conclusions}

In this paper, an innovative prior for image dehazing, called Regional Saturation-Value Transition (RSVT), is proposed. The prior is derived based on two key observations made in the sky and bright areas of hazy-clean image pairs. First, the difference in hue components of a pair of hazy and haze-free points is trivial, indicating that haze has a negligible impact on the hue channel. Second, in the 2D coordinate system formed by the saturation and value channels, most of the lines passing through corresponding pairs of hazy-clean points intersect around the atmospheric light coordinates. This suggests that haze removal can be performed by appropriately translating hazy points in the saturation-value coordinate system. This proposed prior is combined with image decomposition and dark channel prior (DCP) to form a unified dehazing framework. A newly introduced morphological min-max channel can be utilized for image decomposition and estimation of the global atmospheric light. The experimental results have demonstrated that the proposed dehazing framework can perform effectively across a wide range of hazy scenes and can particularly restore the natural appearance of the sky. 


\bibliographystyle{IEEEtran}
\bibliography{ref.bib}

\begin{thebibliography}{10}
\providecommand{\url}[1]{#1}
\csname url@samestyle\endcsname
\providecommand{\newblock}{\relax}
\providecommand{\bibinfo}[2]{#2}
\providecommand{\BIBentrySTDinterwordspacing}{\spaceskip=0pt\relax}
\providecommand{\BIBentryALTinterwordstretchfactor}{4}
\providecommand{\BIBentryALTinterwordspacing}{\spaceskip=\fontdimen2\font plus
\BIBentryALTinterwordstretchfactor\fontdimen3\font minus \fontdimen4\font\relax}
\providecommand{\BIBforeignlanguage}[2]{{%
\expandafter\ifx\csname l@#1\endcsname\relax
\typeout{** WARNING: IEEEtran.bst: No hyphenation pattern has been}%
\typeout{** loaded for the language `#1'. Using the pattern for}%
\typeout{** the default language instead.}%
\else
\language=\csname l@#1\endcsname
\fi
#2}}
\providecommand{\BIBdecl}{\relax}
\BIBdecl

\bibitem{zhao2021refinednet}
S.~Zhao, L.~Zhang, Y.~Shen, and Y.~Zhou, ``Refinednet: A weakly supervised refinement framework for single image dehazing,'' \emph{IEEE Transactions on Image Processing}, vol.~30, pp. 3391--3404, 2021.

\bibitem{he2010single}
K.~He, J.~Sun, and X.~Tang, ``Single image haze removal using dark channel prior,'' \emph{IEEE transactions on pattern analysis and machine intelligence}, vol.~33, no.~12, pp. 2341--2353, 2010.

\bibitem{bui2017single}
T.~M. Bui and W.~Kim, ``Single image dehazing using color ellipsoid prior,'' \emph{IEEE Transactions on Image Processing}, vol.~27, no.~2, pp. 999--1009, 2017.

\bibitem{berman2016non}
D.~Berman, S.~Avidan \emph{et~al.}, ``Non-local image dehazing,'' in \emph{Proceedings of the IEEE conference on computer vision and pattern recognition}, 2016, pp. 1674--1682.

\bibitem{shi2014single}
Z.~Shi, J.~Long, W.~Tang, and C.~Zhang, ``Single image dehazing in inhomogeneous atmosphere,'' \emph{Optik}, vol. 125, no.~15, pp. 3868--3875, 2014.

\bibitem{wang2017dehazing}
W.~Wang, X.~Yuan, X.~Wu, and Y.~Liu, ``Dehazing for images with large sky region,'' \emph{Neurocomputing}, vol. 238, pp. 365--376, 2017.

\bibitem{salazar2020fast}
S.~Salazar-Colores, E.~U. Moya-Sanchez, J.-M. Ramos-Arreguin, E.~Cabal-Yepez, G.~Flores, and U.~Cortes, ``Fast single image defogging with robust sky detection,'' \emph{IEEE access}, vol.~8, pp. 149\,176--149\,189, 2020.

\bibitem{yu2016image}
F.~Yu, C.~Qing, X.~Xu, and B.~Cai, ``Image and video dehazing using view-based cluster segmentation,'' in \emph{2016 Visual Communications and Image Processing (VCIP)}.\hskip 1em plus 0.5em minus 0.4em\relax IEEE, 2016, pp. 1--4.

\bibitem{tran2023single}
L.-A. Tran, D.~Kwon, and D.-C. Park, ``Single image dehazing via regional saturation-value translation,'' in \emph{International Conference on Industry Science and Computer Sciences Innovation}.\hskip 1em plus 0.5em minus 0.4em\relax Elsevier, 2023.

\bibitem{tran2022anovel}
L.-A. Tran, S.~Moon, and D.-C. Park, ``A novel encoder-decoder network with guided transmission map for single image dehazing,'' \emph{Procedia Computer Science}, vol. 204, pp. 682--689, 2022.

\bibitem{middleton1954vision}
W.~Middleton and V.~Twersky, ``Vision through the atmosphere,'' \emph{Physics Today}, vol.~7, no.~3, p.~21, 1954.

\bibitem{fattal2008single}
R.~Fattal, ``Single image dehazing,'' \emph{ACM transactions on graphics (TOG)}, vol.~27, no.~3, pp. 1--9, 2008.

\bibitem{meng2013efficient}
G.~Meng, Y.~Wang, J.~Duan, S.~Xiang, and C.~Pan, ``Efficient image dehazing with boundary constraint and contextual regularization,'' in \emph{Proceedings of the IEEE international conference on computer vision}, 2013, pp. 617--624.

\bibitem{zhu2015fast}
Q.~Zhu, J.~Mai, and L.~Shao, ``A fast single image haze removal algorithm using color attenuation prior,'' \emph{IEEE transactions on image processing}, vol.~24, no.~11, pp. 3522--3533, 2015.

\bibitem{nguyen20133d}
A.~Nguyen and B.~Le, ``3d point cloud segmentation: A survey,'' in \emph{2013 6th IEEE conference on robotics, automation and mechatronics (RAM)}.\hskip 1em plus 0.5em minus 0.4em\relax IEEE, 2013, pp. 225--230.

\bibitem{liu2017single}
Y.~Liu, H.~Li, and M.~Wang, ``Single image dehazing via large sky region segmentation and multiscale opening dark channel model,'' \emph{IEEE Access}, vol.~5, pp. 8890--8903, 2017.

\bibitem{liba2020sky}
O.~Liba, L.~Cai, Y.-T. Tsai, E.~Eban, Y.~Movshovitz-Attias, Y.~Pritch, H.~Chen, and J.~T. Barron, ``Sky optimization: Semantically aware image processing of skies in low-light photography,'' in \emph{Proceedings of the IEEE/CVF Conference on Computer Vision and Pattern Recognition Workshops}, 2020, pp. 526--527.

\bibitem{li2017aod}
B.~Li, X.~Peng, Z.~Wang, J.~Xu, and D.~Feng, ``Aod-net: All-in-one dehazing network,'' in \emph{Proceedings of the IEEE international conference on computer vision}, 2017, pp. 4770--4778.

\bibitem{ren2016single}
W.~Ren, S.~Liu, H.~Zhang, J.~Pan, X.~Cao, and M.-H. Yang, ``Single image dehazing via multi-scale convolutional neural networks,'' in \emph{European conference on computer vision}.\hskip 1em plus 0.5em minus 0.4em\relax Springer, 2016, pp. 154--169.

\bibitem{cai2016dehazenet}
B.~Cai, X.~Xu, K.~Jia, C.~Qing, and D.~Tao, ``Dehazenet: An end-to-end system for single image haze removal,'' \emph{IEEE Transactions on Image Processing}, vol.~25, no.~11, pp. 5187--5198, 2016.

\bibitem{he2012guided}
K.~He, J.~Sun, and X.~Tang, ``Guided image filtering,'' \emph{IEEE transactions on pattern analysis and machine intelligence}, vol.~35, no.~6, pp. 1397--1409, 2012.

\bibitem{li2018benchmarking}
B.~Li, W.~Ren, D.~Fu, D.~Tao, D.~Feng, W.~Zeng, and Z.~Wang, ``Benchmarking single-image dehazing and beyond,'' \emph{IEEE Transactions on Image Processing}, vol.~28, no.~1, pp. 492--505, 2018.

\bibitem{huang2021image}
S.-C. Huang, D.-W. Jaw, W.~Li, Z.~Lu, S.-Y. Kuo, B.~C. Fung, B.-H. Chen, and T.~Numnonda, ``Image dehazing in disproportionate haze distributions,'' \emph{IEEE Access}, vol.~9, pp. 44\,599--44\,609, 2021.

\bibitem{tao2017low}
L.~Tao, C.~Zhu, J.~Song, T.~Lu, H.~Jia, and X.~Xie, ``Low-light image enhancement using cnn and bright channel prior,'' in \emph{2017 IEEE International Conference on Image Processing (ICIP)}.\hskip 1em plus 0.5em minus 0.4em\relax IEEE, 2017, pp. 3215--3219.

\bibitem{kanopoulos1988design}
N.~Kanopoulos, N.~Vasanthavada, and R.~L. Baker, ``Design of an image edge detection filter using the sobel operator,'' \emph{IEEE Journal of solid-state circuits}, vol.~23, no.~2, pp. 358--367, 1988.

\bibitem{li2014weighted}
Z.~Li, J.~Zheng, Z.~Zhu, W.~Yao, and S.~Wu, ``Weighted guided image filtering,'' \emph{IEEE Transactions on Image processing}, vol.~24, no.~1, pp. 120--129, 2014.

\bibitem{qin2020u2}
X.~Qin, Z.~Zhang, C.~Huang, M.~Dehghan, O.~R. Zaiane, and M.~Jagersand, ``U2-net: Going deeper with nested u-structure for salient object detection,'' \emph{Pattern recognition}, vol. 106, p. 107404, 2020.

\bibitem{pizer1987adaptive}
S.~M. Pizer, E.~P. Amburn, J.~D. Austin, R.~Cromartie, A.~Geselowitz, T.~Greer, B.~ter Haar~Romeny, J.~B. Zimmerman, and K.~Zuiderveld, ``Adaptive histogram equalization and its variations,'' \emph{Computer vision, graphics, and image processing}, vol.~39, no.~3, pp. 355--368, 1987.

\bibitem{ren2018gated}
W.~Ren, L.~Ma, J.~Zhang, J.~Pan, X.~Cao, W.~Liu, and M.-H. Yang, ``Gated fusion network for single image dehazing,'' in \emph{Proceedings of the IEEE conference on computer vision and pattern recognition}, 2018, pp. 3253--3261.

\bibitem{chen2019gated}
D.~Chen, M.~He, Q.~Fan, J.~Liao, L.~Zhang, D.~Hou, L.~Yuan, and G.~Hua, ``Gated context aggregation network for image dehazing and deraining,'' in \emph{2019 IEEE winter conference on applications of computer vision (WACV)}.\hskip 1em plus 0.5em minus 0.4em\relax IEEE, 2019, pp. 1375--1383.

\bibitem{li2021you}
B.~Li, Y.~Gou, S.~Gu, J.~Z. Liu, J.~T. Zhou, and X.~Peng, ``You only look yourself: Unsupervised and untrained single image dehazing neural network,'' \emph{International Journal of Computer Vision}, vol. 129, no.~5, pp. 1754--1767, 2021.

\bibitem{mittal2012making}
A.~Mittal, R.~Soundararajan, and A.~C. Bovik, ``Making a “completely blind” image quality analyzer,'' \emph{IEEE Signal processing letters}, vol.~20, no.~3, pp. 209--212, 2012.

\bibitem{mittal2012no}
A.~Mittal, A.~K. Moorthy, and A.~C. Bovik, ``No-reference image quality assessment in the spatial domain,'' \emph{IEEE Transactions on image processing}, vol.~21, no.~12, pp. 4695--4708, 2012.

\bibitem{pal1993review}
N.~R. Pal and S.~K. Pal, ``A review on image segmentation techniques,'' \emph{Pattern recognition}, vol.~26, no.~9, pp. 1277--1294, 1993.

\bibitem{titterington1985statistical}
D.~M. Titterington, A.~F. Smith, and U.~E. Makov, ``Statistical analysis of finite mixture distributions,'' \emph{(No Title)}, 1985.

\bibitem{tran2018vision}
L.-A. Tran, N.-P. Le, T.-D. Do, and M.-H. Le, ``A vision-based method for autonomous landing on a target with a quadcopter,'' in \emph{2018 4th International Conference on Green Technology and Sustainable Development (GTSD)}.\hskip 1em plus 0.5em minus 0.4em\relax IEEE, 2018, pp. 601--606.

\bibitem{tran2019robust}
L.-A. Tran and M.-H. Le, ``Robust u-net-based road lane markings detection for autonomous driving,'' in \emph{2019 International Conference on System Science and Engineering (ICSSE)}.\hskip 1em plus 0.5em minus 0.4em\relax IEEE, 2019, pp. 62--66.

\bibitem{tran2021enhancement}
L.-A. Tran, T.-D. Do, D.-C. Park, and M.-H. Le, ``Enhancement of robustness in object detection module for advanced driver assistance systems,'' in \emph{2021 International Conference on System Science and Engineering (ICSSE)}.\hskip 1em plus 0.5em minus 0.4em\relax IEEE, 2021, pp. 158--163.

\bibitem{tran2024toward}
L.-A. Tran, C.~N. Tran, D.-C. Park, J.~Carrabina, and D.~Castells-Rufas, ``Toward improving robustness of object detectors against domain shift,'' in \emph{2024 IEEE International Conference on Green Energy, Computing and Sustainable Technology (GECOST)}.\hskip 1em plus 0.5em minus 0.4em\relax IEEE, 2024.

\end{thebibliography}

\end{document}